\documentclass{article} 

\usepackage{jmlr2e}

\usepackage{lastpage}

\ShortHeadings{DVMIB -- A Framework for Variational Losses}{Abdelaleem, Nemenman, Martini}
\firstpageno{1}
\usepackage{bm}

\usepackage{hyperref}
\usepackage{url}
\usepackage{multirow}
\usepackage{wrapfig}
\usepackage{amssymb}

\usepackage[utf8]{inputenc} 
\usepackage[T1]{fontenc}    
\usepackage{url}            
\usepackage{booktabs}       
\usepackage{amsfonts}       
\usepackage{nicefrac}       
\usepackage{microtype}      
\usepackage{xcolor}         
\usepackage{amsmath}
\usepackage{graphicx}
\usepackage{tikz}
\usetikzlibrary{positioning, quotes}
\usepackage{enumerate}
\usepackage{adjustbox}
\usepackage{array} 
\usetikzlibrary{shapes, arrows.meta}
\usepackage{float}
\usepackage{minitoc}  

\usepackage{longtable} 

\title{Deep Variational Multivariate Information Bottleneck -\\ A Framework for Variational Losses}

\author{\name Eslam Abdelaleem \thanks{Equal contribution} ~\thanks{Currently eabdelaleem3@gatech.edu at the Schools of Physics and Psychology - Georgia Institute of Technology} \email eslam.abdelaleem@emory.edu\\
\addr Department of Physics\\
Emory University\\
Atlanta, GA 30322, USA \\
\AND
\name Ilya Nemenman \email ilya.nemenman@emory.edu\\
\addr Departments of Physics and Biology\\
Initiative in Theory and Modeling of Living Systems\\
Emory University \\
Atlanta, GA 30322, USA \\
\AND
\name K. Michael Martini \footnotemark[1]~~\thanks{Corresponding author} \email karl.michael.martini@emory.edu \\
\addr Department of Physics \\
Emory University \\
Atlanta, GA 30322, USA \\
}

\editor{}
\begin{document}
\doparttoc 
\faketableofcontents 

\part{} 

\maketitle
\begin{abstract}
Variational dimensionality reduction methods are widely used for their accuracy, generative capabilities, and robustness. We introduce a unifying framework that generalizes both such as traditional and state-of-the-art methods. The framework is based on an interpretation of the multivariate information bottleneck, trading off the information preserved in an encoder graph (defining what to compress) against that in a decoder graph (defining a generative model for data). Using this approach, we rederive existing methods, including the deep variational information bottleneck, variational autoencoders, and deep multiview information bottleneck. We naturally extend the deep variational CCA (DVCCA) family to beta-DVCCA and introduce a new method, the deep variational symmetric information bottleneck (DVSIB). DSIB, the deterministic limit of DVSIB, connects to modern contrastive learning approaches such as Barlow Twins, among others. We evaluate these methods on Noisy MNIST and Noisy CIFAR-100, showing that algorithms better matched to the structure of the problem like DVSIB and beta-DVCCA produce better latent spaces as measured by classification accuracy, dimensionality of the latent variables, sample efficiency, and consistently outperform other approaches under comparable conditions. Additionally, we benchmark against state-of-the-art models, achieving superior or competitive accuracy. Our results demonstrate that this framework can seamlessly incorporate diverse multi-view representation learning algorithms, providing a foundation for designing novel, problem-specific loss functions.
\end{abstract}

\begin{keywords}
  Information Bottleneck, Symmetric Information Bottleneck, Variational Methods, Generative Models, Dimensionality Reduction, Data Efficiency
\end{keywords}

\section{Introduction}

Large dimensional multi-modal datasets are abundant in multimedia systems utilized for language modeling \citep{Rui2016, Corso2018, Schiele2017, Metze2018, Parikh2017, Bowman2018, Steinhardt2020},  neural control of behavior studies \citep{Harris2021, Churchland2022, Poeppel2017, Fairhall2016}, multi-omics approaches in systems biology \citep{Clark2013, Bielas2017, Teichmann2018, O'Donovan2015, Lorenzi2018}, and many other domains. Such data come with the curse of dimensionality, making it hard to learn the relevant statistical correlations from samples. The problem is made even harder by the data often containing information that is irrelevant to the specific questions one asks. To tackle these challenges, a myriad of supervised and unsupervised dimensionality reduction (DR) methods have emerged. By preserving certain aspects of the data while discarding the remainder, DR can decrease the complexity of the problem, yield clearer insights, and provide a foundation for more refined modeling approaches.

DR techniques span linear methods like Principal Component Analysis (PCA) \citep{Hotelling1933}, Partial Least Squares (PLS) \citep{Eriksson2001}, Canonical Correlations Analysis (CCA) \citep{Hotelling1936}, and regularized CCA \citep{Vinod1976, Strother1998}, as well as nonlinear approaches, including Autoencoders (AE) \citep{Salakhutdinov2006}, Deep CCA \citep{Livescu2013}, Deep Canonical Correlated AE \citep{Bilmes2015}, Correlational Neural Networks \citep{Ravindran2015}, Deep Generalized CCA \citep{Arora2017}, and Deep Tensor CCA \citep{Zeng2021}. Of particular interest to us are variational methods, such as Variational Autoencoders (VAE) \citep{Welling2014}, beta-VAE \citep{Lerchner2016}, Joint Multimodal VAE (JMVAE) \citep{Matsuo2016}, Deep Variational CCA (DVCCA) \citep{Livescu2016}, Deep Variational Information Bottleneck (DVIB) \citep{Murphy2017}, Variational Mixture-of-experts AE \citep{shi2019}, and Multiview Information Bottleneck \citep{federici2020}. These  DR methods use deep neural networks and variational approximations to learn robust and accurate representations of the data, while, at the same time, often serving as generative models for creating samples from the learned distributions.

There are many theoretical derivations and justifications for variational DR methods \citep{Welling2014, Lerchner2016, Matsuo2016, Livescu2016, Schuurmans2021, Lin2022, Murphy2017, Bao2021, VanderSchaar2021, Zhou2019, Hu2021, Akata2020, Elgamal2022, Ye2020}. This diversity of derivations, while enabling adaptability, often leaves researchers with no principled ways for choosing a method for a particular application, for designing new methods with distinct assumptions, or for comparing methods to each other.

Here, we introduce the Deep Variational Multivariate Information Bottleneck (DVMIB) framework, offering a unified mathematical foundation for many variational---and deterministic---DR methods. Our framework is grounded in the multivariate information bottleneck loss function \citep{Bialek2000, Tishby2013}. This loss, amenable to approximation through upper and lower variational bounds, provides a system for implementing diverse DR variants using deep neural networks. We demonstrate the framework's efficacy by deriving the loss functions of many existing DR methods starting from the same principles. These include well-known methods, such as AEs \citep{Salakhutdinov2006}, VAEs \citep{Welling2014}, and state-of-the-art methods such as  Contrastive Language-Image Pretraining (CLIP) \citep{radford2021learning} and Barlow Twins \citep{zbontar2021}. Furthermore, our framework naturally allows the adjustment of trade-off parameters, leading to generalizations of these existing methods. For instance, we generalize DVCCA \citep{Livescu2016} to $\beta$-DVCCA. The framework further allows us to introduce and implement in software novel DR methods. We view the DVMIB framework, with its uniform information bottleneck language, conceptual clarity of translating statistical dependencies in data via graphical models of encoder and decoder structures into variational losses, the ability to unify existing approaches, and easy adaptability to new scenarios as one of the main contributions of our work. 

Beyond its unifying role, our framework offers a principled approach for deriving problem-specific loss functions using domain-specific knowledge. Thus, we anticipate its application for multi-view representation learning across diverse fields. To illustrate this, we use the framework to derive a novel dimensionality reduction method,  the Deep Variational Symmetric Information Bottleneck (DVSIB), which compresses two random variables into two distinct latent variables that are maximally informative about one another. This new method produces better representations of classic datasets than previous approaches. The introduction of DVSIB is another major contribution of our paper.

In summary, our paper makes the following contributions to the field:
\begin{enumerate}
    \item \textbf{Introduction of the Variational Multivariate Information Bottleneck Framework:} We provide both intuitive and mathematical insights into this framework, establishing a robust foundation for further exploration.
    
    \item \textbf{Rederivation and Generalization of Existing Methods within a Common Framework:} We demonstrate the versatility of our framework by systematically rederiving and generalizing various existing methods from the literature, showcasing the framework's ability to unify diverse approaches.
    
    \item \textbf{Design of a Novel Method — Deep Variational Symmetric Information Bottleneck (DVSIB):} Employing our framework, we introduce DVSIB as a new method, contributing to the growing repertoire of techniques in variational dimensionality reduction. The method constructs high-accuracy latent spaces from substantially fewer samples than comparable approaches. Additionally, its deterministic version, the Deterministic Symmetric Information Bottleneck (DSIB), can be mapped to a plethora of state-of-the-art methods including CLIP and Barlow Twins. 
\end{enumerate}

The paper is structured as follows: First, in Sec.~\ref{framework}, we introduce the underlying mathematics and an implementation of the DVMIB framework. Then, in Sec.~\ref{different_dr_methods}, we explain how to use the framework to generate new DR methods. In this section, in Tbl.~\ref{table:methods}, we present several known and newly variational DR methods, illustrating how easily they can be derived within the framework. In the Results (Sec.~\ref{results_all}), as a proof of concept, we first benchmark {\em simple} computational implementations of the methods in Tbl.~\ref{table:methods} against the Noisy MNIST dataset (Sec.~\ref{results_noisy_mnist}). We then examine whether the trends observed in Noisy MNIST hold for more complex datasets, i.e., ~Noisy CIFAR-100, and more complex neural network architectures, namely CNNs (Sec.~\ref{results_noisy_cifar}). Next, we demonstrate how the framework extends to more advanced, state-of-the-art methods (Sec.~\ref{results_sota}). The Appendices provide a detailed treatment of all terms in the different loss functions introduced (Appx.~\ref{App:Library}). Appendix~\ref{App:private} discusses auxiliary private variable models. Additional details, including visualizations, and a comprehensive performance analysis of many methods on Noisy MNIST are available in Appx.~\ref{App:results-mnist}. Further results for Noisy CIFAR-100 are provided in Appx.~\ref{app:cifar}.

\section{Multivariate Information Bottleneck Framework}
\label{framework}

We represent DR problems similar to the Multivariate Information Bottleneck (MIB) of \citet{Tishby2013}, which is a generalization of the more traditional Information Bottleneck algorithm \citep{Bialek2000} to multiple variables. The reduced representation is achieved as a trade-off between two Bayesian networks. Bayesian networks are directed acyclic graphs that provide a factorization of the joint probability distribution, $P(X_1,X_2,X_3,..,X_N)=\prod_{i=1}^N P(X_i|Pa_{X_i}^G)$, where $Pa_{X_i}^G$ is the set of parents of $X_i$ in graph $G$.  The multiinformation \citep{Vejnarova1998} of a Bayesian network is defined as the Kullback-Leibler divergence between the joint probability distribution and the product of the marginals, and it serves as a measure of the total correlations among the variables, $I(X_1,X_2,X_3,...,X_N)=D_{KL}(P(X_1,X_2,X_3,...,X_N) \Vert P(X_1)P(X_2)P(X_3)...P(X_N))$. For a Bayesian network, the multiinformation reduces to the sum of all the local informations $I(X_1,X_2,..X_N)=\sum_{i=1}^N I(X_i;Pa_{X_i}^G)$ \citep{Tishby2013}. 

The first of the Bayesian networks is an encoder (compression) graph, which models how compressed (reduced, latent) variables are obtained from the observations. The second network is a decoder graph, which specifies a generative model for the data from the compressed variables, i.e., it is an alternate factorization of the distribution. In MIB, the information of the encoder graph is minimized, ensuring strong compression (corresponding to the approximate posterior).  The information of the decoder graph is maximized, promoting the most accurate model of the data (corresponding to maximizing the log-likelihood). As in  IB \citep{Bialek2000}, the trade-off between the compression and reconstruction is controlled by a trade-off parameter $\beta$:
\begin{equation}
L=I_{\text{encoder}}-\beta I_{\text{decoder}}.
\label{multiInfo}
\end{equation}

In this work, our key contribution is in writing an explicit variational loss for typical information terms found in both the encoder and the decoder graphs.  All terms in the decoder graph use samples of the compressed variables as determined from the encoder graph.  If there are two terms that correspond to the same information in Eq.~(\ref{multiInfo}), one from each of the graphs, they do not cancel each other since they correspond to two different variational expressions. For pedagogical clarity, we do this by first analyzing the Symmetric Information Bottleneck (SIB), a {\em special case} of MIB. We derive the bounds for three types of information terms in SIB, which we then use as building blocks for all other variational MIB methods in subsequent sections.

\subsection{Deep Variational Symmetric Information Bottleneck}\label{sec:DVSIB}
The Deep Variational Symmetric Information Bottleneck (DVSIB)  simultaneously reduces a pair of datasets $X$ and $Y$ into two separate lower dimensional compressed versions $Z_X$ and $Z_Y$. These compressions are done at the same time to ensure that the latent spaces are maximally informative about each other. The joint compression is known to decrease data set size requirements compared to individual ones \citep{Nemenman2023}.  Having distinct latent spaces for each modality usually helps with interpretability. For example, $X$ could be the neural activity of thousands of neurons, and $Y$ could be the recordings of joint angles of the animal. Rather than one latent space representing both, separate latent spaces for the neural activity and the joint angles are sought. By maximizing compression as well as $I(Z_X,Z_Y)$, one constructs the latent spaces that capture only the neural activity pertinent to joint movement and only the movement that is correlated with the neural activity (cf.~\cite{Fairhall2016}). Many other applications could benefit from a similar DR approach.

\begin{wrapfigure}{r}{0.4\textwidth}
\begin{center}
\vspace{-.2in}
\adjustbox{width=.4\textwidth}{
    \begin{tikzpicture}[node distance={15mm}, thick,
    main/.style = {draw, circle,minimum size=11mm},
    comp/.style = {draw, circle,minimum size=9mm},
    labs/.style = {}
    ] 
    \node[labs] (a) {$G_{\text{encoder}}$};
    \node[main] (x) [below=4mm of a] {$X$}; 
    \node[main] (y) [right of=x] {$Y$};
    \node[comp] (zx) [below of=x] {$Z_X$}; 
    \node[comp] (zy) [below of=y] {$Z_Y$};
    \draw (x) -- (y);
    \draw[->] (x) -- (zx);
    \draw[->] (y) -- (zy);

    \node[main] (mx) [right of=y] {$X$}; 
    \node[labs] (b) [above=4mm of mx] {$G_{\text{decoder}}$};
    \node[main] (my) [right of=mx] {$Y$};
    \node[comp] (mzx) [below of=mx] {$Z_X$}; 
    \node[comp] (mzy) [below of=my] {$Z_Y$};
    \draw[->] (mzx) -- (mzy);
    \draw[->] (mzx) -- (mx);
    \draw[->] (mzy) -- (my);
    \end{tikzpicture}
}
\end{center}
\caption{The encoder and decoder graphs for DVSIB.}
\label{Fig:Graph}
\end{wrapfigure}

In Fig.~\ref{Fig:Graph}, we define two Bayesian networks for DVSIB, $G_{\text{encoder}}$ and $G_{\text{decoder}}$. $G_{\text{encoder}}$ encodes the compression of $X$ to $Z_X$ and $Y$ to $Z_Y$. It corresponds to the factorization $p(x,y,z_x,z_y)=p(x,y)p(z_x|x)p(z_y|y)$ and the resultant $I_{\text{encoder}} = I^{E}(X;Y)+I^{E}(X;Z_X)+I^{E}(Y;Z_Y)$. The $I^{E}(X,Y)$ term does not depend on the compressed variables, does not affect the optimization problem, and hence is discarded in what follows. $G_{\text{decoder}}$ represents  a generative model for $X$ and $Y$ given the compressed latent variables $Z_X$ and $Z_Y$. It corresponds to the factorization $p(x,y,z_x,z_y)=p(z_x)p(z_y|z_x)p(x|z_x)p(y|z_y)$ and the resultant $I_{\text{decoder}} = I^{D}(Z_X;Z_Y)+I^{D}(X;Z_X)+I^{D}(Y;Z_Y)$.  Combing the informations from both graphs and  using Eq.~(\ref{multiInfo}), we find the SIB loss:
\begin{equation}
L_{\text{SIB}}=I^{E}(X;Z_X)+I^{E}(Y;Z_Y)
-\beta \left(I^{D}(Z_X;Z_Y)+I^{D}(X;Z_X)+I^{D}(Y;Z_Y)\right).
\label{eq:sib}
\end{equation}

Note that information in the encoder terms is minimized, and information in the decoder terms is maximized. Thus, while it is tempting to simplify Eq.~(\ref{eq:sib}) by canceling $I^{E}(X;Z_X)$ and $I^{D}(X;Z_X)$, this would be a mistake.  Indeed,  these terms come from different factorizations:  the encoder corresponds to learning $p(z_x|x)$, and the decoder to $p(x|z_x)$. 

We now follow a procedure and notation similar to \citet{Murphy2017} and construct variational bounds on all $I^{E}$ and $I^{D}$ terms. Terms without leaf nodes, i.~e., $I^{D}(Z_X,Z_Y)$, require new approaches.

\subsection{Variational bounds on DVSIB encoder terms}
\label{sec:encoder}
The information $I^{E}(Z_X;X)$  corresponds to compressing the random variable $X$ to $Z_X$. Since this is an encoder term, it needs to be minimized in Eq.~(\ref{eq:sib}). Thus, we seek a variational bound $I^{E}(Z_X;X)\le \Tilde{I}^{E}(Z_X;X)$, where $\Tilde{I}^{E}$ is the variational version of $I^{E}$, which can be implemented using a deep neural network. We find $\Tilde{I}^E$ by using the positivity of the Kullback–Leibler divergence. We make $r(z_x)$ be a variational approximation to $p(z_x)$. Then $D_{\rm KL} ( p(z_x) \Vert r(z_x) )\ge 0$, so that $-\int dz_x p(z_x) \ln(p(z_x))\le-\int dz_x p(z_x) \ln(r(z_x))$. Thus, $-\int dx dz_x p(z_x,x) \ln(p(z_x))\le-\int dx dz_x p(z_x,x) \ln(r(z_x))$. We then add $\int dx dz_x p(z_x,x) \ln(p(z_x|x))$ to both sides and find: 
\begin{equation}
I^{E}(Z_X;X)=\int dx dz_x p(z_x,x) \ln\left(\frac{p(z_x|x)}{p(z_x)}\right)\le\int dx dz_x p(z_x,x) \ln\left(\frac{p(z_x|x)}{r(z_x)}\right)\equiv\Tilde{I}^{E}(Z_X;X).
\end{equation}
We further simplify the variational loss by approximating $p(x)\approx\frac{1}{N}\sum_{i=1}^N \delta(x-x_i)$, so that:
\begin{equation}
\Tilde{I}^{E}(Z_X;X)\approx\frac{1}{N}\sum_{i=1}^N\int dz_x p(z_x|x_i) \ln\left(\frac{p(z_x|x_i)}{r(z_x)}\right)=\frac{1}{N}\sum_{i=1}^N D_{\rm KL}(p(z_x|x_i) \Vert r(z_x)).
\end{equation}

The term $I^{E}(Y;Z_Y)$ can be treated in an analogous manner, resulting in:
\begin{equation}
\Tilde{I}^{E}(Z_Y;Y) \approx \frac{1}{N}\sum_{i=1}^N D_{KL}(p(z_y|y_i) \Vert r(z_y)).
\end{equation}
Note that, in the deterministic limit, $p(z_x|x)\rightarrow\delta(z_x-f(x))$. In this limit, $I^E(Z_X;X)\to H(Z_X)$, and the latter is nominally infinite, but diverges only logarithmically with the discretization/regularization scale in $z_x$, and linearly with the dimensionality of $Z_X$. Provided $f(x)$ is sufficiently smooth, it does not affect the (regularized)  $H(Z_X)$.  Hence, compression is enforced not by details of the embedding $z_x=f(x)$, but only by the dimensionality of the latent variable. This deterministic limit is relevant for AEs \citep{Salakhutdinov2006} and several other deterministic methods that we will discuss later (cf. Sec.~\ref{results_sota}).

\subsection{Variational bounds on DVSIB decoder terms}
\label{sec:decoder}
The term $I^{D}(X;Z)$  corresponds to a decoder of $X$ from the compressed variable $Z_X$. It is maximized in Eq.~(\ref{eq:sib}). Thus, we seek its variational version $\Tilde{I}^{D}$, such that $I^{D}\ge\Tilde{I}^{D}$. Here, $q(x|z_x)$ will serve as a variational approximation to $p(x|z_x)$. We use the positivity of the Kullback-Leibler divergence, $D_{\rm KL}(p(x|z_x) \Vert q(x|z_x)) \ge 0$, to find $\int dx\, p(x|z_x)\ln(p(x|z_x))\ge\int dx\, p(x|z_x)\ln(q(x|z_x))$. This gives $\int dz_x dx\, p(x,z_x)\ln(p(x|z_x))\ge\int dz_x dx\, p(x,z_x)\ln(q(x|z_x))$. We add the entropy of $X$ to both sides to arrive at the variational bound:
\begin{equation}
I^{D}(X;Z_X)=\int dz_x dx\, p(x,z_x)\ln\frac{p(x|z_x)}{p(x)} \ge \int dz_x dx p(x,z_x)\ln\frac{q(x|z_x)}{p(x)}\equiv\Tilde{I}^{D}(X;Z_X).
\end{equation}

We further simplify $\Tilde{I}^{D}$ by replacing $p(x)$ by samples, $p(x)\approx\frac{1}{N}\sum_i^N\delta(x-x_i)$ and using the $p(z_x|x)$ that we learned previously from the encoder:
\begin{equation}
\Tilde{I}^{D}(X;Z_X)\approx H(X)+\frac{1}{N}\sum_{i=1}^N\int dz_x p(z_x|x_i)\ln(q(x_i|z_x)).
\end{equation}
Here $H(X)$ does not depend on $p(z_x|x)$ and, therefore, can be dropped from the loss. The variational version of $I^{D}(Y;Z_Y)$ is obtained  analogously:
\begin{equation}
\Tilde{I}^{D}(Y;Z_Y)\approx H(Y)+\frac{1}{N}\sum_{i=1}^N\int dz_y p(z_y|y_i)\ln(q(y_i|z_y)).
\end{equation}

\subsection{Variational Bounds on decoder terms not on a leaf - MINE}
\label{sec:MINE}
The variational bound above cannot be applied to the information terms that do not contain leaves in $G_{\rm decoder}$. For SIB, this corresponds to the $I^{D}(Z_X,Z_Y)$ term. This information is maximized. To find a variational bound such that $I^{D}(Z_X,Z_Y)\ge\Tilde{I}^{D}(Z_X,Z_Y)$, we use a simple approximation, the Mutal Information Neural Estimator (MINE) \citep{Hjelm2018}, which samples both $Z_X$ and $Z_Y$ from their respective variational encoders. Other mutual information estimators, such as $I_{\rm Info NCE}$ \citep{oord2018representation}, and $I_{\rm SMILE}$ \citep{song2019understanding} can be used as long as they are differentiable. Different estimators might be better suited for different problems, and we explore some effects of the choice later (cf.~Sec.~\ref{results_noisy_cifar} \& Sec.~\ref{results_sota}). However, for simple applications and for developing intuition, $I_{\rm MINE}$ is sufficient. We variationally approximate $p(z_x,z_y)$ as $p(z_x)p(z_y)e^{T(z_x,z_y)}/\mathcal{Z}_{\text{norm}}$, where $\mathcal{Z}_{\text{norm}}=\int dz_x dz_y p(z_x)p(z_y)e^{T(z_x,z_y)}=\mathbb{E}_{z_x \sim p(z_x), z_y \sim p(z_y)}[e^{T(z_x,z_y)}]$ is the normalization factor\footnote{The term $\mathcal{Z}_{\text{norm}}$ can be challenging to implement when doing optimization over batches. For example, \cite{Hjelm2018} propose using an expected moving average over the batches. Further discussion can be found in \cite{Tucker2019}. Simple implementations for the MINE estimator ignore this issue, but newer implementations (cf.~\cite{song2019understanding,abdelaleem2025accurate}) provide better estimates, as we discuss later in Appx.~\ref{app:estimators}.}. Here $T(z_x,z_y)$, a critic function, is parameterized by a neural network that takes in samples of the latent spaces $z_x$ and $z_y$ and returns a single number. We again  use the positivity of the Kullback-Leibler divergence, $D_{\rm KL}(p(z_x,z_y) \Vert p(z_x)p(z_y)e^{T(z_x,z_y)}/\mathcal{Z}_{\text{norm}}) \ge 0$, which implies $\int dz_x dz_y p(z_x,z_y)\ln(p(z_x,z_y))\ge\int dz_x dz_y p(z_x,z_y)\ln\frac{p(z_x)p(z_y)e^{T(z_x,z_y)}}{\mathcal{Z}_{\text{norm}}}$. Subtracting $\int dz_x dz_y p(z_x,z_y)\ln(p(z_x)p(z_y))$ from both sides, we find:
\begin{equation}
I^{D}(Z_X;Z_Y) \ge
\int dz_x dz_y p(z_x,z_y)\ln\frac{e^{T(z_x,z_y)}}{\mathcal{Z}_{\text{norm}}}\equiv \Tilde{I}_{\text{MINE}}^{D}(Z_X;Z_Y).
\end{equation}

\subsection{Parameterizing the distributions and the reparameterization trick}
$H(X)$, $H(Y)$, and $I(X,Y)$ do not depend on $p(z_x|x)$ and $p(z_y|y)$ and are dropped from the loss. Further, we can use any ansatz for the variational distributions we introduced. We choose parametric probability distribution families and learn the nearest distribution in these families consistent with the data. We assume $p(z_x|x)$ is a normal distribution with mean $\mu_{Z_X}(x)$ and a diagonal variance $\Sigma_{Z_X}(x)$. We learn the mean and the log variance as neural networks. We also assume that $q(x|z_x)$ is normal with a mean $\mu_{X}(z_x)$ and a unit variance.\footnote{In principle, we could also learn the variance for this distribution, but practically we did not find the need for that. Also note that, because we assume a unit variance for the decoder, both probabilistic and deterministic decoders are functionally equivalent: one models an exact map $x(z_x)$, and the other uses the same neural network to represent the mean, $\mathbb{E}_x(x|z_x)$.}. Finally, we assume that $r(z_x)$ is a standard normal distribution.  We use the reparameterization trick to produce samples of ${z_x}_{i,j}=z_{x_j}(x_i)=\mu(x_i)+\sqrt{\Sigma_{Z_X}(x_i)} \eta_j$  from $p(z_x|x_i)$, where $\eta_j$ is drawn from a standard normal distribution \citep{Welling2014}. We choose the same types of distributions for the corresponding $z_y$ terms. 

To sample from $p(z_x,z_y)$ we use  $p(z_x,z_y)=\int dx dy\, p(z_x,z_y,x,y)=\int dx dy\, p(z_x|x)p(z_y|y)\times p(x,y)\approx\frac{1}{N}\sum_{i=1}^N p(z_x|x_i)p(z_y|y_i)=\frac{1}{N M^2}\sum_{i=1}^N (\sum_{j=1}^M\delta(z_x-{z_x}_{i,j}))(\sum_{j=1}^M\delta(z_y-{z_y}_{i,j}))$, where ${z_x}_{i,j} \in p(z_x|x_i)$ and ${z_y}_{i,j} \in p(z_y|y_i)$, and $M$ is the number of new samples being generated. To sample from $p(z_x)p(z_y)$, we generate samples from $p(z_x,z_y)$ and scramble the generated entries $z_x$ and $z_y$, destroying all correlations.
With this, the components of the loss function become

\begin{align}
\Tilde{I}^{E}(X;Z_X) &\approx \frac{1}{2N}\sum_{i=1}^N \left[\text{Tr}({\Sigma_{Z_X}(x_i)}) +||\vec{\mu}_{Z_X}(x_i)||^2-k_{Z_X}-\ln \det(\Sigma_{Z_X}(x_i)) \right],\label{IExzx}\\
\Tilde{I}^{D}(X;Z_X)&\approx
\frac{1}{MN}\sum_{i,j=1}^{N,M} {-\frac{1}{2}}||(x_i - \mu_{X}({z_x}_{i,j}))||^2,\label{IDxzx}\\
\Tilde{I}^{D}_{\rm MINE}(Z_X;Z_Y)&
\approx
\frac{1}{M^2N}\sum_{i,j_x,j_y=1}^{N,M,M}\left[T({z_x}_{i,j_x},{z_y}_{i,j_y}) - \ln \mathcal{Z}_{\text{norm}}\right],\label{Imine}
\end{align}
where $\mathcal{Z}_{\text{norm}}=\mathbb{E}_{z_x \sim p(z_x), z_y \sim p(z_y)}[e^{T(z_x,z_y)}]$, $k_{Z_X}$ is the dimension of $Z_X$, and the corresponding terms for $Y$ are similar. Combining these terms results in the variational loss for DVSIB:
\begin{equation}
\label{dvsib_eq}
L_{\text{DVSIB}}=\Tilde{I}^{E}(X;Z_X)+\Tilde{I}^{E}(Y;Z_Y)
-\beta \left(\Tilde{I}^{D}_{\rm MINE}(Z_X;Z_Y)+\Tilde{I}^{D}(X;Z_X)+\Tilde{I}^{D}(Y;Z_Y)\right).
\end{equation}

\section{Deriving other DR methods}
\label{different_dr_methods}
The variational bounds used in DVSIB can be used to implement loss functions that correspond to other encoder-decoder graph pairs and hence to other DR algorithms. 

\subsection{DVSIB variants}
One can build another variant of DVSIB, but without reconstruction: DVSIB-noRecon. This variant is obtained by modifying the decoder graph in Fig.~\ref{Fig:Graph}, $G_\text{decoder}$, by removing the arrows that go from the embeddings $Z_X,Z_Y$ to $X,Y$  which removes the two decoder terms $\Tilde{I}^{D}(X;Z_X), \Tilde{I}^{D}(Y;Z_Y)$ from the loss function (Eq.~\ref{dvsib_eq}). In addition, a generic decoder not-on-a-leaf can be used $\Tilde{I}^D(Z_X;Z_Y)$.
\begin{equation}
\label{dvsib_no_recon_eq}
    L_{\text{DVSIB-noRecon}}=\Tilde{I}^{E}(X;Z_X)+\Tilde{I}^{E}(Y;Z_Y) - \beta \Tilde{I}^{D}(Z_X;Z_Y).
\end{equation}

The deterministic variant of DVSIB-noRecon, DSIB-noRecon, is obtained by using deterministic encoders instead of probabilistic encoders, $p(z_x|x)=\delta(\vec z_x-\vec \mu_{Z_X}(x))$ and $p(z_y|y)=\delta(\vec z_y-\vec \mu_{Z_Y}(y))$, in addition to taking the limit of $\beta \rightarrow\infty$.
\begin{equation}
\label{dsib_no_recon_eq}
    L_{\text{DSIB-noRecon}}= - \Tilde{I}^{D}(Z_X;Z_Y).
\end{equation}  
The deterministic variant will be useful in deriving other methods (cf. Sec~\ref{results_sota}).

\begin{table}[h]
    \caption{Method descriptions, variational losses, and the Bayesian Network graphs for each DR method derived in our framework. See Appx.\ref{App:Library} for details.}
    \label{table:methods}
    \vspace{-.2in}

    \begin{center}
    \begin{tabular}
    {|m{0.68\textwidth}|>{\centering\arraybackslash}m{0.16\textwidth}|>{\centering\arraybackslash}m{0.16\textwidth}|}
    \hline
        \textbf{Method Description}
        & \textbf{$G_{\text{encoder}}$} & \textbf{$G_{\text{decoder}}$} \\
        \hline
    \textbf{beta-VAE} \citep{Welling2014, Lerchner2016}: Two independent Variational Autoencoder (VAE) models trained, one for each view, $X$ and $Y$ (only $X$ graphs/loss shown).\newline
        $\sloppy L_{\text{VAE}}= \Tilde{I}^{E}(X;Z_X)-\beta \Tilde{I}^{D}(X;Z_X)$
        &
        \centering
        \adjustbox{height=18mm}{
            \begin{tikzpicture}[node distance={15mm}, thick,
                main/.style = {draw, circle,minimum size=11mm},
                comp/.style = {draw, circle,minimum size=9mm},
                labs/.style = {}
            ] 
            \node[main] (1) {$X$}; 
            \node[comp] (2) [below of=1] {$Z_X$};
            \draw[->] (1) -- (2);
            
            \end{tikzpicture}
        }
        &
        \adjustbox{height=18mm}{
            \begin{tikzpicture}[node distance={15mm}, thick,
                main/.style = {draw, circle,minimum size=11mm},
                comp/.style = {draw, circle,minimum size=9mm},
                labs/.style = {}
            ]

            \node[main] (3) {$X$};
            \node[comp] (4) [below of=3] {$Z_X$};
            \draw[->] (4) -- (3);
            \end{tikzpicture}
        }
        \\
        \hline
        \textbf{DVIB} \citep{Murphy2017}: Two bottleneck models trained, one for each view, $X$ and $Y$, using the other view as the supervising signal. (Only $X$ graphs/loss shown).\newline
        $\sloppy L_{\text{DVIB}}=\Tilde{I}^{E}(X;Z_X)-\beta \Tilde{I}^{D}(Y;Z_X)$
        &
        \adjustbox{height=18mm}{
        \begin{tikzpicture}[node distance={15mm}, thick,
            main/.style = {draw, circle,minimum size=11mm},
            comp/.style = {draw, circle,minimum size=9mm},
            labs/.style = {}
            ] 
            \node[main] (x) {$X$}; 
            \node[main] (y) [right of=x] {$Y$};
            \node[comp] (zx) [below of=x] {$Z_X$}; 
            \draw (x) -- (y);
            \draw[->] (x) -- (zx);
            
            \end{tikzpicture}
        }
        &
        \adjustbox{height=18mm}{
            \begin{tikzpicture}[node distance={15mm}, thick,
                main/.style = {draw, circle,minimum size=11mm},
                comp/.style = {draw, circle,minimum size=9mm},
                labs/.style = {}
            ]
            
            \node[main] (mx) {$X$}; 
            \node[main] (my) [right of=mx] {$Y$};
            \node[comp] (mzx) [below of=mx] {$Z_X$}; 
            \draw[->] (mzx) -- (my);
            \end{tikzpicture}    
        } \\
        \hline
        \textbf{beta-DVCCA}: Similar to DVIB \citep{Murphy2017}, but with reconstruction of both views. Two models trained, compressing either $X$ or $Y$,  while reconstructing both $X$ and $Y$. (Only $X$ graphs/loss shown).
        \newline
        $\sloppy L_{\text{DVCCA}}=\Tilde{I}^{E}(X;Z_X)-\beta (\Tilde{I}^{D}(Y;Z_X)+\Tilde{I}^{D}(X;Z_X))$
        \newline
        \textbf{DVCCA} \citep{Livescu2016}: $\beta$-DVCCA with $\beta=1$.
        &
        \adjustbox{height=18mm}{
        \begin{tikzpicture}[node distance={15mm}, thick,
            main/.style = {draw, circle,minimum size=11mm},
            comp/.style = {draw, circle,minimum size=9mm},
            labs/.style = {}
            ] 
            \node[main] (x) {$X$}; 
            \node[main] (y) [right of=x] {$Y$};
            \node[comp] (zx) [below of=x] {$Z_X$}; 
            \draw (x) -- (y);
            \draw[->] (x) -- (zx);
                        
            \end{tikzpicture}
        }
        &
        \adjustbox{height=18mm}{
            \begin{tikzpicture}[node distance={15mm}, thick,
                main/.style = {draw, circle,minimum size=11mm},
                comp/.style = {draw, circle,minimum size=9mm},
                labs/.style = {}
            ] 
            
            \node[main] (mx) {$X$}; 
            \node[main] (my) [right of=mx] {$Y$};
            \node[comp] (mzx) [below of=mx] {$Z_X$}; 
            \draw[->] (mzx) -- (my);
            \draw[->] (mzx) -- (mx);
            \end{tikzpicture}    
        } \\
        \hline
        \textbf{beta-joint-DVCCA}: A single model trained using a concatenated variable $[X,Y]$, learning one latent representation $Z$. \newline
        $\sloppy L_{\text{jDVCCA}}=\Tilde{I}^{E}((X,Y);Z)-\beta (\Tilde{I}^{D}(Y;Z)+\Tilde{I}^{D}(X;Z))$
        \newline
        \textbf{joint-DVCCA} \citep{Livescu2016}: $\beta$-jDVCCA with $\beta=1$.
        &
        \adjustbox{height=12mm}{
                    \begin{tikzpicture}[node distance={15mm}, thick,
            main/.style = {draw, circle,minimum size=11mm},
            comp/.style = {draw, circle,minimum size=9mm},
            labs/.style = {}
            ] 
            \node[main] (1) {$X$};
            \node[comp] (3) [below right of=1] {$Z$};
            \node[main] (2) [above right of=3] {$Y$};
            \draw (1) -- (2);
            \draw[->] (1) -- (3);
            \draw[->] (2) -- (3);
                        
            \end{tikzpicture}
        }
        &
        \adjustbox{height=12mm}{
            \begin{tikzpicture}[node distance={15mm}, thick,
                main/.style = {draw, circle,minimum size=11mm},
                comp/.style = {draw, circle,minimum size=9mm},
                labs/.style = {}
            ]

            \node[main] (x) {$X$};
            \node[comp] (z) [below right of=x] {$Z$};
            \node[main] (y) [above right of=z] {$Y$};
            \draw[->] (z) -- (x);
            \draw[->] (z) -- (y);
            \end{tikzpicture}
        } \\
        \hline

        \textbf{DVSIB}: A symmetric model trained, producing $Z_X$ and $Z_Y$.\newline
        $\sloppy L_{\text{DVSIB}}=\Tilde{I}^{E}(X;Z_X)+\Tilde{I}^{E}(Y;Z_Y) $\newline $- \beta \left(\Tilde{I}^{D}_{\text{MINE}}(Z_X;Z_Y)+\Tilde{I}^{D}(X;Z_X)+\Tilde{I}^{D}(Y;Z_Y)\right)$\newline
    &
    \adjustbox{height=18mm}{
    \begin{tikzpicture}[node distance={15mm}, thick,
        main/.style = {draw, circle,minimum size=11mm},
        comp/.style = {draw, circle,minimum size=9mm},
        labs/.style = {}
        ] 
        \node[main] (x) {$X$}; 
        \node[main] (y) [right of=x] {$Y$};
        \node[comp] (zx) [below of=x] {$Z_X$}; 
        \node[comp] (zy) [below of=y] {$Z_Y$};
        \draw (x) -- (y);
        \draw[->] (x) -- (zx);
        \draw[->] (y) -- (zy);
    \end{tikzpicture}
    }
    &
    \adjustbox{height=18mm}{
    \begin{tikzpicture}[node distance={15mm}, thick,
        main/.style = {draw, circle,minimum size=11mm},
        comp/.style = {draw, circle,minimum size=9mm},
        labs/.style = {}
        ] 
        \node[main] (mx) {$X$}; 
        \node[main] (my) [right of=mx] {$Y$};
        \node[comp] (mzx) [below of=mx] {$Z_X$}; 
        \node[comp] (mzy) [below of=my] {$Z_Y$};
        \draw[->] (mzx) -- (mzy);
        \draw[->] (mzx) -- (mx);
        \draw[->] (mzy) -- (my);
    \end{tikzpicture}
    }\\
    \hline
    \end{tabular}
    \end{center}
    \vspace{-0.3in}
\end{table}

\subsection{Other common DR methods}
Aside from DVSIB and its variants, the simplest other common DR method to derive using the Framework is the beta variational auto-encoder \citep{Welling2014}. Here $G_{\rm encoder}$  consists of one term: $X$ compressed into $Z_X$. Similarly $G_{\rm decoder}$  consists of one term: $X$ decoded from $Z_X$ (see Table~\ref{table:methods}). Using this simple set of Bayesian networks, we find the variational loss:
\begin{equation}
L_{\beta-\text{VAE}}=\Tilde{I}^{E}(X;Z_X)-\beta\Tilde{I}^{D}(X;Z_X).
\label{beta-VAE}
\end{equation}

Both terms in Eq.~(\ref{beta-VAE}) are the same as Eqs.~(\ref{IExzx}, \ref{IDxzx}) and can be approximated and implemented by neural networks. Note that taking the deterministic limit of Eq.~(\ref{beta-VAE}) results in a traditional AE. 

If we have a supervising variable $Y$ that we wish to reconstruct instead of $X$, modifying $G_{\rm decoder}$ accordingly results in DVIB, the Deep Variational Information Bottleneck \citep{Murphy2017}.
\begin{equation}
L_{\text{DVIB}}=\Tilde{I}^{E}(X;Z_X)-\beta\Tilde{I}^{D}(Y;Z_X).
\label{dvib-eq}
\end{equation}

Similarly, we can re-derive the DVCCA family of losses \citep{Livescu2016}, Table~\ref{table:methods}.
Here $G_{\rm encoder}$  is $X$ compressed into $Z_X$. $G_{\rm decoder}$ reconstructs both $X$ and $Y$ from the same compressed latent space $Z_X$. In fact, our loss function is more general than the DVCCA loss and has an additional compression-reconstruction trade-off parameter $\beta$. We call this more general loss $\beta$-DVCCA, and the original  DVCCA emerges when $\beta=1$:
\begin{equation}
L_{\rm DVCCA}=\Tilde{I}^{E}(X;Z_X)-\beta (\Tilde{I}^{D}(Y;Z_X)+\Tilde{I}^{D}(X;Z_X)).
\end{equation}
Using the same library of terms as we found for DVSIB, Eqs.~(\ref{IExzx}, \ref{IDxzx}), we find:
\begin{multline}
L_{\text{DVCCA}}\approx\frac{1}{N}\sum_{i=1}^N D_{\rm KL}(p(z_x|x_i) \Vert r(z_x))\\
-\beta\left( \frac{1}{N}\sum_{i=1}^N\int dz_x p(z_x|x_i)\ln(q(y_i|z_x)) + \frac{1}{N}\sum_{i=1}^N\int dz_x p(z_x|x_i)\ln(q(x_i|z_x))\right).
\end{multline}
This is similar to the loss function of the deep variational CCA \citep{Livescu2016}, but now it has a trade-off parameter $\beta$. It trades off the compression into $Z_X$ against the reconstruction of $X$ and $Y$ from the compressed variable $Z_X$. This apparent small change of adding $\beta$ have a significant impact on the accuracy of DVCCA methods, as shown in the results (Sec.~\ref{results_noisy_mnist} \& Sec.~\ref{results_noisy_cifar}) 

Table~\ref{table:methods} shows how our framework reproduces and generalizes other traditional DR losses (see Appx.~\ref{App:Library}). Our framework naturally extends beyond two variables and more state-of-the-art methods as well (see Tbl.~\ref{table:methodsII}). It also extends to models with private variables (see Appx.~\ref{App:private}).

\section{Results}
\label{results_all}
\subsection{Noisy MNIST}
\label{results_noisy_mnist}
\begin{figure}[h]
\begin{center}
\includegraphics[width=.85\textwidth]{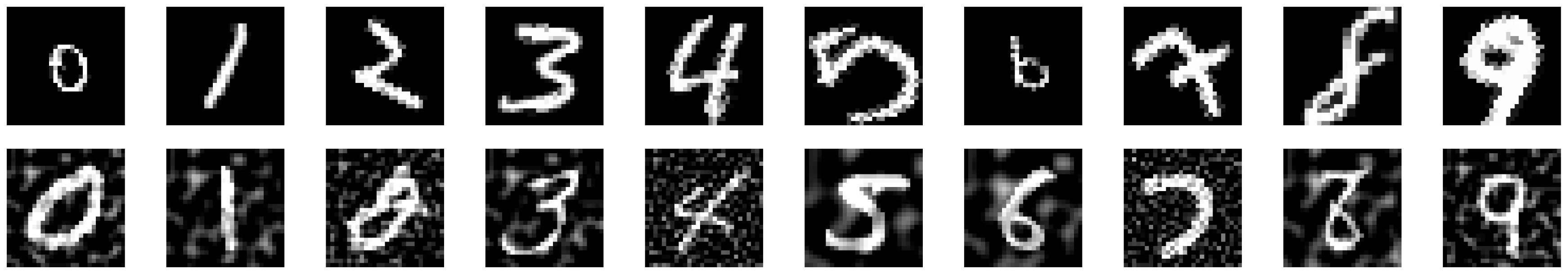}
\includegraphics[width=.32\textwidth]{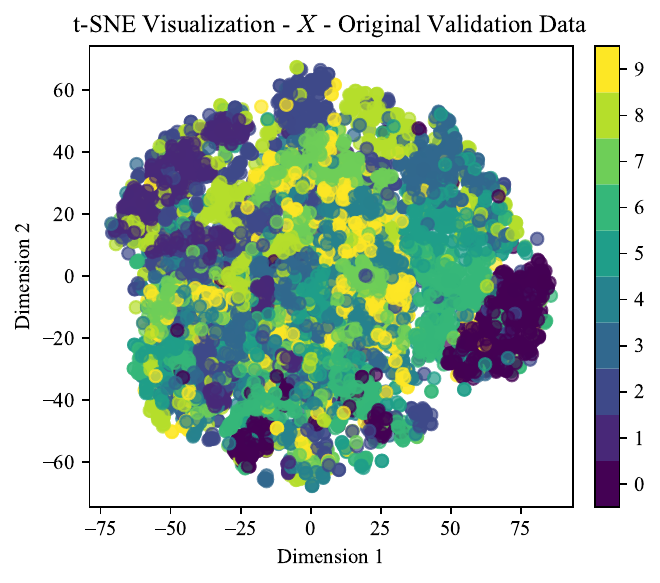}
\includegraphics[width=.32\textwidth]{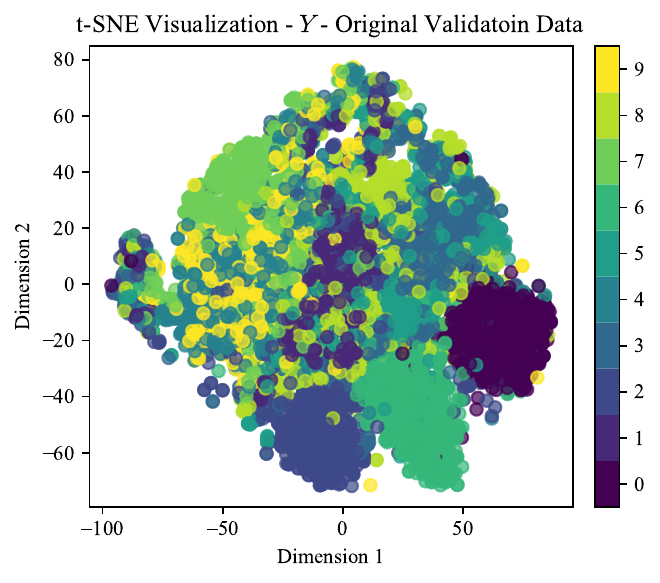}
\includegraphics[width=.32\textwidth]{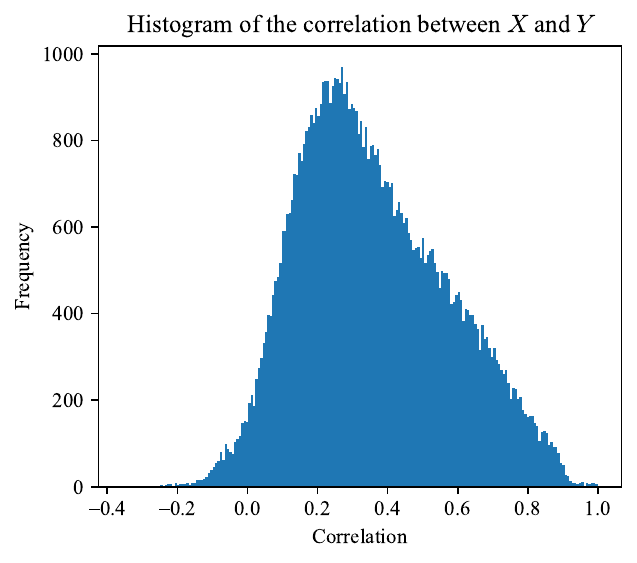}
\end{center}
\caption{Dataset consisting of pairs of digits drawn from MNIST that share an identity. Top row, $X$:  MNIST digits randomly scaled $(0.5-1.5)$ and rotated $(0-\pi/2)$. Bottom row, $Y$: MNIST digits with a background Perlin noise. t-SNE of $X$ and $Y$ datasets (left and middle) shows poor separation by digit, and there is a wide range of correlation between $X$ and $Y$ (right).}
\label{Fig:data}
\end{figure}

To test our methods, we created a dataset inspired by the Noisy MNIST dataset \citep{Haffner1998, Bilmes2015, Livescu2016}, consisting of two distinct views of data, both with dimensions of $28 \times 28$~pixels, cf.~Fig.~\ref{Fig:data}. The first view comprises the original image randomly rotated by an angle uniformly sampled between $0$ and $\frac{\pi}{2}$ and scaled by a factor uniformly distributed between $0.5$ and $1.5$. The second view consists of the original image with an added background Perlin noise \citep{Perlin1985} with the noise factor uniformly distributed between $0$ and $1$. Both image intensities are scaled to the range of $[0,1)$. The dataset was shuffled within labels, retaining only the shared label identity between two images, while disregarding the view-specific details, i.e., the random rotation and scaling for $X$, and the correlated background noise for $Y$. The dataset, totaling $70,000$ images, was partitioned into training ($80\%$), testing ($10\%$), and validation ($10\%$) subsets. Visualization via t-SNE \citep{Roweis2002} plots of the original dataset suggest poor separation by digit, and the two digit views have diverse correlations,  making this a sufficiently hard problem. 

The DR methods we evaluated include all methods from  Tbl.~\ref{table:methods}. PCA and CCA \citep{Hotelling1933, Hotelling1936} served as a baseline for linear dimensionality reduction. 
We emphasize that none of the algorithms were given labeled data. They had to infer compressed latent representations that presumably should cluster into ten different digits based simply on the fact that images come in pairs, and the (unknown) digit label is the only information that relates the two images.

Each method was trained for 100 epochs using fully connected neural networks with layer sizes $(\text{input\_dim}, 1024, 1024, (k_Z, k_Z))$, where $k_Z$ is the latent dimension size, employing ReLU activations for the hidden layers. The input dimension $(\text{input\_dim})$ was either the size of $X$ (784) or the size of the concatenated $[X, Y]$ (1568). The last two layers of size $k_Z$ represented the means and $\log(\text{variance})$ learned. For the decoders, we employed regular decoders, fully connected neural networks with layer sizes $(k_Z, 1024, 1024, \text{output\_dim})$, using ReLU activations for the hidden layers and sigmoid activation for the output layer. Again, the output dimension $(\text{output\_dim})$ could either be the size of $X$ (784) or the size of the concatenated $[X, Y]$ (1568). 
\begin{wraptable}{r}{0.68\textwidth} 
    \centering
    \vspace{-0.2in}
    \caption{Maximum accuracy from a linear SVM and the optimal $k_Z$ and $\beta$ for variational DR methods reported on the $Y$ (above the line) and the joint $[X,Y]$ (below the line) datasets. ($^\dag$ fixed values)}
    \label{table:SVM-Y}

    \begin{tabular}{|l|c|c|c|c|c|}
    \hline
    \textbf{Method} & \textbf{Acc. \%} & \textbf{${k_Z}_\textbf{best}$} & $95\%$ \textbf{${k_Z}$} & \bm{$\beta_\text{best}$} & $95\%$ \textbf{$\beta$} \\
    \hline
    Baseline & 90.8 & 784$^\dag$ & - & - & - \\
    PCA & 90.5 & 256 & [64,256*] & - & - \\
    CCA & 85.7 & 256 & [32,256*] & - & - \\
    $\beta$-VAE & 96.3 & 256 & [64,256*] & 32 & [2,1024*] \\
    DVIB & 90.4 & 256 & [16,256*] & 512 & [8,1024*] \\
    DVCCA & 89.6 & 128 & [16,256*] & 1$^\dag$ & - \\
    $\beta$-DVCCA & 95.4 & 256 & [64,256*] & 16 & [2,1024*] \\
    DVSIB & \textbf{97.8} & 256 & [\textbf{8},256*] & 128 & [2,1024*] \\
    \hline
    jBaseline & 91.9 & 1568$^\dag$ & - & - & - \\
    jDVCCA & 92.5 & 256 & [64,265*] & 1$^\dag$ & - \\
    $\beta$-jDVCCA & 96.7 & 256 & [16,265*] & 256 & [1,1024*] \\
    \hline
    \end{tabular}
\end{wraptable}

The latent dimension $(k_{Z})$ could be $k_{Z_X}$ or $k_{Z_Y}$ for regular decoders, or $k_{Z_X}+k_{W_X}$ or $k_{Z_Y}+k_{W_Y}$ for decoders with private information. Additionally, another decoder based on the MINE estimator for estimating $I(Z_X, Z_Y)$, was used in DVSIB and DVSIB with private information. The critic of $I_{\text{MINE}}(Z_X, Z_Y)$ is a fully connected neural network with layer sizes $(k_{Z_X}+k_{Z_Y}, 1024, 1024, 1)$ and ReLU activations for the hidden layers. Optimization was conducted using the ADAM optimizer with default parameters.

\begin{figure}[t]
\begin{center}
\includegraphics[width=.48\textwidth]{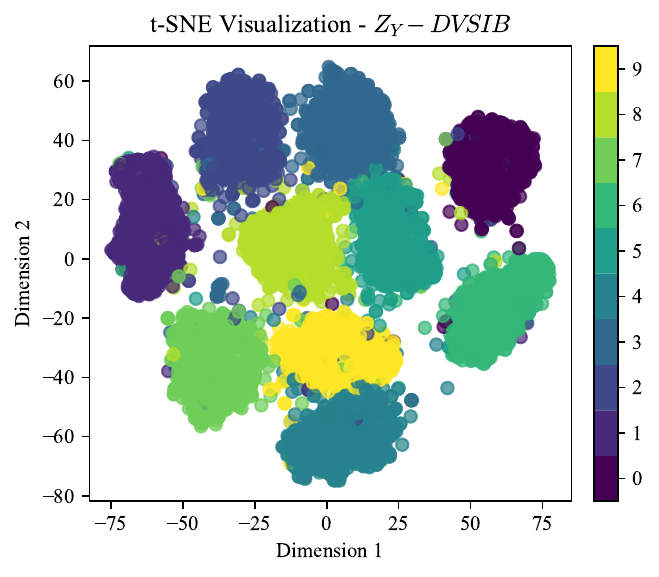}
\includegraphics[width=.5\textwidth]{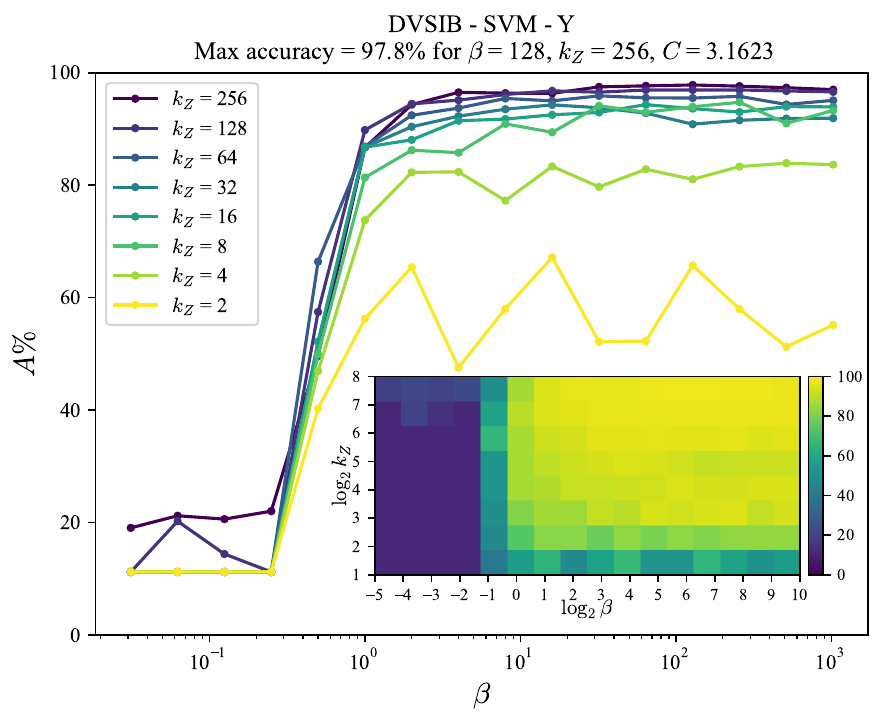}
\includegraphics[width=.85\textwidth]{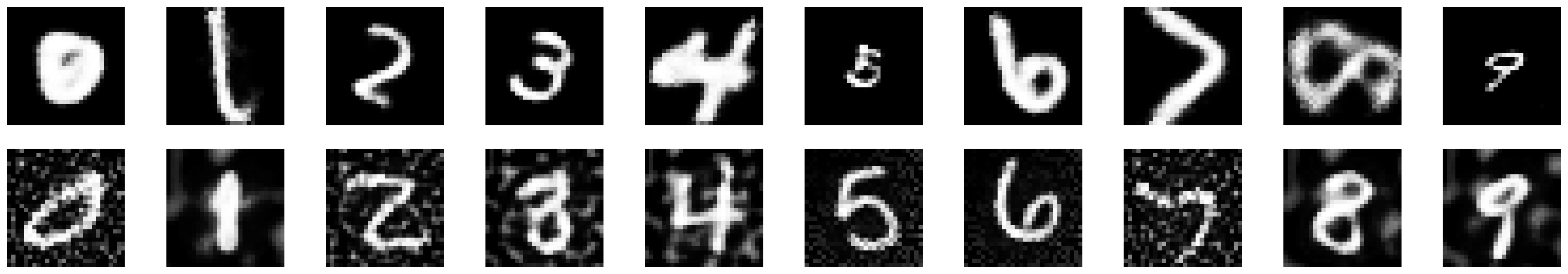}
\end{center}
\vspace{-.15in}
\caption{Top: t-SNE plot of the latent space $Z_Y$ of DVSIB colored by the identity of digits. Top Right: Classification accuracy of an SVM trained on DVSIB's $Z_Y$ latent space. The accuracy was evaluated for DVSIB with a parameter sweep of the trade-off parameter $\beta=2^{-5},...,2^{10}$ and the latent dimension $k_Z=2^1, ..., 2^8$. The max accuracy was  $97.8\%$ for $\beta=128$ and $k_Z=256$. Bottom: Example digits generated by sampling from the DVSIB decoder, $X$ and $Y$ branches.}
\vspace{-.2in}
\label{Fig:DVSIB}
\end{figure}

To evaluate the methods, we trained them on the training portions of $X$ and $Y$ without exposure to the true labels. Subsequently, we used the trained encoders to compute $Z_{\text{train}}$, $Z_{\text{test}}$, and $Z_{\text{validation}}$ on the respective datasets. To assess the quality of the learned representations, we revealed the labels of $Z_{\text{train}}$ and trained a linear SVM classifier with $Z_{\text{train}}$ and  $\text{labels}_{\text{train}}$. Hyper-parameter tuning of the classifier was performed to identify the optimal SVM slack parameter ($C$ value), maximizing accuracy on $Z_{\text{validation}}$. This best classifier was then used to predict $Z_{\text{test}}$, yielding the reported accuracy. We also conducted classification experiments using fully connected neural networks, with detailed results available in Appx.~\ref{App:results-mnist}. For both SVM and the fully connected network, we find the baseline accuracy on the original training data and labels  $(X_\text{train}, \text{labels}_\text{train})$ and $(Y_\text{train}, \text{labels}_\text{train})$, choose $C$ based on the validation datasets, and report the results of the test datasets.

Using a linear SVM allows us to assess the linear separability of the clusters of $Z_X$ and $Z_Y$ obtained through DR methods. While neural networks can uncover complex nonlinear relationships that may lead to higher classification accuracy, a linear SVM ensures that classification performance reflects the quality of the embeddings themselves rather than the classifier’s ability to compensate for any shortcomings in the DR methods. Additionally, if the embeddings are already linearly separable, they are likely to generalize more easily. Notably, we have also evaluated classification performance using neural networks, and the results (cf.~Appx.~\ref{App:results-mnist}) show that DVSIB consistently achieves the highest or among the highest classification accuracies across both linear and nonlinear classifiers, further supporting the effectiveness of its embeddings.

Here, we focus on the results of the $Y$ datasets (MNIST with correlated noise background); results for $X$ are in Appx.~\ref{App:results-mnist}. A parameter sweep was performed to identify optimal $k_Z$ values, ranging from $2^1$ to $2^8$ dimensions on $\log_2$ scale, as well as optimal $\beta$ values, ranging from $2^{-5}$ to $2^{10}$. 
For methods with private variables, $k_{W_X}$ and $k_{W_Y}$ were varied from $2^1$ to $2^6$ (results for these methods are available in Appx.~\ref{App:private}). The highest accuracy is reported in Tbl.~\ref{table:SVM-Y}, along with the optimal parameters used to obtain this accuracy. 

\begin{wrapfigure}{r}{0.55\textwidth} 
    \centering
    \vspace{-0.2in}
    \includegraphics[width=0.53\textwidth]{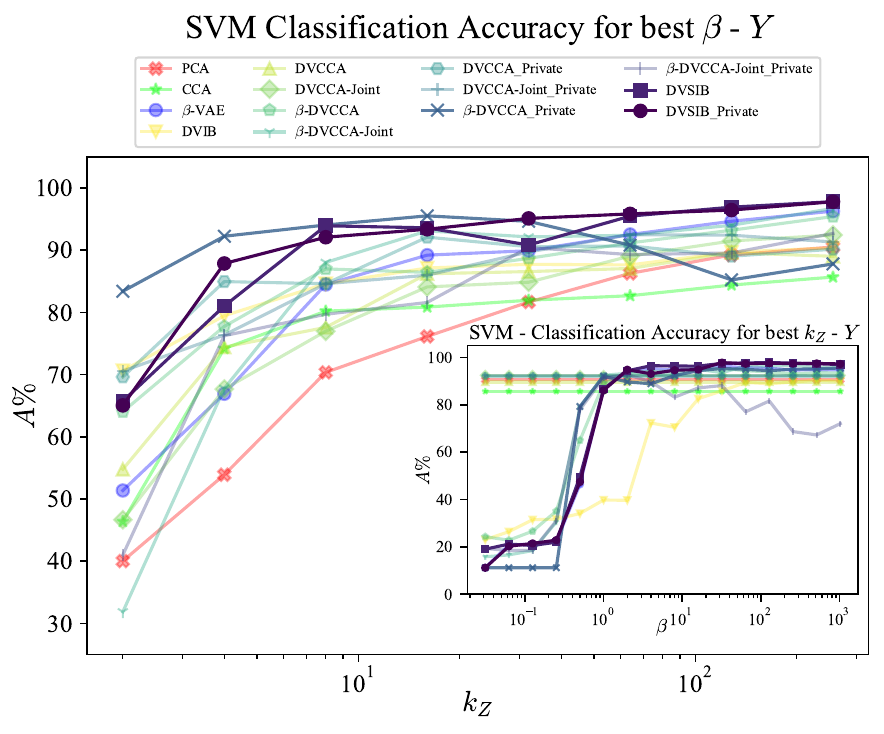}
    \vspace{-0.1in}
    \caption{The best SVM classification accuracy curves for each method. Here, DVSIB and DVSIB-private achieved the highest accuracy, and, together with $\beta$-DVCCA-private, they performed best for low-dimensional latent spaces.}
    \vspace{-0.2in}
    \label{Fig:SVM-Y}
\end{wrapfigure}
Additionally, for every method we find the range of $\beta$ and the dimensionality $k_Z$ of the latent variable $Z_Y$ that gives 95\% of the method's maximum accuracy. If the range includes the limits of the parameter, this is indicated by an asterisk.

Figure~\ref{Fig:DVSIB} shows a t-SNE plot of DVSIB's latent space, $Z_Y$, colored by the identity of digits. The resulting latent space has 10 clusters, each corresponding to one digit. The clusters are well separated and interpretable. Further, DVSIB's $Z_Y$ latent space provides the best classification of digits using a linear method such as an SVM showing the latent space is linearly separable.

DVSIB maximum classification accuracy obtained for the linear SVM is 97.8\%. Crucially, DVSIB maintains  accuracy of at least 92.9\% (95\% of 97.8\%) for  $\beta\in [2,1024^*]$ and $k_Z\in  [8,256^*]$. This accuracy is high compared to other methods and has a large range of parameters that maintain its ability to correctly capture information about the identity of the shared digit. Since DVSIB is a generative method, we also provided sample generated digits from the decoders that were trained from the model graph.

In Fig.~\ref{Fig:SVM-Y}, we show the highest SVM classification accuracy curves for each method. DVSIB and DVSIB-private tie for the best classification accuracy for $Y$. Together with $\beta$-DVCCA-private they have the highest accuracy for all dimensions of the latent space, $k_Z$. In theory, only one dimension should be needed to capture the identity of a digit, but our data sets also contain information about the rotation and scale for $X$ and the strength of the background noise for $Y$. $Y$ should then need at least two latent dimensions to be reconstructed and $X$ should need at least three. Since DVSIB, DVSIB-private, and $\beta$-DVCCA-private performed with the best accuracy starting with the smallest $k_Z$, we conclude that methods with the encoder-decoder graphs that more closely match the structure of the data produce higher accuracy with lower dimensional latent spaces. 

\begin{wrapfigure}{l}{0.58\textwidth} 
    \centering
    \includegraphics[width=0.56\textwidth]{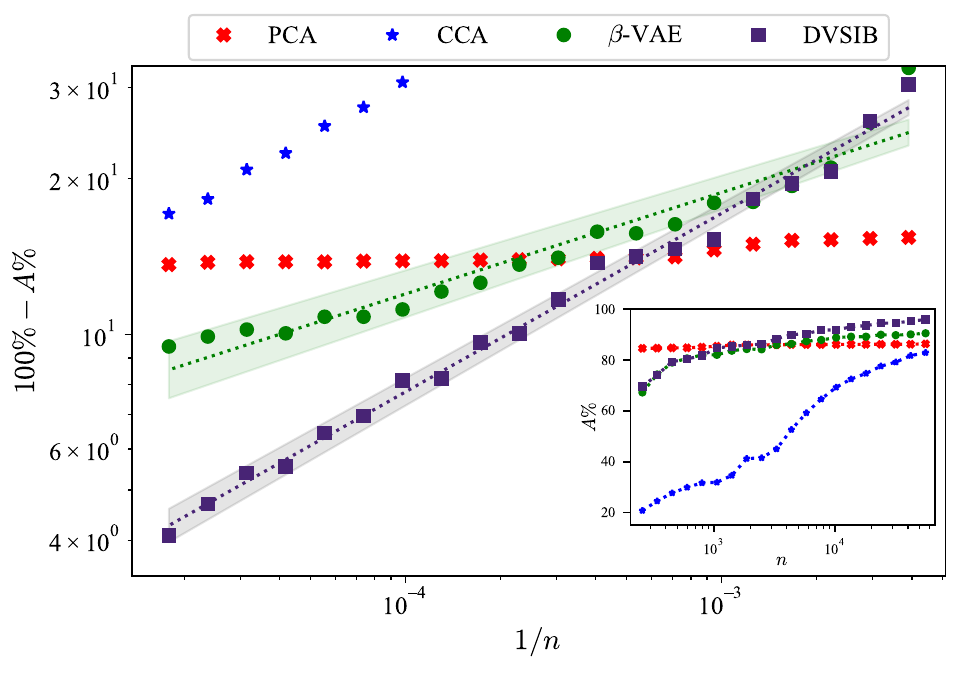}
    \vspace{-0.1in}
    \caption{Classification accuracy ($A$) of DVSIB scales better with sample size ($n$). Main: a log-log plot of $100\%-A$ vs. $1/n$. The slope for fitted lines is $0.345\pm0.007$ for DVSIB and $0.196\pm0.013$ for $\beta$-VAE, indicating a faster increase in accuracy of DVSIB with $n$. Inset: same data plotted as $A$ vs. $n$.}
    \label{Fig:Acc_T_SVM-Y}
    \vspace{-0.2in}
\end{wrapfigure}

Next, in Fig.~\ref{Fig:Acc_T_SVM-Y}, we compare the sample training efficiency of DVSIB and $\beta$-VAE by training new instances of these methods on a geometrically increasing number of samples $n=[256, 339, 451,\dots,\sim42k, \sim56k]$, consisting of 20 subsamples of the full training data ($X,Y$) to get ($X_{\text{train}_n},Y_{\text{train}_n}$), where each larger subsample includes the previous one. 

Each method was trained for 60 epochs, and we used $\beta=1024$ (as defined by the DVMIB framework). Further, all reported results are with the latent space size $k_Z=64$. We explored other numbers of training epochs and latent space dimensions (see Appx.~\ref{App:subsamples}), but did not observe qualitative differences. We follow the same procedure as outlined earlier, using the 20 trained encoders for each method to compute $Z_{\text{train}_{n}}$, $Z_{\text{test}}$, and $Z_{\text{validation}}$ for the training, test, and validation datasets. As before, we then train and evaluate the classification accuracy of SVMs for the $Z_Y$ representation learned by each method. Fig.~\ref{Fig:Acc_T_SVM-Y}, inset, shows the classification accuracy of each method as a function of the number of samples used in training. Again, CCA and PCA serve as linear methods baselines. PCA is able to capture the linear correlations in the dataset consistently, even at low sample sizes. However,  it is unable to capture the nonlinearities of the data, and its accuracy does not improve with the sample size. Because of the iterative nature of the implementation of the PCA algorithm \citep{scikit-learn}, it is able to capture some linear correlations in a relatively low number of dimensions, which are sufficiently sampled even with small-sized datasets. Thus the accuracy of PCA barely depends on the training set size. CCA, on the other hand, does not work in the under-sampled regime (see \cite{abdelaleem2024simultaneous} for discussion of this). DVSIB performs uniformly better, at all training set sizes, than the $\beta$-VAE. Furthermore, DVSIB improves its quality faster, with a different sample size scaling. Specifically,  DVSIB and $\beta$-VAE accuracy ($A$, measured in percent) appears to follow the scaling form $A=100-c/n^m$, where $c$ is a constant, and the scaling exponent $m=0.345\pm0.007$ for DVSIB, and $0.196\pm0.013$ for $\beta$-VAE. We illustrate this scaling in Fig.~\ref{Fig:Acc_T_SVM-Y} by plotting a log-log plot of $100-A$ vs $1/n$ and observing a linear relationship. 


\subsection{More complex tests: Noisy CIFAR-100 and CNN architecture}
\label{results_noisy_cifar}

Next, we extend our analysis to more complex datasets and architectures to verify whether the trends observed in Noisy MNIST results hold. Similar to Noisy MNIST, we construct a Noisy CIFAR-100 dataset, which consists of $60,000$ RGB images of size $32 \times 32$, totaling $3072$ pixels per image, with $100$ classes and $600$ images per class. We generate two distinct views in the same manner as Noisy MNIST: the first view ($X$) is randomly rotated by an angle uniformly sampled between $0$ and $\frac{\pi}{2}$ and scaled by a factor uniformly drawn from $[0.5, 1.5]$. The second view ($Y$) incorporates background Perlin noise \citep{Perlin1985}, where the noise factor for each RGB channel is independently sampled from a uniform distribution over $[0,1]$. Both views are rescaled to have pixel intensities in the range $[0,1)$, and the dataset is shuffled within labels while preserving the shared label identity between the two transformed views ($X$ and $Y$).

On the architecture side, instead of feedforward networks, we use a \textit{simple} convolutional neural network (CNN) for the encoders and decoders of the different DR methods implemented. We observe classification accuracy trends similar to those in Noisy MNIST, where conv-DVSIB (and, to a great extent, conv-$\beta$-DVCCA) outperforms other methods with comparable architectures and computational costs. See Appx.~\ref{app:cifar} for details.

\subsection{Towards the state of the art methods}
\label{results_sota}

Having demonstrated that various common DR methods, Table~\ref{table:methods}, here we extend the approach to more recent, state-of-the-art methods. The fundamental trade-off between an encoder and a decoder graph can encapsulate widely used two-view learning approaches such as Barlow Twins \cite{zbontar2021}, CLIP \cite{radford2021learning}, and several other methods \cite{bardes2024revisiting, assran2023self}. Similarly, this framework can accommodate multi-view approaches, including Deep Multi-View Information Bottleneck methods \cite{VanderSchaar2021, Hu2021, Elgamal2022}, among others. We show that, by making simple assumptions (such as treating embeddings as deterministic and jointly Gaussian in the case of Barlow Twins), or by constructing the appropriate computational graphs, as in the Multi-View IB, these methods naturally fit within the DVMIB framework.

Table~\ref{table:methodsII} provides an overview of these extended methods, with additional details in Appx.~\ref{more_methods_details}. Moreover, we demonstrate that DVSIB, when adapted to match the structure and complexity of methods such as Barlow Twins and trained under comparable conditions, achieves similar or even superior accuracy. For details, see Appx.~\ref{app:results-sota}.

\begin{table}
    \vspace{-.5in}
    \caption{Method descriptions, variational losses, and the Bayesian Network graphs for more DR methods derived within the DVMIB framework.}
    \label{table:methodsII}
    \vspace{-.2in}

    \begin{center}
    \begin{tabular}
    {|m{0.7\textwidth}|>{\centering\arraybackslash}m{0.16\textwidth}|>{\centering\arraybackslash}m{0.16\textwidth}|}
    \hline
        \textbf{Method Description}
        & \textbf{$G_{\text{encoder}}$} & \textbf{$G_{\text{decoder}}$} \\
        \hline
        \textbf{Unsupervised Multi-View IB (UMVIB)}: Methods like \citep{Hu2021,Ye2020}, where one compresses multiple views/modalities $X_i$ into one (or more) latent variable $Z$ (or $Z_i$) and reconstruct $X_i$ from $Z$ (or $Z$ and $Z_i$). 
        \newline
        $\sloppy L_{\text{UMVIB}_1}= \Tilde{I}^{E}(X_1,X_2,X_3;Z)-\beta (\Tilde{I}^{D}(X_1;Z)+\Tilde{I}^{D}(X_2;Z)+\Tilde{I}^{D}(X_3;Z))$
        &
        \adjustbox{width=25mm}{
            \begin{tikzpicture}[node distance={15mm}, thick,
            main/.style = {draw, circle,minimum size=11mm},
            comp/.style = {draw, circle,minimum size=9mm},
            labs/.style = {}
            ] 
            \node[main] (x1)  {$X_1$}; 
            \node[main] (x2) [right of=x1] {$X_2$};
            \node[main] (x3) [right of=x2] {$X_3$}; 
            \node[comp] (z) [below of=x2] {$Z$};
            
            \draw[->] (x1) -- (z);
            \draw[->] (x2) -- (z);
            \draw[->] (x3) -- (z);            
            \end{tikzpicture}
        }
        
        &
        \adjustbox{width=25mm}{
            \begin{tikzpicture}[node distance={15mm}, thick,
            main/.style = {draw, circle,minimum size=11mm},
            comp/.style = {draw, circle,minimum size=9mm},
            labs/.style = {}
            ] 
            
            \node[main] (mx1) {$X_1$}; 
            \node[main] (mx2) [right of=mx1] {$X_2$};
            \node[main] (mx3) [right of=mx2] {$X_3$}; 
            \node[comp] (mz) [below of=mx2] {$Z$};
            
            \draw[->] (mz) -- (mx1);
            \draw[->] (mz) -- (mx2);
            \draw[->] (mz) -- (mx3);
            
            \end{tikzpicture}
        }
        \\
        $\sloppy L_{\text{UMVIB}_2}=\Tilde{I}^{E}((X_1,X_2,X_3);Z)+\Tilde{I}^{E}(X_1;Z_1)+\Tilde{I}^{E}(X_2;Z_2)+\Tilde{I}^{E}(X_3;Z_3)-\beta (\Tilde{I}^{D}(X_1;(Z,Z_1))+\Tilde{I}^{D}(X_2;(Z,Z_2))+\Tilde{I}^{D}(X_3;(Z,Z_3)))$
        &
        \adjustbox{width=25mm}{
            \begin{tikzpicture}[node distance={15mm}, thick,
            main/.style = {draw, circle,minimum size=11mm},
            comp/.style = {draw, circle,minimum size=9mm},
            labs/.style = {}
            ] 
            \node[main] (x1) {$X_1$}; 
            \node[main] (x2) [right of=x1] {$X_2$};
            \node[main] (x3) [right of=x2] {$X_3$}; 
            \node[comp] (z1) [below of=x1] {$Z_1$};
            \node[comp] (z2) [below of=x2] {$Z_2$};
            \node[comp] (z3) [below of=x3] {$Z_3$};
            \node[comp] (z) [below left of=x1] {$Z$};

            \draw[->] (x1) -- (z1);
            \draw[->] (x2) -- (z2);
            \draw[->] (x3) -- (z3);
            \draw[->] (x1) -- (z);
            \draw[->] (x2) -- (z);
            \draw[->] (x3) -- (z);
            \end{tikzpicture}
        }
&
        \adjustbox{width=25mm}{
            \begin{tikzpicture}[node distance={15mm}, thick,
            main/.style = {draw, circle,minimum size=11mm},
            comp/.style = {draw, circle,minimum size=9mm},
            labs/.style = {}
            ] 
            \node[main] (mx1) {$X_1$}; 
            \node[main] (mx2) [right of=mx1] {$X_2$};
            \node[main] (mx3) [right of=mx2] {$X_3$}; 
            \node[comp] (mz1) [below of=mx1] {$Z_1$};
            \node[comp] (mz2) [below of=mx2] {$Z_2$};
            \node[comp] (mz3) [below of=mx3] {$Z_3$};
            \node[comp] (mz) [below left of=mx1] {$Z$};
            
            \draw[->] (mz) -- (mx1);
            \draw[->] (mz) -- (mx2);
            \draw[->] (mz) -- (mx3);
            \draw[->] (mz1) -- (mx1);
            \draw[->] (mz2) -- (mx2);
            \draw[->] (mz3) -- (mx3);
            \end{tikzpicture}

        }
        \\
        \hline

        \textbf{Supervised Multi-View IB (SMVIB)}: Methods like \citep{VanderSchaar2021,Zhou2019,Elgamal2022,Meiser2022}, where one compresses multiple views/modalities $X_i$ into a shared latent variable $Z$ (or $Z$ and $Z_i$), and uses $Z$ to predict a supervisory signal/label $Y$ (or in addition to using both $Z$ and $Z_i$ to reconstruct $X_i$ as well).
        \newline
        $\sloppy L_{\text{SVMIB}_1}= \Tilde{I}^{E}(X_1,X_2,X_3;Z)-\beta (\Tilde{I}^{D}(Y;Z))$
        &
        \adjustbox{width=25mm}{
            \begin{tikzpicture}[node distance={15mm}, thick,
            main/.style = {draw, circle,minimum size=11mm},
            comp/.style = {draw, circle,minimum size=9mm},
            labs/.style = {}
            ] 
            \node[main] (x1)  {$X_1$}; 
            \node[main] (x2) [right of=x1] {$X_2$};
            \node[main] (x3) [right of=x2] {$X_3$};
            \node[main] (y) [right of=x3] {$Y$};
            \node[comp] (z) [below of=x3] {$Z$};
            
            \draw[->] (x1) -- (z);
            \draw[->] (x2) -- (z);
            \draw[->] (x3) -- (z);            
            \end{tikzpicture}
        }
        
        &
        \adjustbox{width=25mm}{
            \begin{tikzpicture}[node distance={15mm}, thick,
            main/.style = {draw, circle,minimum size=11mm},
            comp/.style = {draw, circle,minimum size=9mm},
            labs/.style = {}
            ] 
            
            \node[main] (mx1) {$X_1$}; 
            \node[main] (mx2) [right of=mx1] {$X_2$};
            \node[main] (mx3) [right of=mx2] {$X_3$};
            \node[main] (my) [right of=mx3] {$Y$};
            \node[comp] (mz) [below of=mx3] {$Z$};
            
            \draw[->] (mz) -- (my);
            
            \end{tikzpicture}
        }
        \\
        $\sloppy L_{\text{SMVIB}_2}=\Tilde{I}^{E}((X_1,X_2,X_3);Z)+\Tilde{I}^{E}(X_1;Z_1)+\Tilde{I}^{E}(X_2;Z_2)+\Tilde{I}^{E}(X_3;Z_3)-\beta (\Tilde{I}^{D}(X_1;(Z,Z_1))+\Tilde{I}^{D}(X_2;(Z,Z_2))+\Tilde{I}^{D}(X_3;(Z,Z_3))+\Tilde{I}^{D}(Z;Y))$
        &
        \adjustbox{width=25mm}{
            \begin{tikzpicture}[node distance={15mm}, thick,
            main/.style = {draw, circle,minimum size=11mm},
            comp/.style = {draw, circle,minimum size=9mm},
            labs/.style = {}
            ] 
            \node[main] (x1) {$X_1$}; 
            \node[main] (x2) [right of=x1] {$X_2$};
            \node[main] (x3) [right of=x2] {$X_3$}; 
            \node[comp] (z1) [below of=x1] {$Z_1$};
            \node[comp] (z2) [below of=x2] {$Z_2$};
            \node[comp] (z3) [below of=x3] {$Z_3$};
            \node[comp] (z) [below of=y] {$Z$};
            \node[main] (y) [right of=x3] {$Y$};

            \draw[->] (x1) -- (z1);
            \draw[->] (x2) -- (z2);
            \draw[->] (x3) -- (z3);
            \draw[->] (x1) -- (z);
            \draw[->] (x2) -- (z);
            \draw[->] (x3) -- (z);

            \end{tikzpicture}
        }
&
        \adjustbox{width=25mm}{
            \begin{tikzpicture}[node distance={15mm}, thick,
            main/.style = {draw, circle,minimum size=11mm},
            comp/.style = {draw, circle,minimum size=9mm},
            labs/.style = {}
            ] 
            \node[main] (mx1) {$X_1$}; 
            \node[main] (mx2) [right of=mx1] {$X_2$};
            \node[main] (mx3) [right of=mx2] {$X_3$}; 
            \node[comp] (mz1) [below of=mx1] {$Z_1$};
            \node[comp] (mz2) [below of=mx2] {$Z_2$};
            \node[comp] (mz3) [below of=mx3] {$Z_3$};
            \node[comp] (mz) [below of=y] {$Z$};
            \node[main] (my) [right of=x3] {$Y$};
            
            \draw[->] (mz) -- (my);
            \draw[->] (mz) -- (mx1);
            \draw[->] (mz) -- (mx2);
            \draw[->] (mz) -- (mx3);
            \draw[->] (mz1) -- (mx1);
            \draw[->] (mz2) -- (mx2);
            \draw[->] (mz3) -- (mx3);
            \end{tikzpicture}

        }
        \\
        \hline

        \textbf{Barlow Twins (BT)/DSIB} \citep{zbontar2021}: Two different views $X$ and $Y$ are compressed using the same deterministic compression function $\mu(\cdot)$ into $Z_X$ and $Z_Y$ respectively ($p(z_x|x)=\delta(z_x-\mu(x))$ and $p(z_y|y)=\delta(z_y-\mu(y))$). One then optimizes the cross-correlation matrix, $C$, between $Z_X$ and $Z_Y$ to be as close to identity as possible, which is equivalent to maximizing the information between $I(Z_X,Z_Y)$, when $Z_X$ and $Z_Y$ are jointly Gaussian.
        \newline
        $\sloppy L_{\text{DSIB-noRecon}}=-\Tilde{I}^{D}(Z_X;Z_Y)=\frac{1}{2}\ln(\det(I-CC^T))$
        \newline
        Same minimum as: $\sloppy L_{\text{BT}}=\sum_i (1-C_{ii})^2 + \lambda \sum_{i,j}C_{ij}^2$
        
        &
        \adjustbox{width=25mm}{
        \begin{tikzpicture}[node distance={15mm}, thick,
            main/.style = {draw, circle,minimum size=11mm},
            comp/.style = {draw, circle,minimum size=9mm},
            labs/.style = {}
            ] 
            \node[main] (x) {$X$}; 
            \node[main] (y) [right of=x] {$Y$};
            \node[comp] (zx) [below of=x] {$Z_X$};
            \node[comp] (zy) [below of=y] {$Z_Y$};
            \draw (x) -- (y);
            \draw[->] (x) -- (zx);
            \draw[->] (y) -- (zy);
                        
            \end{tikzpicture}
        }
        &
        \adjustbox{width=25mm}{
            \begin{tikzpicture}[node distance={15mm}, thick,
                main/.style = {draw, circle,minimum size=11mm},
                comp/.style = {draw, circle,minimum size=9mm},
                labs/.style = {}
            ] 
            \node[main] (x) {$X$}; 
            \node[main] (y) [right of=x] {$Y$};
            \node[comp] (zx) [below of=x] {$Z_X$};
            \node[comp] (zy) [below of=y] {$Z_Y$};
            \draw (zx) -- (zy);
            \end{tikzpicture}    
        }
        \\
        \hline
        \textbf{CLIP/DSIB} \citep{radford2021learning} : Two different views $X$ and $Y$ are compressed  using different deterministic encoder functions into $Z_X$ and $Z_Y$ respectively, $p(z_x|x)=\delta(\vec z_x-\vec \mu_{Z_X}(x))$ and $p(z_y|y)=\delta(\vec z_y-\vec \mu_{Z_Y}(y))$. The CLIP loss is equivalent to $-I(Z_X,Z_Y)$ (and hence to the no reconstruction version of DSIB) if $p(z_x|z_y)=p(z_y|z_x)$ or $-I(Z_X,Z_Y)+\mathrm{correction}$ otherwise.\newline 
        $\sloppy L_{\text{DSIB-noRecon}}= -\Tilde{I}^{D}(Z_X;Z_Y)$.\newline
        $\sloppy L_{\text{CLIP}}=\frac{-1}{2N}\sum_{i=1}^N\ln\left(\frac{\exp{(\vec{z}_{x_i}\cdot\vec{z}_{y_i}/T)}}{\frac{1}{N}\sum_{j=1}^N\exp{(\vec{z}_{x_i}\cdot\vec{z}_{y_j}/T)}}\right)$\newline
        $+\frac{-1}{2N}\sum_{i=1}^N\ln\left(\frac{\exp{(\vec{z}_{x_i}\cdot\vec{z}_{y_i}/T)}}{\frac{1}{N}\sum_{j=1}^N\exp{(\vec{z}_{x_j}\cdot\vec{z}_{y_i}/T)}}\right)$
        
        &
        \adjustbox{width=25mm}{
        \begin{tikzpicture}[node distance={15mm}, thick,
            main/.style = {draw, circle,minimum size=11mm},
            comp/.style = {draw, circle,minimum size=9mm},
            labs/.style = {}
            ] 
            \node[main] (x) {$X$}; 
            \node[main] (y) [right of=x] {$Y$};
            \node[comp] (zx) [below of=x] {$Z_X$};
            \node[comp] (zy) [below of=y] {$Z_Y$};
            \draw (x) -- (y);
            \draw[->] (x) -- (zx);
            \draw[->] (y) -- (zy);
                        
            \end{tikzpicture}
        }
        &
        \adjustbox{width=25mm}{
            \begin{tikzpicture}[node distance={15mm}, thick,
                main/.style = {draw, circle,minimum size=11mm},
                comp/.style = {draw, circle,minimum size=9mm},
                labs/.style = {}
            ] 
            \node[main] (x) {$X$}; 
            \node[main] (y) [right of=x] {$Y$};
            \node[comp] (zx) [below of=x] {$Z_X$};
            \node[comp] (zy) [below of=y] {$Z_Y$};
            \draw (zx) -- (zy);
            \end{tikzpicture}    
        }
        \\
        \hline
        \textbf{Possible Extensions}: Methods like \cite{bardes2024revisiting, assran2023self,caron2021emerging, chen2020big,zhou2022image,baevski2022data2vec,bardes2022vicreg,zhang2022contrastive,jia2021scaling, meng2022compressed}: probabilistically or deterministically compress $X$ and $Y$ into latent variables $Z_X$ and $Z_Y$ respectively and then learn a specific mapping between $Z_Y$ and $Z_X$.
        
        &
        \adjustbox{width=25mm}{
        \begin{tikzpicture}[node distance={15mm}, thick,
            main/.style = {draw, circle,minimum size=11mm},
            comp/.style = {draw, circle,minimum size=9mm},
            labs/.style = {}
            ] 
            \node[main] (x) {$X$}; 
            \node[main] (y) [right of=x] {$Y$};
            \node[comp] (zx) [below of=x] {$Z_X$};
            \node[comp] (zy) [below of=y] {$Z_Y$};
            \draw[->] (x) -- (y);
            \draw[->] (x) -- (zx);
            \draw[->] (y) -- (zy);
                        
            \end{tikzpicture}
        }
        &
        \adjustbox{width=25mm}{
            \begin{tikzpicture}[node distance={15mm}, thick,
                main/.style = {draw, circle,minimum size=11mm},
                comp/.style = {draw, circle,minimum size=9mm},
                labs/.style = {}
            ] 
            \node[main] (x) {$X$}; 
            \node[main] (y) [right of=x] {$Y$};
            \node[comp] (zx) [below of=x] {$Z_X$};
            \node[comp] (zy) [below of=y] {$Z_Y$};
            \draw[->] (zx) -- (zy);
            \end{tikzpicture}    
        }
        \\
        \hline        
    \end{tabular}
    \end{center}
\end{table}

\section{Conclusion}
We developed an MIB-based framework for deriving variational loss functions for DR applications. We demonstrated the use of this framework by developing a novel variational method, DVSIB. DVSIB compresses the variables $X$ and $Y$ into latent variables $Z_X$ and $Z_Y$ respectively, while maximizing the information between $Z_X$ and $Z_Y$.  The method generates two distinct latent spaces---a feature highly sought after in various applications---but it accomplishes this with superior data efficiency, compared to other methods. The example of DVSIB demonstrates the process of deriving variational bounds for terms present in all examined DR methods. A comprehensive library of typical terms is included in Appx.~\ref{App:Library} for reference, which can be used to derive additional DR methods. Further, we (re)-derive several DR methods, as outlined in Tables~\ref{table:methods}\&\ref{table:methodsII}. These include well-known techniques such as $\beta$-VAE, DVIB, DVCCA, and DVCCA-private, in addition to more complicated multiview techniques such as supervised and unsupervised MultiView IB, and state-of-the-art ones such as Barlow Twins and CLIP.
MIB naturally introduces a trade-off parameter into the DVCCA family of methods, resulting in what we term the $\beta$-DVCCA DR methods, of which DVCCA is a special case. We implement this new family of methods and show that it produces better latent spaces than DVCCA at $\beta=1$, cf.~Tbl.~\ref{table:SVM-Y} and Tbl.~\ref{tab:results_cifar_y}.

We observe that methods that more closely match the structure of dependencies in the data can give better latent spaces as measured by the dimensionality of the latent space and the accuracy of reconstruction (see Figure~\ref{Fig:SVM-Y}).  This makes DVSIB, DVSIB-private, and $\beta$-DVCCA-private perform the best. DVSIB and DVSIB-private both have separate latent spaces for $X$ and $Y$. The private methods allow us to learn additional aspects about $X$ and $Y$ that are not important for the shared digit label, but allow reconstruction of the rotation and scale for $X$ and the background noise of $Y$. We also found that DVSIB can make more efficient use of data when producing latent spaces as compared to $\beta$-VAEs and linear methods. 

Our framework extends beyond variational approaches and provides a unified foundation for diverse learning methods, including dimensionality reduction, supervised and unsupervised multi-view learning, and self-supervised approaches. In particular, we have explicitly demonstrated how the deterministic limit of DVSIB recovers DSIB, and we have shown that this limit naturally maps onto state-of-the-art methods such as Barlow Twins and CLIP (Table.~\ref{table:methodsII}). Additionally, we have implemented convolutional and ResNet-based encoder-decoder architectures (Sec.~\ref{results_noisy_cifar}, Sec.~\ref{results_sota}), illustrating that the framework is highly flexible: embedding functions can be easily swapped for more suitable architectures depending on the task. While we have not explicitly derived linear methods within our framework, prior work has shown that approaches like CCA can be viewed as special cases of the information bottleneck \citep{Weiss2003}, suggesting that they may also emerge naturally from our formulation. Overall, this versatility highlights the framework’s ability to unify broad classes of learning techniques. With the provided tools and code, we aim to facilitate the adaptation of the framework to a wide range of problems.

\acks{EA and KMM contributed equally to this work. We thank Sean Ridout and Michael Pasek for providing feedback on the manuscript, and Paarth Gulati for carefully verifying mathematical derivations in this work. EA and KMM also thank Ahmed Roman for useful and helpful discussions. EA is also grateful to Abdelrahman Nasser for his early help with coding when starting the project that eventually evolved into this work. IN is particularly grateful to the late Tali Tishby for many discussions about life, science, and information bottleneck over the years. This work was funded, in part, by NSF Grants Nos.\ 2010524 and 2014173, by the Simons Investigator award to IN, and the Simons-Emory International Consortium on Motor Control. We acknowledge support of our work through the use of the HyPER C3 cluster of Emory University's AI.Humanity Initiative.}
\clearpage

\appendix
\addcontentsline{toc}{section}{Appendix} 
\part{Appendix} 
\parttoc 

\section{Deriving and designing variational losses and their components}

\label{App:Library}
In the next two sections, we provide a library of typical terms found in encoder graphs, Appx.~\ref{App:Comp}, and decoder graphs, Appx.~\ref{App:Model}. In Appx.~\ref{App:Methods}, we provide examples of combining these terms to produce variational losses corresponding to beta-VAE, DVIB, beta-DVCCA, beta-DVCCA-joint, beta-DVCCA-private, DVSIB, and DVSIB-private.

\subsection{Encoder graph components}
\label{App:Comp}
We expand  Sec.~\ref{sec:encoder} and present a range of common components found in encoder graphs across various DR methods, cf.~Fig.~(\ref{Fig:Gencoder}).

\begin{figure}[h]
\begin{center}
\begin{tikzpicture}[node distance={15mm}, thick,
main/.style = {draw, circle,minimum size=11mm},
comp/.style = {draw, circle,minimum size=9mm},
labs/.style = {},
every edge quotes/.style = {auto, font=\footnotesize, sloped}
] 
\node[labs] (a) {a.};
\node[main] (1) [below right of=a] {$X$}; 
\node[comp] (2) [below of=1] {$Z_X$}; 
\draw[->] (1) -- (2);

\node[labs] (b) [above right of=1] {b.};
\node[main] (3) [below right of=b] {$X$}; 
\node[comp] (4) [below left of=3] {$W_X$};
\node[comp] (5) [below right of=3] {$Z_X$}; 
\draw[->] (3) -- (4);
\draw[->] (3) -- (5);

\node[labs] (c) [above right of=3] {c.};
\node[main] (6) [below right of=c] {$X$}; 
\node[comp] (8) [below right of=6] {$Z$};
\node[main] (7) [above right of=8] {$Y$};

\draw[->] (6) -- (8);
\draw[->] (7) -- (8);

\node[labs] (d) [above right of=7] {d.};
\node[main] (9) [below right of=d] {$X$};
\node[main] (10) [right of=9] {$Y$};

\draw (9) -- (10);
\end{tikzpicture}
\end{center}
\caption{Encoder graph components.}
\label{Fig:Gencoder}
\end{figure}

\begin{enumerate}[a.]
\item 
This graph corresponds to compressing the random variable $X$ to $Z_X$. Variational bounds for encoders of this type were derived in the main text in Sec.~\ref{sec:encoder} and correspond to the loss:
\begin{align}
\Tilde{I}^{E}(X;Z_X) &=\frac{1}{N}\sum_{i=1}^N D_{\rm KL}(p(z_x|x_i) \Vert r(z_x))\nonumber\\
&\approx \frac{1}{2N}\sum_{i=1}^N \left[\text{Tr}({\Sigma_{Z_X}(x_i)}) +||\vec{\mu}_{Z_X}(x_i)||^2-k_{Z_X}-\ln \det(\Sigma_{Z_X}(x_i)) \right].
\end{align}

\item
This type of encoder graph is similar to the first, but now with two outputs, $Z_X$ and $W_X$. This corresponds to making two encoders, one for $Z_X$ and one for $W_X$, $\Tilde{I}^{E}(Z_X;X)+\Tilde{I}^{E}(W_X;X)$, where
\begin{align}
\Tilde{I}^{E}(Z_X;X)&\approx\frac{1}{N}\sum_{i=1}^N D_{\rm KL}(p(z_x|x_i) \Vert r(z_x)),\\
\Tilde{I}^{E}(W_X;X)&\approx\frac{1}{N}\sum_{i=1}^N D_{\rm KL}(p(w_x|x_i) \Vert r(w_x)).
\end{align}

\item
This type of encoder consists of compressing $X$ and $Y$ into a single variable $Z$. It corresponds to the information loss $I^{E}(Z;(X,Y))$. This again has a similar encoder structure to type (a), but  $X$ is replaced by a joint variable $(X,Y)$. For this loss, we find a variational version:
\begin{equation}
\Tilde{I}^{E}(Z;(X,Y))\approx\frac{1}{N}\sum_{i=1}^N D_{\rm KL}(p(z|x_i,y_i) \Vert r(x_i,y_i)).
\end{equation}

\item
This final type of an encoder term corresponds to information $I^{E}(X,Y)$, which is constant with respect to our minimization. In practice, we drop terms of this type.

\end{enumerate}

\subsection{Decoder graph components}
\label{App:Model}
In this section, we elaborate on the decoder graphs that happen in our considered DR methods,  cf.~Fig.~(\ref{Fig:Gdecoder}).

\begin{figure}[h!]
\begin{center}
\begin{tikzpicture}[node distance={15mm}, thick,
main/.style = {draw, circle,minimum size=11mm},
comp/.style = {draw, circle,minimum size=9mm},
labs/.style = {}
] 
\node[labs] (a) {a.};
\node[main] (1) [below right of=a] {$X$}; 
\node[comp] (2) [below of=1] {$Z_X$}; 
\draw[->] (2) -- (1);

\node[labs] (b) [above right of=1] {b.};
\node[main] (3) [below right of=b] {$X$}; 
\node[comp] (4) [below left of=3] {$W_X$};
\node[comp] (5) [below right of=3] {$Z_X$}; 
\draw[->] (4) -- (3);
\draw[->] (5) -- (3);

\node[labs] (c) [above right of=3] {c.};
\node[main] (6) [below right of=c] {$X$}; 
\node[comp] (8) [below right of=6] {$Z$}; 
\node[main] (7) [above right of=8] {$Y$};

\draw[->] (8) -- (6);
\draw[->] (8) -- (7);
\end{tikzpicture}
\end{center}
\caption{Decoder graph components.}
\label{Fig:Gdecoder}
\end{figure}

All decoder graphs sample from their methods' corresponding encoder graph.
\begin{enumerate}[a.]
\item 
In this decoder graph, we decode $X$ from the compressed variable $Z_X$. 
Variational bounds for decoders of this type were derived in the main text, Sec.~\ref{sec:decoder}, and they correspond to the loss:
\begin{align}
\Tilde{I}^{D}(X;Z_X)&=H(X)+\frac{1}{N}\sum_{i=1}^N\int dz_x p(z_x|x_i)\ln  q(x_i|z_x)\nonumber\\ 
&\approx H(X)+\frac{1}{MN}\sum_{i,j=1}^{N,M} {-\frac{1}{2}}||(x_i - \mu_{X}({z_x}_{i,j}))||^2,
\end{align}
where $H(X)$ can be dropped from the loss since it doesn't change in optimization.

\item 
This type of decoder term is similar to that in part (a), but $X$ is decoded from two variables simultaneously. The corresponding loss term is $I^{D}(X;(Z_X,W_X))$. We find a variational loss by replacing $Z_X$ in part (a) by $(Z_X,W_X)$:
\begin{equation}
\Tilde{I}^{D}(X;(Z_X,W_X))\approx H(X)+\frac{1}{N}\sum_{i=1}^N\int dz_x dw_x p(z_x,w_x|x_i)\ln(q(x_i|z_x,w_x)),
\end{equation}
where, again, the entropy of $X$ can be dropped.

\item
This decoder term can be obtained by adding two decoders of type (a) together. In this case, the loss term is $I^{D}(X;Z)+I^{D}(Y;Z)$:
\begin{multline}
\Tilde{I}^{D}(X;Z)+\Tilde{I}^{D}(Y;Z)\approx H(X)+H(Y)\\+\frac{1}{N}\sum_{i=1}^N\int dz p(z|x_i)\ln(q(x_i|z))+\frac{1}{N}\sum_{i=1}^N\int dz p(z|y_i)\ln(q(y_i|z)),
\end{multline}
and the entropy terms can be dropped, again.
\end{enumerate}

\subsection{Internal decoders (decoders not on a leaf)}
\label{app:estimators}
\begin{figure}[h!]
\begin{center}
\begin{tikzpicture}[node distance={15mm}, thick,
main/.style = {draw, circle,minimum size=11mm},
comp/.style = {draw, circle,minimum size=9mm},
labs/.style = {}
] 
\node[comp] (1) {$Z_X$}; 
\node[comp] (2) [right of=1] {$Z_Y$}; 
\draw[->] (1) -- (2);

\end{tikzpicture}
\end{center}
\caption{Internal Decoders}
\label{Fig:Gdecoderleaf}
\end{figure}

Decoders of this type were discussed in the main text in Sec.~\ref{sec:MINE}. They correspond to the information between latent variables $Z_X$ and $Z_Y$. We use the MINE estimator to find variational bounds for such terms:
\begin{align}
\Tilde{I}^{D}_{\rm MINE}(Z_X;Z_Y)&=\int dz_x dz_y p(z_x,z_y)\ln \frac{e^{T(z_x,z_y)}}{\mathcal{Z}_{\text{norm}}}
\approx
\frac{1}{NM^2}\sum_{i,{j_x},{j_y}=1}^{N,M,M}\left[T(z_{x_{i,j_{x}}},z_{y_{i,j_{y}}}) - \ln \mathcal{Z}_{\text{norm}}\right].
\end{align}

We use $I_{\rm MINE}(X,Y)$ for the implementations in Sec.~\ref{results_noisy_mnist}. It is also possible to estimate terms of this type using other mutual information estimators such as SMILE and InfoNCE. The SMILE estimator \citep{song2019understanding} is a clipped version of the MINE estimator. The SMILE estimator improves the robustness of MINE by clipping the joint to marginal density ratio between $e^{-\tau}$ and $e^\tau$, where $\tau$ is some parameter that we set to  $5$ in our implementation:
\begin{equation}
I_{\rm SMILE}(X,Y)\ge\mathbb{E}_P [T(x,y)] - \log\left[\mathbb{E}_Q \left({\rm clip}(e^{T(x,y)},e^{-\tau},e^\tau)\right)\right].
\end{equation}
Smaller $\tau$ decreases the variance, but at a cost of a larger bias. At $\tau\rightarrow\infty$, $I_{\rm SMILE} \rightarrow I_{\rm MINE}$. We use $I_{\rm SMILE}(X,Y)$ for the implementations mentioned in Sec.~\ref{results_noisy_cifar}.

InfoNCE \citep{oord2018representation} is a mutual information estimator that uses a contrastive loss. When using a separable critic, it consists of two networks, $g$ and $h$. These networks compute $g(Z_X)$ and $h(Z_Y)$, which are then combined via a dot product to form the critic function $f(Z_X, Z_Y) = g(X) \cdot h(Y)$. The InfoNCE estimate of mutual information is then given by  
\begin{equation}
    \label{infonce}
    I_{\mathrm{InfoNCE}}(Z_X;Z_Y) \approx \mathbb{E} \left[ \log \frac{e^{f(Z_X, Z_Y)}}{\sum_{Z_Y'} e^{f(Z_X, Z_Y')}} \right],
\end{equation}  
where the denominator sums over negative samples $Z_Y'$. This loss can easily be recognized as mutual information if we make the identification $e^{f(Z_X, Z_Y)}=p(Z_X|Z_Y)$. We use $I_{\rm InfoNCE}(X,Y)$ for the implementations mentioned in Sec.~\ref{results_sota}. A detailed treatment of the different estimators and best practices of using them can also be found in \citet{abdelaleem2025accurate}.

\section{Deriving and designing variational losses: detailed implementations}
\subsection{Two variable Losses}
\label{App:Methods}
For completeness, we provide detailed implementations of methods outlined in Tbl.~\ref{table:methods}.

\subsubsection{Beta Variational Auto-Encoder}

\begin{figure}[h]
\begin{center}
\begin{tikzpicture}[node distance={15mm}, thick,
main/.style = {draw, circle,minimum size=11mm},
comp/.style = {draw, circle,minimum size=9mm},
labs/.style = {}
] 
\node[labs] (a) {$G_{\text{encoder}}$};
\node[main] (1) [below right of=a] {$X$}; 
\node[comp] (2) [below of=1] {$Z_X$};
\draw[->] (1) -- (2);

\node[labs] (b) [above right of=1] {$G_{\text{decoder}}$};
\node[main] (3) [below right of=b] {$X$}; 
\node[comp] (4) [below of=3] {$Z_X$};
\draw[->] (4) -- (3);
\end{tikzpicture}
\end{center}
\caption{Encoder and decoder graphs for the beta-variational auto-encoder method}
\label{Fig:betaVAE}
\end{figure}

A variational autoencoder \citep{Welling2014, Lerchner2016} compresses $X$ into a latent variable $Z_X$ and then reconstructs $X$ from the latent variable, cf.~Fig.~(\ref{Fig:betaDVIB}). The overall loss is a trade-off between the compression $I^{E}(X;Z_X)$ and the reconstruction $I^{D}(X;Z_X)$:
\begin{multline}
I^{E}(X;Z_X)-\beta I^{D}(X;Z_X) \le
\Tilde{I}^{E}(X;Z_X)-\beta \Tilde{I}^{D}(X;Z_X) \\
\lesssim \frac{1}{N}\sum_{i=1}^N D_{\rm KL}(p(z_x|x_i) \Vert r(z_x)) -\beta \left(H(X)+\frac{1}{N}\sum_{i=1}^N\int dz_x p(z_x|x_i)\ln(q(x_i|z_x))\right).
\end{multline}
$H(X)$ is a constant with respect to the minimization, and it can be omitted from the loss. Similar to the main text, DVSIB case, we make ansatzes for forms of each of the variational distributions. We choose parametric distribution families and learn the nearest distribution in these families consistent with the data. Specifically, we assume $p(z_x|x)$ is a normal distribution with mean $\mu_{Z_X}(X)$ and variance $\Sigma_{Z_X}(X)$. We learn the mean and the log-variance as neural networks. We also assume that $q(x|z_x)$ is normal with a mean $\mu_{X}(z_x)$ and a unit variance.  Finally, we assume that $r(z_x)$ is drawn from a standard normal distribution. We then use the re-parameterization trick to produce samples of $z_{x_j}(x)=\mu(x)+\sqrt{\Sigma_{Z_X}(x)} \eta_j$ from $p(z_x|x)$, where $\eta$ is drawn from a standard normal distribution. Overall, this gives:
\begin{multline}
L_{\text{VAE}}=\frac{1}{2N}\sum_{i=1}^N \left[\text{Tr}({\Sigma_{Z_X}(x_i)}) +\vec{\mu}_{Z_X}(x_i)^T\vec{\mu}_{Z_X}(x_i)-k_{Z_X}-\ln\det(\Sigma_{Z_X}(x_i)) \right]\\ -\beta \left(\frac{1}{MN}\sum_{i=1}^N\sum_{j=1}^M {-\frac{1}{2}}(x_i - \mu_{X}(z_{x_j}))^T(x_i - \mu_{X}(z_{x_j})) \right).
\end{multline}
This is the same loss as for a beta auto-encoder. However, following the convention in the Information Bottleneck literature \citep{Bialek2000,Tishby2013}, our $\beta$ is the inverse of the one typically used for beta auto-encoders. A small $\beta$ in our case results in a stronger compression, while a large $\beta$ results in a better reconstruction.

\subsubsection{Deep Variational Information Bottleneck}
\begin{figure}[h]
\begin{center}
\begin{tikzpicture}[node distance={15mm}, thick,
main/.style = {draw, circle,minimum size=11mm},
comp/.style = {draw, circle,minimum size=9mm},
labs/.style = {}
] 
\node[labs] (a) {$G_{\text{encoder}}$};
\node[main] (x) [below right of=a] {$X$}; 
\node[main] (y) [right of=x] {$Y$};
\node[comp] (zx) [below of=x] {$Z_X$}; 
\draw (x) -- (y);
\draw[->] (x) -- (zx);

\node[labs] (b) [above right of=y] {$G_{\text{decoder}}$};
\node[main] (mx) [below right of=b] {$X$}; 
\node[main] (my) [right of=mx] {$Y$};
\node[comp] (mzx) [below of=mx] {$Z_X$}; 
\draw[->] (mzx) -- (my);
\end{tikzpicture}
\end{center}
\caption{Encoder and decoder graphs for the Deep Variational Information Bottleneck.}
\label{Fig:betaDVIB}
\end{figure}

Just as in the beta auto-encoder, we immediately write down the loss function for the information bottleneck. Here, the encoder graph compresses $X$ into $Z_X$, while the decoder tries to maximize the information between the compressed variable and the relevant variable $Y$, cf.~Fig.~(\ref{Fig:betaDVIB}). The resulting loss function is:
\begin{equation}
L_{\text{IB}} = I^{E}(X;Y)+I^{E}(X;Z_X)-\beta I^{D}(Y;Z_X).
\end{equation}
Here the information between $X$ and $Y$ does not depend on $p(z_x|x)$ and can dropped in the optimization.

Thus the Deep Variational Information Bottleneck \citep{Murphy2017} becomes :
\begin{multline}
L_{\text{DVIB}}\approx \frac{1}{N}\sum_{i=1}^N D_{\rm KL}(p(z_x|x_i) \Vert r(z_x)) -\beta\left( \frac{1}{N}\sum_{i=1}^N\int dz_x p(z_x|x_i)\ln(q(y_i|z_x))\right),
\end{multline}
where we dropped $H(Y)$ since it doesn't change in the optimization. 

As we have been doing before, we choose to parameterize all these distributions by Gaussians and their means and their log variances are learned by neural networks. Specifically, we parameterize $p(z_x|x)=N(\mu_{z_x}(x),\Sigma_{z_x})$, $r(z_x)=N(0,I)$, and $q(y|z_x)=N(\mu_Y,I)$. Again we can use the reparameterization trick and sample from $p(z_x|x_i)$ by $z_{x_j}(x)=\mu(x)+\sqrt{\Sigma_{z_x}(x)} \eta_j$ where $\eta$ is drawn from a standard normal distribution.

\subsubsection{Beta Deep Variational CCA}
\begin{figure}[h]
\begin{center}
\begin{tikzpicture}[node distance={15mm}, thick,
main/.style = {draw, circle,minimum size=11mm},
comp/.style = {draw, circle,minimum size=9mm},
labs/.style = {}
] 
\node[labs] (a) {$G_{\text{encoder}}$};
\node[main] (x) [below right of=a] {$X$}; 
\node[main] (y) [right of=x] {$Y$};
\node[comp] (zx) [below of=x] {$Z_X$}; 
\draw (x) -- (y);
\draw[->] (x) -- (zx);

\node[labs] (b) [above right of=y] {$G_{\text{decoder}}$};
\node[main] (mx) [below right of=b] {$X$}; 
\node[main] (my) [right of=mx] {$Y$};
\node[comp] (mzx) [below of=mx] {$Z_X$}; 
\draw[->] (mzx) -- (my);
\draw[->] (mzx) -- (mx);
\end{tikzpicture}
\end{center}
\caption{Encoder and decoder graphs for beta Deep Variational CCA.}
\label{Fig:betaDCCA}
\end{figure}
beta-DVCCA, cf.~Fig.~\ref{Fig:betaDCCA}, is similar to the traditional information bottleneck, but now $X$ and $Y$ are both used as relevance variables:
\begin{equation}
L_{\rm DVCCA}=\Tilde{I}^{E}(X;Y)+\Tilde{I}^{E}(X;Z_X)-\beta (\Tilde{I}^{D}(Y;Z_X)+\Tilde{I}^{D}(X;Z_X)).
\end{equation}

Using the same library of terms as before, we find:
\begin{multline}
L_{\text{DVCCA}}\approx\frac{1}{N}\sum_{i=1}^N D_{\rm KL}(p(z_x|x_i) \Vert r(z_x))\\
-\beta\left( \frac{1}{N}\sum_{i=1}^N\int dz_x p(z_x|x_i)\ln(q(y_i|z_x)) + \frac{1}{N}\sum_{i=1}^N\int dz_x p(z_x|x_i)\ln(q(x_i|z_x))\right).
\end{multline}
This is similar to the loss function of the deep variational CCA \citep{Livescu2016}, but now it has a trade-off parameter $\beta$. It trades off the compression into $Z$ against the reconstruction of $X$ and $Y$ from the compressed variable $Z$.

\subsubsection{beta joint-Deep Variational CCA}
\begin{figure}[h]
\begin{center}
\begin{tikzpicture}[node distance={15mm}, thick,
main/.style = {draw, circle,minimum size=11mm},
comp/.style = {draw, circle,minimum size=9mm},
labs/.style = {}
] 
\node[labs] (a) {$G_{\text{encoder}}$};
\node[main] (1) [below right of=a] {$X$};
\node[comp] (3) [below right of=1] {$Z$};
\node[main] (2) [above right of=3] {$Y$};
\draw (1) -- (2);
\draw[->] (1) -- (3);
\draw[->] (2) -- (3);

\node[labs] (b) [above right of=2] {$G_{\text{decoder}}$};
\node[main] (x) [below right of=b] {$X$};
\node[comp] (z) [below right of=x] {$Z$};
\node[main] (y) [above right of=z] {$Y$};
\draw[->] (z) -- (x);
\draw[->] (z) -- (y);
\end{tikzpicture}
\end{center}
\caption{Encoder and decoder graphs for beta joint-Deep Variational CCA.}
\label{Fig:betaDCCAj}
\end{figure}

Joint deep variational CCA \citep{Livescu2016}, cf.~Fig.~\ref{Fig:betaDCCAj}, compresses $(X,Y)$ into one $Z$ and then reconstructs the individual terms $X$ and $Y$,
\begin{equation}
L_{\rm DVCCA}=I^{E}(X;Y)+I^{E}((X,Y);Z)-\beta (I^{D}(Y;Z)+I^{D}(X;Z)).
\end{equation}

Using the terms we derived, the loss function is:
\begin{multline}
L_{\text{DVCCA}}\approx\frac{1}{N}\sum_{i=1}^N D_{KL}(p(z|x_i,y_i) \Vert r(z)) \\-\beta \left(\frac{1}{N}\sum_{i=1}^N\int dz p(z|x_i)\ln(q(y_i|z)) + \frac{1}{N}\sum_{i=1}^N\int dz p(z|x_i)\ln(q(x_i|z))\right).
\end{multline}
The information between $X$ and $Y$ does not change under the minimization and can be dropped.

\subsection{Auxiliary private variable models}
\label{App:private}

\begin{table}[h!]
    \caption{Method descriptions, variational losses, and the Bayesian Network graphs for each DR method with private variables derived in our framework. See Appx.~\ref{App:Library} for details.}
    \label{table:methods-private}

    \begin{center}
    \begin{tabular}
    {|m{0.68\textwidth}|>{\centering\arraybackslash}m{0.16\textwidth}|>{\centering\arraybackslash}m{0.16\textwidth}|}
    \hline
        \textbf{Method Description}
        & \textbf{$G_{\text{encoder}}$} & \textbf{$G_{\text{decoder}}$} \\
        \hline
        \textbf{beta-DVCCA-private}: Two models trained, compressing either $X$ or $Y$, while reconstructing both $X$ and $Y$, and simultaneously learning private information $W_X$ and $W_Y$. (Only $X$ graphs/loss shown).
        
        $\sloppy L_{\text{DVCCA-p}}=\Tilde{I}^{E}(X;Z)+\Tilde{I}^{E}(X;W_X)+\Tilde{I}^{E}(Y;W_Y)-\beta (\Tilde{I}^{D}(X;(W_X,Z))+\Tilde{I}^{D}(Y;(W_Y,Z)))$\newline

        \textbf{DVCCA-private} \citep{Livescu2016}: $\beta$-DVCCA-p with $\beta=1$.
        &
        \adjustbox{width=25mm}{
        \begin{tikzpicture}[node distance={15mm}, thick,
            main/.style = {draw, circle,minimum size=11mm},
            comp/.style = {draw, circle,minimum size=9mm},
            labs/.style = {}
            ]
            \node[main] (1) {$X$};
            \node[comp] (3) [below right of=1] {$Z$};
            \node[main] (2) [above right of=3] {$Y$};
            \node[comp] (4) [below of=1] {$W_X$};
            \node[comp] (5) [below of=2] {$W_Y$};
            \draw (1) -- (2);
            \draw[->] (1) -- (3);
            \draw[->] (1) -- (4);
            \draw[->] (2) -- (5);
        \end{tikzpicture}
        }
        &
        \adjustbox{width=25mm}{
        \begin{tikzpicture}[node distance={15mm}, thick,
            main/.style = {draw, circle,minimum size=11mm},
            comp/.style = {draw, circle,minimum size=9mm},
            labs/.style = {}
        ] 
            \node[main] (x) {$X$};
            \node[comp] (z) [below right of=x] {$Z$};
            \node[main] (y) [above right of=z] {$Y$};
            \node[comp] (wx) [below of=x] {$W_X$};
            \node[comp] (wy) [below of=y] {$W_Y$};
            \draw[->] (wx) -- (x);
            \draw[->] (z) -- (x);
            \draw[->] (wy) -- (y);
            \draw[->] (z) -- (y);
        \end{tikzpicture}
        
        }\\
        \hline
        \textbf{beta-joint-DVCCA-private}: A single model trained using a concatenated variable $[X,Y]$, learning one latent representation $Z$, and simultaneously learning private information $W_X$ and $W_Y$.
        $\sloppy L_{\text{jDVCCA-p}}=\Tilde{I}^{E}((X,Y);Z)+\Tilde{I}^{E}(X;W_X)+$$\sloppy\Tilde{I}^{E}(Y;W_Y)-\beta (\Tilde{I}^{D}(X;(W_X,Z))+\Tilde{I}^{D}(Y;(W_Y,Z)))$ \newline

        \textbf{joint-DVCCA-private}\citep{Livescu2016}: $\beta$-jDVCCA-p with $\beta=1$.
        &
        \adjustbox{width=25mm}{
        \begin{tikzpicture}[node distance={15mm}, thick,
            main/.style = {draw, circle,minimum size=11mm},
            comp/.style = {draw, circle,minimum size=9mm},
            labs/.style = {}
            ] 
            \node[main] (1) {$X$};
            \node[comp] (3) [below right of=1] {$Z$};
            \node[main] (2) [above right of=3] {$Y$};
            \node[comp] (4) [below of=1] {$W_X$};
            \node[comp] (5) [below of=2] {$W_Y$};
            \draw (1) -- (2);
            \draw[->] (1) -- (3);
            \draw[->] (2) -- (3);
            \draw[->] (1) -- (4);
            \draw[->] (2) -- (5);
        \end{tikzpicture}
        }
        &
        \adjustbox{width=25mm}{
        \begin{tikzpicture}[node distance={15mm}, thick,
            main/.style = {draw, circle,minimum size=11mm},
            comp/.style = {draw, circle,minimum size=9mm},
            labs/.style = {}
        ] 
            \node[main] (x) {$X$};
            \node[comp] (z) [below right of=x] {$Z$};
            \node[main] (y) [above right of=z] {$Y$};
            \node[comp] (wx) [below of=x] {$W_X$};
            \node[comp] (wy) [below of=y] {$W_Y$};
            \draw[->] (wx) -- (x);
            \draw[->] (z) -- (x);
            \draw[->] (wy) -- (y);
            \draw[->] (z) -- (y);
        \end{tikzpicture}
        
        }\\
        \hline
    
    \textbf{DVSIB-private}: A symmetric model trained, producing $Z_X$ and $Z_Y$, while simultaneously learning private information $W_X$ and $W_Y$.\newline
    $\sloppy L_{\text{DVSIBp}}=\Tilde{I}^{E}(X;W_X)+\Tilde{I}^{E}(X;Z_X) +$\newline$\Tilde{I}^{E}(Y;Z_Y)+\Tilde{I}^{E}(Y;W_Y)
-$\newline
$\sloppy\beta \left(\Tilde{I}^{D}_{\text{MINE}}(Z_X;Z_Y)+\Tilde{I}^{D}(X;(Z_X,W_X))+\Tilde{I}^{D}(Y;(Z_Y,W_Y))\right)$
    &
    \adjustbox{width=25mm}{
    \begin{tikzpicture}[node distance={15mm}, thick,
        main/.style = {draw, circle,minimum size=11mm},
        comp/.style = {draw, circle,minimum size=9mm},
        labs/.style = {}
        ] 
        \node[main] (x) {$X$}; 
        \node[main] (y) [right of=x] {$Y$};
        \node[comp] (zx) [below of=x] {$Z_X$}; 
        \node[comp] (zy) [below of=y] {$Z_Y$};
        \node[comp] (wx) [below left of=x] {$W_X$}; 
        \node[comp] (wy) [below right of=y] {$W_Y$};
        \draw (x) -- (y);
        \draw[->] (x) -- (zx);
        \draw[->] (y) -- (zy);
        \draw[->] (x) -- (wx);
        \draw[->] (y) -- (wy);
    \end{tikzpicture}
    }
    &
    \adjustbox{width=25mm}{
    \begin{tikzpicture}[node distance={15mm}, thick,
        main/.style = {draw, circle,minimum size=11mm},
        comp/.style = {draw, circle,minimum size=9mm},
        labs/.style = {}
        ] 
        \node[comp] (mwx) {$W_X$}; 
        \node[main] (mx) [above right of=mwx] {$X$}; 
        \node[main] (my) [right of=mx] {$Y$};
        \node[comp] (mzx) [below of=mx] {$Z_X$}; 
        \node[comp] (mzy) [below of=my] {$Z_Y$};
        \node[comp] (mwy) [below right of=my] {$W_Y$};
        \draw[->] (mzx) -- (mzy);
        \draw[->] (mzx) -- (mx);
        \draw[->] (mzy) -- (my);
        \draw[->] (mwx) -- (mx);
        \draw[->] (mwy) -- (my);
        
    \end{tikzpicture}
    }\\
    \hline
    
    \end{tabular}
    \end{center}
\end{table}

Most of the losses examined so far aim to capture the correlation between $X$ and $Y$. If, however, one wishes additionally to separate out details about just $X$ and just $Y$, it is helpful to introduce additional auxiliary private variables $W_X$, $W_Y$ intended to capture these details. In DVCCA-private (beta-DVCCA-private) for example, the decoder factorization illustrates this clearly: $P(x|z,w_x)P(x|z,w_y)P(z)P(w_x)P(w_y)$. This shows that $X$ is reconstructed from the private variable $W_X$ that only has information about $X$ and the shared variable $Z$, and similarly $Y$ is reconstructed from the private variable $W_Y$ which only has information about $Y$ and the shared variable $Z$. In the Noisy MNIST dataset this type of factorization encourages $Z$ to exclusively capture details about digit identity, $W_X$ to capture details about scale and rotation, and $W_Y$ details about noise. Similar results hold for beta-joint-DVCCA-private, and DVSIB-private. See \ref{App:results-mnist} for Noisy MNIST visualizations.

\subsubsection{beta (joint) Deep Variational CCA-private}

\begin{figure}[h]
\begin{center}
\begin{tikzpicture}[node distance={15mm}, thick,
main/.style = {draw, circle,minimum size=11mm},
comp/.style = {draw, circle,minimum size=9mm},
labs/.style = {}
] 
\node[labs] (a) {$G_{\text{encoder}}$};
\node[main] (1) [below right of=a] {$X$};
\node[comp] (3) [below right of=1] {$Z$};
\node[main] (2) [above right of=3] {$Y$};
\node[comp] (4) [below of=1] {$W_X$};
\node[comp] (5) [below of=2] {$W_Y$};
\draw (1) -- (2);
\draw[->] (1) -- (3);
\draw[->] (1) -- (4);
\draw[->] (2) -- (5);

\node[labs] (b) [above right of=2] {$G_{\text{decoder}}$};
\node[main] (x) [below right of=b] {$X$};
\node[comp] (z) [below right of=x] {$Z$};
\node[main] (y) [above right of=z] {$Y$};
\node[comp] (wx) [below of=x] {$W_X$};
\node[comp] (wy) [below of=y] {$W_Y$};
\draw[->] (wx) -- (x);
\draw[->] (z) -- (x);
\draw[->] (wy) -- (y);
\draw[->] (z) -- (y);
\end{tikzpicture}
\end{center}
\caption{Encoder and decoder graphs for beta Deep Variational CCA-private}
\label{Fig:betaDVCCAp}
\end{figure}

This is a generalization of the Deep Variational CCA \cite{Livescu2016} to include private information, cf.~Fig.~\ref{Fig:betaDVCCAp}. Here $X$ is encoded into a shared latent variable $Z$ and a private latent variable $W_X$. Similarly $Y$ is encoded into the same shared variable and a different private latent variable $W_Y$. $X$ is reconstructed from $Z$ and $W_X$, and $Y$ is reconstructed from $Z$ and $W_Y$. In the joint version $(X,Y)$ are compressed jointly in $Z$ similar to the previous joint methods. What follows is the loss $X$ version of beta Deep Variational CCA-private.
\begin{multline}
L_{\rm DVCCAp}=I^{E}(X;Y)+I^{E}((X,Y);Z)+I^{E}(X;W_X)+I^{E}(Y;W_Y)\\-\beta (I^{D}(X;(W_X,Z))+I^{D}(Y;(W_Y,Z))).
\end{multline}
After the usual variational manipulations, this becomes:
\begin{multline}
L_{\text{DVCCAp}}\approx\frac{1}{N}\sum_{i=1}^N D_{\rm KL}(p(z|x_i) \Vert r(z))+\frac{1}{N}\sum_{i=1}^N D_{\rm KL}(p(w_x|x_i) \Vert r(w_x))\\+
\frac{1}{N}\sum_{i=1}^N D_{\rm KL}(p(w_y|y_i) \Vert r(w_y)) -\beta \left(\frac{1}{N}\sum_{i=1}^N\int dz dw_x p(w_x|x_i)p(z|x_i)\ln(q(y_i|z,w_x)) \right.\\
+ \left. \frac{1}{N}\sum_{i=1}^N\int dz dw_y p(w_y|y_i)p(z|x_i)\ln(q(x_i|z,w_y))\right).
\end{multline}

\subsubsection{Deep Variational Symmetric Information Bottleneck}
This has been analyzed in detail in the main text, Sec.~\ref{sec:DVSIB}, and will not be repeated here.

\subsubsection{Deep Variational Symmetric Information Bottleneck-private}
\begin{figure}[h]
\begin{center}
\begin{tikzpicture}[node distance={15mm}, thick,
main/.style = {draw, circle,minimum size=11mm},
comp/.style = {draw, circle,minimum size=9mm},
labs/.style = {}
] 
\node[labs] (a) {$G_{\text{encoder}}$};
\node[main] (x) [below right of=a] {$X$}; 
\node[main] (y) [right of=x] {$Y$};
\node[comp] (zx) [below of=x] {$Z_X$}; 
\node[comp] (zy) [below of=y] {$Z_Y$};
\node[comp] (wx) [below left of=x] {$W_X$}; 
\node[comp] (wy) [below right of=y] {$W_Y$};
\draw (x) -- (y);
\draw[->] (x) -- (zx);
\draw[->] (y) -- (zy);
\draw[->] (x) -- (wx);
\draw[->] (y) -- (wy);

\node[comp] (mwx) [right of=wy] {$W_X$}; 
\node[main] (mx) [above right of=mwx] {$X$}; 
\node[labs] (b) [above left of=mx] {$G_{\text{decoder}}$};
\node[main] (my) [right of=mx] {$Y$};
\node[comp] (mzx) [below of=mx] {$Z_X$}; 
\node[comp] (mzy) [below of=my] {$Z_Y$};
\node[comp] (mwy) [below right of=my] {$W_Y$};
\draw[->] (mzx) -- (mzy);
\draw[->] (mzx) -- (mx);
\draw[->] (mzy) -- (my);
\draw[->] (mwx) -- (mx);
\draw[->] (mwy) -- (my);

\end{tikzpicture}
\end{center}
\caption{Encoder and decoder graphs for DVSIB-private.}
\label{Fig:betaDVSIBp}
\end{figure}

This is a generalization of the Deep Variational Symmetric Information Bottleneck to include private information. Here $X$ is encoded into a shared latent variable $Z_X$ and a private latent variable $W_X$. Similarly, $Y$ is encoded into its own shared $Z_Y$ variable and a private latent variable $W_Y$. $X$ is reconstructed from $Z_X$ and $W_X$, and $Y$ is reconstructed from $Z_Y$ and $W_Y$. $Z_X$ and $Z_Y$ are constructed to be maximally informative about each another. This results in
\begin{multline}
L_{\text{DVSIBp}}=I^{E}(X;W_X)+I^{E}(X;Z_X)+I^{E}(Y;Z_Y)+I^{E}(Y;W_Y)\\
-\beta \left(I^{D}(Z_X;Z_Y)+I^{D}(X;(Z_X,W_X))+I^{D}(Y;(Z_Y,W_Y))\right).
\end{multline}
After the usual variational manipulations, this becomes (see also main text):
\begin{multline}
L_{\text{DVSIBp}}\approx \frac{1}{N}\sum_{i=1}^N D_{\rm KL}(p(z_x|x_i) \Vert r(z_x)) + \frac{1}{N}\sum_{i=1}^N D_{\rm KL}(p(z_y|x_i) \Vert r(z_y)) \\+\frac{1}{N}\sum_{i=1}^N D_{\rm KL}(p(w_x|x_i) \Vert r(w_x))+\frac{1}{N}\sum_{i=1}^N D_{\rm KL}(p(w_y|y_i) \Vert r(w_y)) \\-\beta \left(\int dz_x dz_y p(z_x,z_y)\ln\frac{e^{T(z_x,z_y)}}{\mathcal{Z}_{\text{norm}}}+\frac{1}{N}\sum_{i=1}^N\int dz_ydw_y p(w_y|y_i)p(z_y|y_i)\ln(q(y_i|z_y,w_y))\right. \\\left.+ \frac{1}{N}\sum_{i=1}^N\int dz_x dw_x p(w_x|x_i)p(z_x|x_i)\ln(q(x_i|z_x,w_x))\right),
\end{multline}
where 
\begin{equation}
\mathcal{Z}_{\rm norm}=\int dz_x dz_y p(z_x)p(z_y)e^{T(z_x,z_y)}.
\end{equation}

\subsubsection{Private Variable Model Results on Noisy MNIST}
Details of experiments can be found in Tbl.~\ref{table:SVM-Y-private}.
\begin{table}[h!]
\caption{Maximum accuracy from a linear SVM and the optimal $k_Z$ and $\beta$ for variational DR methods with private variables reported on the $Y$ (above the line) and the joint $[X,Y]$ (below the line) datasets. ($^\dag$ fixed values)}
\label{table:SVM-Y-private}
\begin{center}
\begin{tabular}{|l|c|c|c|c|c|c|}
\hline
\textbf{Method} & \textbf{Acc. \%} & \textbf{${k_Z}_\textbf{best}$} & $95\%$ \textbf{${k_Z}_{\text{range}}$} & \bm{$\beta_\text{best}$} & $95\%$ \textbf{$\beta_\text{range}$} \\
\hline
Baseline & 90.8 & 784$^\dag$ & - & - & - \\
DVCCA-p & 92.1 & 16 & [16,256*] & 1$^\dag$ & - \\
$\beta$-DVCCA-p & 95.5 & 16 & [\textbf{4},256*] & 1024 & [1,1024*] \\
DVSIB-p & \textbf{97.8} & 256 & [\textbf{8},256*] & 32 & [2,1024*] \\
\hline
jBaseline & 91.9 & 1568$^\dag$ & - & - & -  \\
jDVCCA-p & 92.5 & 64 & [32,265*] & 1$^\dag$ & - \\
$\beta$-jDVCCA-p & 92.7 & 256 & [\textbf{4},265*] & 2 & [1,1024*] \\
\hline

\end{tabular}
\end{center}
\end{table}

\subsection{Multi-variable losses (more than two views/variables)}
\label{App:MultiViewLosses}
Several multi-variable losses that have appeared in the literature can be rederived within our framework. These methods can be broadly categorized into two groups: supervised multiview information bottleneck (SMVIB) and unsupervised multiview information bottleneck (UMVIB). The key distinction between them is the presence of a supervising signal $Y$ in addition to the multiple views or modalities of the data $X_i$. Here, we demonstrate how our DVMIB framework provides a natural way to extend and unify these methods.

\subsubsection{Unsupervised Multi-View IB}

Several approaches in the literature, such as \cite{Hu2021,Ye2020}, focus on compressing multiple views or modalities $X_i$ into one or more latent variables $Z$ (or $Z_i$) and reconstructing $X_i$ from $Z$ (or both $Z$ and $Z_i$). Within this framework, we can consider two primary formulations:

\begin{itemize}
    \item \textbf{Single shared latent representation (UMVIB$_1$):}  
    \begin{figure}[h]
    \begin{center}
    \begin{tikzpicture}[node distance={15mm}, thick,
    main/.style = {draw, circle,minimum size=11mm},
    comp/.style = {draw, circle,minimum size=9mm},
    labs/.style = {}
    ] 
    \node[labs] (a) {$G_{\text{encoder}}$};
    \node[main] (x1) [below right of=a] {$X_1$}; 
    \node[main] (x2) [right of=x1] {$X_2$};
    \node[main] (x3) [right of=x2] {$X_3$}; 
    \node[comp] (z) [below of=x2] {$Z$};
    
    \draw[->] (x1) -- (z);
    \draw[->] (x2) -- (z);
    \draw[->] (x3) -- (z);
    
    \node[labs] (b) [above right of=x3] {$G_{\text{decoder}}$};
    \node[main] (mx1) [below right of=b] {$X_1$}; 
    \node[main] (mx2) [right of=mx1] {$X_2$};
    \node[main] (mx3) [right of=mx2] {$X_3$}; 
    \node[comp] (mz) [below of=mx2] {$Z$};
    
    \draw[->] (mz) -- (mx1);
    \draw[->] (mz) -- (mx2);
    \draw[->] (mz) -- (mx3);
    
    \end{tikzpicture}
    \end{center}
    \caption{Encoder and decoder graphs for a single shared latent representation UMVIB}
    \label{app:umvib1}
    \end{figure}
    
    All views $X_1$, $X_2$, and $X_3$ are compressed into a single latent variable $Z$, which is then used to reconstruct each view. The model is illustrated in Figure~\ref{app:umvib1}, and the corresponding loss function is:
    \begin{equation}
    L_{\text{UMVIB}_1}= \Tilde{I}^{E}(X_1,X_2,X_3;Z)-\beta (\Tilde{I}^{D}(X_1;Z)+\Tilde{I}^{D}(X_2;Z)+\Tilde{I}^{D}(X_3;Z)).
    \end{equation}

    Using the same library of terms as before:

    \begin{multline}
    L_{\text{UMVIB}_1}\approx\frac{1}{N}\sum_{i=1}^N D_{\rm KL}(p(z|{x_1}_i,{x_2}_i,{x_3}_i) \Vert r(z))\\
    -\beta\left( \frac{1}{N}\sum_{i=1}^N\int dz p(z|{x_1}_i,{x_2}_i,{x_3}_i)\ln(q({x_1}_i|z)) + \frac{1}{N}\sum_{i=1}^N\int dz p(z|{x_1}_i,{x_2}_i,{x_3}_i)\ln(q({x_2}_i|z))\right.\\
    \left.+\frac{1}{N}\sum_{i=1}^N\int dz p(z|{x_1}_i,{x_2}_i,{x_3}_i)\ln(q({x_3}_i|z))\right).
    \end{multline}
    
    \item \textbf{Shared and private latent representations (UMVIB$_2$):}  
    \begin{figure}[h]
    \begin{center}
    \begin{tikzpicture}[node distance={15mm}, thick,
    main/.style = {draw, circle,minimum size=11mm},
    comp/.style = {draw, circle,minimum size=9mm},
    labs/.style = {}
    ] 
    \node[labs] (a) {$G_{\text{encoder}}$};
    \node[main] (x1) [below right of=a] {$X_1$}; 
    \node[main] (x2) [right of=x1] {$X_2$};
    \node[main] (x3) [right of=x2] {$X_3$}; 
    \node[comp] (z1) [below of=x1] {$Z_1$};
    \node[comp] (z2) [below of=x2] {$Z_2$};
    \node[comp] (z3) [below of=x3] {$Z_3$};
    \node[comp] (z) [below left of=x1] {$Z$};

    \draw[->] (x1) -- (z1);
    \draw[->] (x2) -- (z2);
    \draw[->] (x3) -- (z3);
    \draw[->] (x1) -- (z);
    \draw[->] (x2) -- (z);
    \draw[->] (x3) -- (z);
    
    \node[labs] (b) [above right of=x3] {$G_{\text{decoder}}$};
    \node[main] (mx1) [below right of=b] {$X_1$}; 
    \node[main] (mx2) [right of=mx1] {$X_2$};
    \node[main] (mx3) [right of=mx2] {$X_3$}; 
    \node[comp] (mz1) [below of=mx1] {$Z_1$};
    \node[comp] (mz2) [below of=mx2] {$Z_2$};
    \node[comp] (mz3) [below of=mx3] {$Z_3$};
    \node[comp] (mz) [below left of=mx1] {$Z$};
    
    \draw[->] (mz) -- (mx1);
    \draw[->] (mz) -- (mx2);
    \draw[->] (mz) -- (mx3);
    \draw[->] (mz1) -- (mx1);
    \draw[->] (mz2) -- (mx2);
    \draw[->] (mz3) -- (mx3);
    \end{tikzpicture}
    \end{center}
    \caption{Encoder and decoder graphs for shared and private latent representations UMVIB}
    \label{app:umvib2}
    \end{figure}
    
    An alternative formulation introduces both a shared latent variable $Z$ and private latent variables $Z_1$, $Z_2$, and $Z_3$, corresponding to each view. This structure extends DVCCA-private to three variables. Each view $X_i$ is reconstructed from its private latent representation $Z_i$ along with the shared latent variable $Z$. The model is illustrated in Fig.~\ref{app:umvib2}, and the corresponding loss function is:
    \begin{multline}
    L_{\text{UMVIB}_2}=\Tilde{I}^{E}((X_1,X_2,X_3);Z)+\Tilde{I}^{E}(X_1;Z_1)+\Tilde{I}^{E}(X_2;Z_2)+\Tilde{I}^{E}(X_3;Z_3)\\
    -\beta (\Tilde{I}^{D}(X_1;(Z,Z_1))+\Tilde{I}^{D}(X_2;(Z,Z_2))+\Tilde{I}^{D}(X_3;(Z,Z_3))).
    \end{multline}

    Using the same library of terms as before, this becomes:
    \begin{multline}
    L_{\text{UMVIB}_2}\approx\frac{1}{N}\sum_{i=1}^N D_{\rm KL}(p(z|{x_1}_i,{x_2}_i,{x_3}_i) \Vert r(z))+\sum_{j=1}^3\frac{1}{N}\sum_{i=1}^N D_{\rm KL}(p(z_j|{x_j}_i) \Vert r_j(z))\\
    -\beta\left( \frac{1}{N}\sum_{i=1}^N\int dz dz_1 dz_2 dz_3 p(z|{x_1}_i,{x_2}_i,{x_3}_i)p(z_1|{x_1}_i)p(z_2|{x_2}_i)p(z_3|{x_3}_i)\ln(q({x_1}_i|z,z_1))\right.\\
    \left.+ \frac{1}{N}\sum_{i=1}^N\int dz dz_1 dz_2 dz_3 p(z|{x_1}_i,{x_2}_i,{x_3}_i)p(z_1|{x_1}_i)p(z_2|{x_2}_i)p(z_3|{x_3}_i)\ln(q({x_2}_i|z,z_2))\right.\\
    \left.+\frac{1}{N}\sum_{i=1}^N\int dz dz_1 dz_2 dz_3 p(z|{x_1}_i,{x_2}_i,{x_3}_i)p(z_1|{x_1}_i)p(z_2|{x_2}_i)p(z_3|{x_3}_i)\ln(q({x_3}_i|z,z_3))\right).
    \end{multline}

\end{itemize}

\subsubsection{Supervised Multi-View IB}

Several approaches in the literature, such as \cite{VanderSchaar2021,Zhou2019,Elgamal2022,Meiser2022}, focus on compressing multiple views or modalities $X_i$ into one or more latent variables $Z$ (or $Z$ and $Z_i$) and uses $Z$ to predict a supervisory signal/label $Y$ (or in addition to using both $Z$ and $Z_i$ to reconstruct $X_i$ as well). Within this framework, we can also consider two primary formulations:

\begin{itemize}
    \item \textbf{Single shared latent representation (SMVIB$_1$):}  
    \begin{figure}[h]
    \begin{center}
    \begin{tikzpicture}[node distance={15mm}, thick,
    main/.style = {draw, circle,minimum size=11mm},
    comp/.style = {draw, circle,minimum size=9mm},
    labs/.style = {}
    ] 
    \node[labs] (a) {$G_{\text{encoder}}$};
    \node[main] (x1) [below right of=a] {$X_1$}; 
    \node[main] (x2) [right of=x1] {$X_2$};
    \node[main] (x3) [right of=x2] {$X_3$}; 
    \node[main] (y) [right of=x3] {$Y$}; 
    \node[comp] (z) [below of=x3] {$Z$};
    
    \draw[->] (x1) -- (z);
    \draw[->] (x2) -- (z);
    \draw[->] (x3) -- (z);
    
    \node[labs] (b) [above right of=y] {$G_{\text{decoder}}$};
    \node[main] (mx1) [below right of=b] {$X_1$}; 
    \node[main] (mx2) [right of=mx1] {$X_2$};
    \node[main] (mx3) [right of=mx2] {$X_3$}; 
    \node[main] (my) [right of=mx3] {$Y$}; 
    \node[comp] (mz) [below of=mx3] {$Z$};
    
    \draw[->] (mz) -- (my);
    
    \end{tikzpicture}
    \end{center}
    \caption{Encoder and decoder graphs for a single shared latent representation SMVIB}
    \label{app:smvib1}
    \end{figure}
    
    All views $X_1$, $X_2$, and $X_3$ are compressed into a single latent variable $Z$, which is then used to predict the supervising signal $Y$. The model is illustrated in Figure~\ref{app:smvib1}, and the corresponding loss function is:  
    \begin{equation}
        L_{\text{SVMIB}_1}= \Tilde{I}^{E}(X_1,X_2,X_3;Z)-\beta (\Tilde{I}^{D}(Y;Z)).
    \end{equation}

    Using the same library of terms as before, we get:
    \begin{multline}
    L_{\text{SMVIB}_1}\approx\frac{1}{N}\sum_{i=1}^N D_{\rm KL}(p(z|{x_1}_i,{x_2}_i,{x_3}_i) \Vert r(z))
    -\beta\left( \frac{1}{N}\sum_{i=1}^N\int dz p(z|{x_1}_i,{x_2}_i,{x_3}_i)\ln(q(y_i|z))\right).
    \end{multline}
    
    \item \textbf{Shared and private latent representations (SMVIB$_2$):}  
    \begin{figure}[h]
    \begin{center}
    \begin{tikzpicture}[node distance={15mm}, thick,
    main/.style = {draw, circle,minimum size=11mm},
    comp/.style = {draw, circle,minimum size=9mm},
    labs/.style = {}
    ] 
    \node[labs] (a) {$G_{\text{encoder}}$};
    \node[main] (x1) [below right of=a] {$X_1$}; 
    \node[main] (x2) [right of=x1] {$X_2$};
    \node[main] (x3) [right of=x2] {$X_3$}; 
    \node[main] (y) [right of=x3] {$Y$}; 
    \node[comp] (z1) [below of=x1] {$Z_1$};
    \node[comp] (z2) [below of=x2] {$Z_2$};
    \node[comp] (z3) [below of=x3] {$Z_3$};
    \node[comp] (z) [below of=y] {$Z$};

    \draw[->] (x1) -- (z1);
    \draw[->] (x2) -- (z2);
    \draw[->] (x3) -- (z3);
    \draw[->] (x1) -- (z);
    \draw[->] (x2) -- (z);
    \draw[->] (x3) -- (z);
    
    \node[labs] (b) [above right of=y] {$G_{\text{decoder}}$};
    \node[main] (mx1) [below right of=b] {$X_1$}; 
    \node[main] (mx2) [right of=mx1] {$X_2$};
    \node[main] (mx3) [right of=mx2] {$X_3$}; 
    \node[main] (my) [right of=mx3] {$Y$}; 
    \node[comp] (mz1) [below of=mx1] {$Z_1$};
    \node[comp] (mz2) [below of=mx2] {$Z_2$};
    \node[comp] (mz3) [below of=mx3] {$Z_3$};
    \node[comp] (mz) [below of=my] {$Z$};
    
    \draw[->] (mz) -- (mx1);
    \draw[->] (mz) -- (mx2);
    \draw[->] (mz) -- (mx3);
    \draw[->] (mz) -- (my);
    \draw[->] (mz1) -- (mx1);
    \draw[->] (mz2) -- (mx2);
    \draw[->] (mz3) -- (mx3);
    \end{tikzpicture}
    \end{center}
    \caption{Encoder and decoder graphs for shared and private latent representations SMVIB}
    \label{app:smvib2}
    \end{figure}
    
    An alternative formulation introduces both a shared latent variable $Z$ and private latent variables $Z_1$, $Z_2$, and $Z_3$, corresponding to each view. The shared latent variable $Z$ reconstructs the supervising signal $Y$. In addition, each view $X_i$ is reconstructed from its private latent representation $Z_i$ along with the shared latent variable $Z$. The model is illustrated in Figure~\ref{app:smvib2}, and the corresponding loss function is:
    \begin{multline}
    L_{\text{SMVIB}_2}=\Tilde{I}^{E}((X_1,X_2,X_3);Z)+\Tilde{I}^{E}(X_1;Z_1)+\Tilde{I}^{E}(X_2;Z_2)+\Tilde{I}^{E}(X_3;Z_3)\\
    -\beta (\Tilde{I}^{D}(X_1;(Z,Z_1))+\Tilde{I}^{D}(X_2;(Z,Z_2))+\Tilde{I}^{D}(X_3;(Z,Z_3))+\Tilde{I}^{D}(Z;Y)).
    \end{multline}

    Using the same library of terms as before, this becomes:
    \begin{multline}
    L_{\text{SMVIB}_2}\approx\frac{1}{N}\sum_{i=1}^N D_{\rm KL}(p(z|{x_1}_i,{x_2}_i,{x_3}_i) \Vert r(z))+\sum_{j=1}^3\frac{1}{N}\sum_{i=1}^N D_{\rm KL}(p(z_j|{x_j}_i) \Vert r_j(z))\\
    -\beta\bigg( \frac{1}{N}\sum_{i=1}^N\int dz dz_1 dz_2 dz_3 p(z|{x_1}_i,{x_2}_i,{x_3}_i)p(z_1|{x_1}_i)p(z_2|{x_2}_i)p(z_3|{x_3}_i)\ln(q({x_1}_i|z,z_1))\\
    + \frac{1}{N}\sum_{i=1}^N\int dz dz_1 dz_2 dz_3 p(z|{x_1}_i,{x_2}_i,{x_3}_i)p(z_1|{x_1}_i)p(z_2|{x_2}_i)p(z_3|{x_3}_i)\ln(q({x_2}_i|z,z_2))\\
    +\frac{1}{N}\sum_{i=1}^N\int dz dz_1 dz_2 dz_3 p(z|{x_1}_i,{x_2}_i,{x_3}_i)p(z_1|{x_1}_i)p(z_2|{x_2}_i)p(z_3|{x_3}_i)\ln(q({x_3}_i|z,z_3))\\
    +\frac{1}{N}\sum_{i=1}^N\int dz dy p(z|{x_1}_i,{x_2}_i,{x_3}_i)p(y_i|z) \bigg).
    \end{multline}
\end{itemize}

\subsubsection{Discussion}
There exist many other structures that have been explored in the multi-view representation learning literature, including conditional VIB \citep{shi2019, hwang2021}, which is formulated in terms of conditional information. These types of structures are beyond the current scope of our framework. However, they could be represented by an encoder mapping from all independent views $X_\nu$ to $Z$, subtracted from another encoder mapping from the joint view $\vec{X}$ to $Z$. Coupled with this would be a decoder mapping from $Z$ to the independent views $X_\nu$ (or the joint view $\vec{X}$, analogous to the Joint-DVCCA). Similarly, one can use our framework to represent other multi-view approaches, or their approximations \citep{VanderSchaar2021, Hu2021,hwang2021}. This underscores the breadth of methods seeking to address specific questions by exploring known or assumed statistical dependencies within data, and also the generality of our approach, which can re-derive these methods.

\subsection{Widely used state of the art methods}
\label{more_methods_details}

\subsubsection{Barlow Twins - BT}
For Barlow Twins \citep{zbontar2021}, two different views $X$ and $Y$ are compressed using the same deterministic compression function $\mu(\cdot)$ into $Z_X$ and $Z_Y$ respectively ($p(z_x|x)=\delta(z_x-\mu(x))$ and $p(z_y|y)=\delta(z_y-\mu(y))$). In training, one optimizes the cross-correlation matrix, $C$, between $Z_X$ and $Z_Y$ to be as close to identity as possible. This is equivalent to maximizing the information between $I(Z_X,Z_Y)$ when $Z_X$ and $Z_Y$ are Gaussian. This has the same loss structure as the deterministic symmetric information bottleneck with no reconstruction (Eq~\ref{dsib_no_recon_eq}), but with a shared deterministic encoder for both $Z_X$ and $Z_Y$.
\begin{equation}
    \sloppy L_{\text{BT}}=-\Tilde{I}^{D}(Z_X;Z_Y)=\frac{1}{2}(\ln(\det(I-C C ^T))).
\end{equation}        
This has the same minimum as:
\begin{equation}
    L_{\text{BT}}=\sum_i (1-C_{ii})^2 + \lambda \sum_{i,j}C_{ij}^2.
\end{equation}
        
\subsubsection{Contrastive Language-Image Pretraining - CLIP}
Contrastive Language-Image Pretraining \citep{radford2021learning} was introduced to learn visual concepts from natural language supervision. CLIP simultaneously compresses both the text and the images into a common latent space by maximizing a similarity metric between the compression found for the image and the text. Crucially, it also penalizes the similarity between samples of images and text that are not paired with one another. The CLIP loss is
\begin{equation}L_{\text{CLIP}}=\frac{-1}{2N}\sum_{i=1}^N\ln\left(\frac{\exp{({\vec{z}_{x_i}}\cdot{\vec{z}_{y_i}}/T)}}{\frac{1}{N}\sum_{j=1}^N\exp{({\vec{z}_{x_i}}\cdot{\vec{z}_{y_j}}/T)}}\right)+\frac{-1}{2N}\sum_{i=1}^N\ln\left(\frac{\exp{({\vec{z}_{x_i}}\cdot{\vec{z}_{y_i}}/T)}}{\frac{1}{N}\sum_{j=1}^N\exp{({\vec{z}_{x_j}}\cdot{\vec{z}_{y_i}}/T)}}\right)
\label{eq:cliploss}
\end{equation}
where ${\vec{z}_{x_i}}=\vec\mu_{Z_X}(x_i)$, ${\vec{z}_{y_i}}=\vec\mu_{Z_Y}(y_i)$ and the functions $\vec\mu_{Z_X}$ and $\vec\mu_{Z_Y}$ are parameterized by networks.

We show now that $L_{\rm CLIP}$ is related to the Symmetric Information Bottleneck, but with deterministic encoders for both $Z_X$ and $Z_Y$ and in the $\beta\rightarrow\infty$ limit. Specifically, CLIP's compression of the views $X$ and $Y$ amounts to using deterministic encoder functions into $Z_X$ and $Z_Y$, respectively, such that $p(z_x|x)=\delta(\vec z_x-\vec\mu_{Z_X}(x))$ and $p(z_y|y)=\delta(\vec z_y-\vec\mu_{Z_Y}(y))$. Below we show that, if one represents $p(z_x,z_y)=\frac{\exp{({\vec{z_x}}\cdot\vec{z_y}/T)}}{Z}$, then, in the limit $N\to\infty$,  $L_{\text{CLIP}}\to-I(Z_X;Z_Y)+\mathrm{correction}$, where the correction term is defined below and both it and $I$ are estimated from samples. Further, if one instead interprets  $p(z_x|z_y)=p(z_y|z_x)=\frac{\exp{({\vec{z_x}}\cdot\vec{z_y}/T)}}{Z}$, then  $L_{\text{CLIP}}\to-I(Z_X;Z_Y)$ in the same $N\to\infty$ limit. In both of these limits,  $L_{\rm CLIP}$ has the same structure as the deterministic symmetric information bottleneck with no reconstruction, Eq.~(\ref{dsib_no_recon_eq}), but in the first case, there is an extra correction term.

In the expressions below, we use the limit sign, $\to$, to indicate convergence for $N\to\infty$. That is, by using the limit, we replace the average over a sample with the expectation value. We first start with the case of  $p(z_x,z_y)=\frac{\exp{({\vec{z_x}}\cdot\vec{z_y}/T)}}{Z}$. Then:
\begin{align}
\sloppy L_{\text{CLIP}}&=\frac{-1}{2N}\sum_{i=1}^N\ln\left(\frac{\exp{({\vec{z}_{x_i}}\cdot{\vec{z}_{y_i}}/T)}}{\frac{1}{N}\sum_{j=1}^N\exp{({\vec{z}_{x_i}}\cdot{\vec{z}_{y_j}}/T)}}\right)+\frac{-1}{2N}\sum_{i=1}^N\ln\left(\frac{\exp{({\vec{z}_{x_i}}\cdot{\vec{z}_{y_i}}/T)}}{\frac{1}{N}\sum_{j=1}^N\exp{({\vec{z}_{x_j}}\cdot{\vec{z}_{y_i}}/T)}}\right)\nonumber\\
&=\frac{-1}{2N}\sum_{i=1}^N\ln\left(\frac{p(z_{x_i},{z_{y_i})}}{\frac{1}{N}\sum_{j=1}^N p({z_{x_i},{z_{y_j}}})}\right)+\frac{-1}{2N}\sum_{i=1}^N\ln\left(\frac{p(z_{x_i},z_{y_i})}{\frac{1}{N}\sum_{j=1}^Np(z_{x_j},z_{y_i})}\right)\nonumber\\
&\to\frac{-1}{2N}\sum_{i=1}^N\ln\left(\frac{p(z_{x_i},z_{y_i})}{\int p(z_y)p(z_{x_i},z_y)dz_y}\right)+\frac{-1}{2N}\sum_{i=1}^N\ln\left(\frac{p(z_{x_i},z_{y_i})}{\int p(z_x)p(z_x,z_{y_i})dz_x}\right)\nonumber\\
&=\frac{-1}{N}\sum_{i=1}^N\ln\left(\frac{p(z_{x_i},z_{y_i})}{\sqrt{\int p(z_x)p(z_x,z_{y_i})dz_x\int p(z_y)p(z_{x_i},z_y)dz_y}}\right)\nonumber\\
&=\frac{-1}{N}\sum_{i=1}^N\ln\left(\frac{p(z_{x_i},z_{y_i})}{p(z_{x_i})p(z_{y_i})}\right)-\frac{1}{N}\sum_{i=1}^N\ln\left(p(z_{x_i})p(z_{y_i})\right)\nonumber\\
&\quad +\frac{1}{N}\sum_{i=1}^N\ln\left(\sqrt{\int p(z_x)p(z_x,z_{y_i})dz_x\int p(z_y)p(z_{x_i},z_y)dz_y}\right)\nonumber\\
&\to-I(Z_X,Z_Y)+\int p(z_x,z_y) \ln\left(\frac{\sqrt{\int p(z_x^\prime)p(z_x^\prime,z_{y})dz_x^\prime\int p(z_y^\prime)p(z_{x},z_y^\prime)dz_y^\prime}}{p(z_{x})p(z_{y})}\right)dz_x dz_y,
\end{align}
proving the assertion.

Now consider the case $p(z_y|z_x)=p(z_x|z_y)=\frac{\exp{({\vec{z}_x}\cdot\vec{z}_y/T)}}{Z}$. Then:
\begin{align}
\sloppy L_{\text{CLIP}}&=\frac{-1}{2N}\sum_{i=1}^N\ln\left(\frac{\exp{({\vec{z}_{x_i}}\cdot\vec{z}_{y_i}/T)}}{\frac{1}{N}\sum_{j=1}^N\exp{({\vec{z}_{x_i}}\cdot\vec{z}_{y_j}/T)}}\right)+\frac{-1}{2N}\sum_{i=1}^N\ln\left(\frac{\exp{({\vec{z}_{x_i}}\cdot\vec{z}_{y_i}/T)}}{\frac{1}{N}\sum_{j=1}^N\exp{({\vec{z}_{x_j}}\cdot\vec{z}_{y_i}/T)}}\right)\nonumber\\
&=\frac{-1}{2N}\sum_{i=1}^N\ln\left(\frac{p({z_{x_i}}|{z_{y_i}})}{\frac{1}{N}\sum_{j=1}^N p({z_{x_i}}|{z_{y_j}})}\right)+\frac{-1}{2N}\sum_{i=1}^N\ln\left(\frac{p({z_{y_i}}|{z_{x_i}})}{\frac{1}{N}\sum_{j=1}^Np({z_{y_i}}|{z_{x_j}})}\right)\nonumber\\
&\to\frac{-1}{2N}\sum_{i=1}^N\ln\left(\frac{p(z_{x_i}|z_{y_i})}{\int p(z_{x_i}|{z_y})p(z_y)dz_y}\right)+\frac{-1}{2N}\sum_{i=1}^N\ln\left(\frac{p(z_{y_i}|z_{x_i})}{\int p(z_{y_i}|{z_x})p(z_x)dz_x}\right)\nonumber\\
&=\frac{-1}{2N}\sum_{i=1}^N\ln\left(\frac{p(z_{x_i}|z_{y_i})}{ p(z_{x_i})}\right)+\frac{-1}{2N}\sum_{i=1}^N\ln\left(\frac{p(z_{y_i}|z_{x_i})}{ p(z_{y_i})}\right)\nonumber\\
&\to\frac{-1}{2}I(Z_X;Z_Y)+\frac{-1}{2}I(Z_X;Z_Y) = -I(Z_X;Z_Y),
\end{align}
which completes the proof.

We end this section by discussing these two interpretations of the CLIP loss function (the original work by \cite{radford2021learning} did not offer an interpretation).

For the first interpretation, we note that combining mutual information and the correction, which is a nonlinear combination of probabilities under the log, is an unconventional choice. One cannot argue with the exceptional success of CLIP on real-world data, and yet one must question whether simplifying the loss by subtracting the empirically sampled version of the correction term from Eq.~(\ref{eq:cliploss}) would lead to even better results. 

For the second interpretation, we note that the assumption $p(z_x|z_y)=p(z_y|z_x)$ is very restrictive. It amounts to requiring either (i) a near-identity mapping between $z_x$ and $z_y$, or (ii) that $p(z_x)=p(z_y)$ for all admissible values of $z_x$ and $z_y$. The latter case is equivalent to saying that the marginal distributions are uniform, and yet the joint is not. 

In view of these arguments, it would be interesting to explore if replacing the CLIP loss function with an empirically sampled version of the mutual information, $I(Z_X;Z_Y)$---that is, making CLIP exactly an SIB problem---would result in better performance. It would also be interesting to explore whether the embeddings learned by CLIP on real-world data result in satisfying $p(z_x|z_y)=p(z_y|z_x)$. More generally, it is interesting to explore whether there is another interpretation of $\frac{\exp{({\vec{z}_x}\cdot\vec{z}_y/T)}}{Z}$ that would result in a more meaningful limit of the CLIP loss function $L_{\rm CLIP}$ for $N\to\infty$.

\section{Additional MNIST Results}
\label{App:results-mnist}

In this section, we present supplementary results derived from the methods in Tables~\ref{table:methods}\&\ref{table:methods-private}.

\subsection{Additional results tables for the best parameters}
We report classification accuracy using SVM on data $X$ (SVM on $Y$ is in the main text Tbl.~\ref{table:SVM-Y}), and using neural networks on both $X$ and $Y$.
\begin{table}[h!]
\caption{Maximum  accuracy from a linear SVM and the optimal $k_Z$ and $\beta$ for variational DR methods on the $X$ dataset. ($^\dag$ fixed values)}
\label{table:SVM-X}
\begin{center}
\begin{tabular}{|l|c|c|c|c|c|c|}
\hline
\textbf{Method} & \textbf{Acc. \%} & \textbf{${k_Z}_\textbf{best}$} & $95\%$ \textbf{${k_Z}_\text{range}$} & \bm{$\beta_\text{best}$} & $95\%$ \textbf{$\beta_\text{range}$} \\
\hline
Baseline & 57.8 & 784$^\dag$ & - & - & - \\
PCA & 58.0 & 256 & [32,265*] & - & -  \\
CCA & 54.4 & 256 & [8,265*] & - & -  \\
$\beta$-VAE & 84.4 & 256 & [128,265*] & 4 & [2,8] \\
DVIB & 87.3 & 128 & [4,265*] & 512 & [8,1024*] \\
DVCCA & 86.1 & 256 & [64,265*] & 1$^\dag$ & -  \\
$\beta$-DVCCA & 88.9 & 256 & [128,265*] & 4 & [1,128]  \\
DVCCA-private & 85.3 & 128 & [32,265*] & 1$^\dag$ & -  \\
$\beta$-DVCCA-private & 85.3 & 128 & [32,265*] & 1 & [1,8]  \\
DVSIB & \textbf{92.9} & 256 & [64,265*] & 256 & [4,1024*] \\
DVSIB-private & \textbf{92.6} & 256 & [\textbf{32},265*] & 128 & [8,1024*] \\
\hline

\end{tabular}
\end{center}
\end{table}

\begin{table}[h!]
\caption{Maximum accuracy from a feed forward neural network and the optimal $k_Z$ and $\beta$ for variational DR methods on the $Y$ and the joined $[X,Y]$ datasets. ($^\dag$ fixed values)}
\label{table:NN-Y}
\begin{center}
\begin{tabular}{|l|c|c|c|c|c|}
\hline
\textbf{Method} & \textbf{Acc. \%} & \textbf{${k_Z}_\textbf{best}$} & $95\%$ \textbf{${k_Z}_\text{range}$} & \bm{$\beta_\text{best}$} & $95\%$ \textbf{$\beta_\text{range}$} \\
\hline
Baseline & 92.8 & 784$^\dag$ & - & - & - \\
PCA & 97.6 & 128 & [16,256*] & - & - \\
CCA & 90.2 & 256 & [32,256*] & - & - \\
$\beta$-VAE & \textbf{98.4} & 64 & [8,256*] & 64 & [2,1024*] \\
DVIB & 90.4 & 128 & [8,256*] & 1024 & [8,1024*] \\
DVCCA & 91.3 & 16 & [4,256*] & 1$^\dag$  & - \\
$\beta$-DVCCA & 97.5 & 128 & [8,256*] & 512 & [2,1024*] \\
DVCCA-private & 93.8 & 16 & [2,256*] & 1$^\dag$ & - \\
$\beta$-DVCCA-private & 97.5 & 256 & [\textbf{2},256*] & 32 & [1,1024*] \\
DVSIB & \textbf{98.3} & 256 & [\textbf{4},256*] & 32 & [2,1024*] \\
DVSIB-private & \textbf{98.3} & 256 & [\textbf{4},256*] & 32 & [2,1024*] \\
\hline
Baseline-joint & 97.7 & 1568$^\dag$ & - & - & - \\
joint-DVCCA & 93.7 & 256 & [8,256*] & 1$^\dag$ & - \\
$\beta$-joint-DVCCA & \textbf{98.9} & 64 & [8,256*] & 512 & [2,1024*] \\
joint-DVCCA-private & 93.5 & 16 & [4,256*] & 1$^\dag$ & - \\
$\beta$-joint-DVCCA-private & 95.6 & 32 & [4,256*] & 512 & [1,1024*] \\
\hline
\end{tabular}
\end{center}
\end{table}

\begin{table}[h!]
\caption{Maximum  accuracy from a neural network  the optimal $k_Z$ and $\beta$ for variational DR methods on the $X$  dataset. ($^\dag$ fixed values)}
\label{table:NN-X}
\begin{center}
\begin{tabular}{|l|c|c|c|c|c|}
\hline
\textbf{Method} & \textbf{Acc. \%} & \textbf{${k_Z}_\textbf{best}$} & $95\%$ \textbf{${k_Z}_\text{range}$} & \bm{$\beta_\text{best}$} & $95\%$ \textbf{$\beta_\text{range}$} \\
\hline
Baseline & 92.8 & 784$^\dag$ & - & - & - \\
PCA & 91.9 & 64 & [32,256*] & -  & - \\
CCA & 72.6 & 256 & [256,256*] & -  & - \\
$\beta$-VAE & 93.3 & 256 & [16,256*] & 256 & [2,1024*] \\
DVIB & 87.5 & 4 & [2,256*] & 1024 & [4,1024*] \\
DVCCA & 87.5 & 128 & [8,256*] & 1$^\dag$ & - \\
$\beta$-DVCCA & 92.2 & 64 & [8,256*] & 32 & [2,1024*] \\
DVCCA-private & 88.2 & 8 & [8,256*] & 1$^\dag$ & - \\
$\beta$-DVCCA-private & 90.7 & 256 & [4,256*] & 8 & [1,1024*] \\
DVSIB & \textbf{93.9} & 128 & [\textbf{8},256*] & 16 & [2,1024*] \\
DVSIB-private & 92.8 & 32 & [8,256*] & 256 & [4,1024*] \\
\hline
\end{tabular}
\end{center}
\end{table}
\newpage

\subsection{t-SNE embeddings at best parameters}
\label{App:best-latent-mnist}
\begin{figure}[h!]
\begin{center}
\includegraphics[width=.4\textwidth]{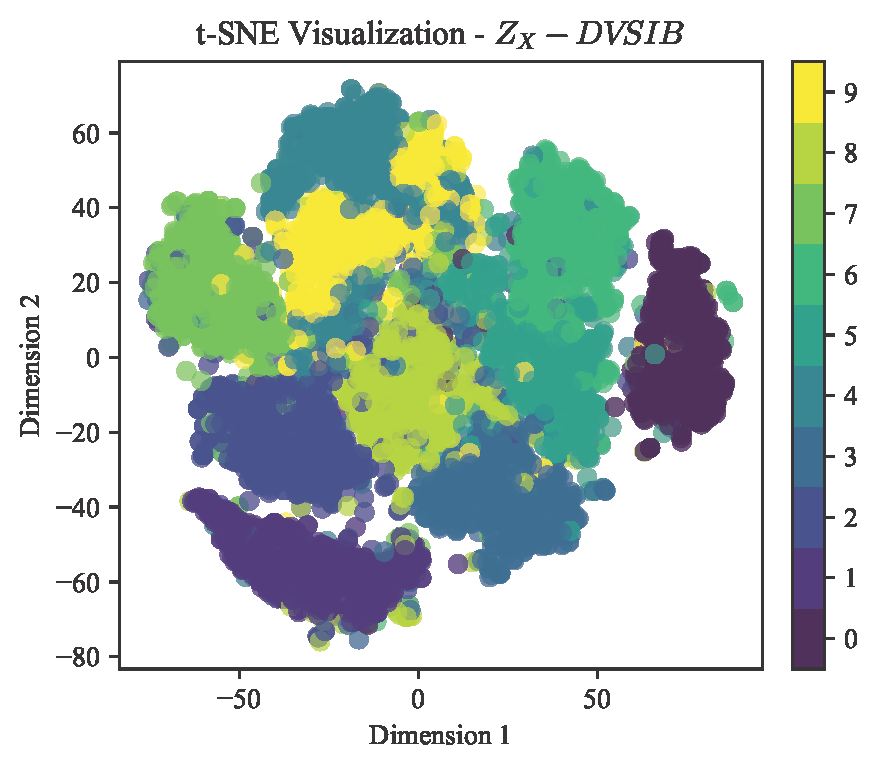}
\includegraphics[width=.4\textwidth]{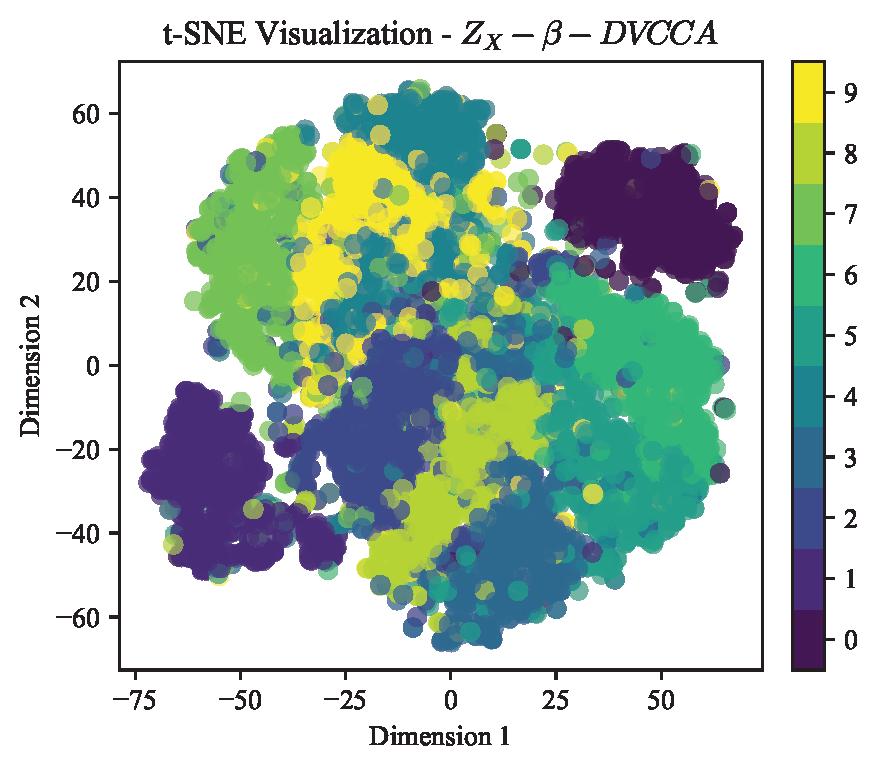}
\end{center}
\begin{center}
\includegraphics[width=.4\textwidth]{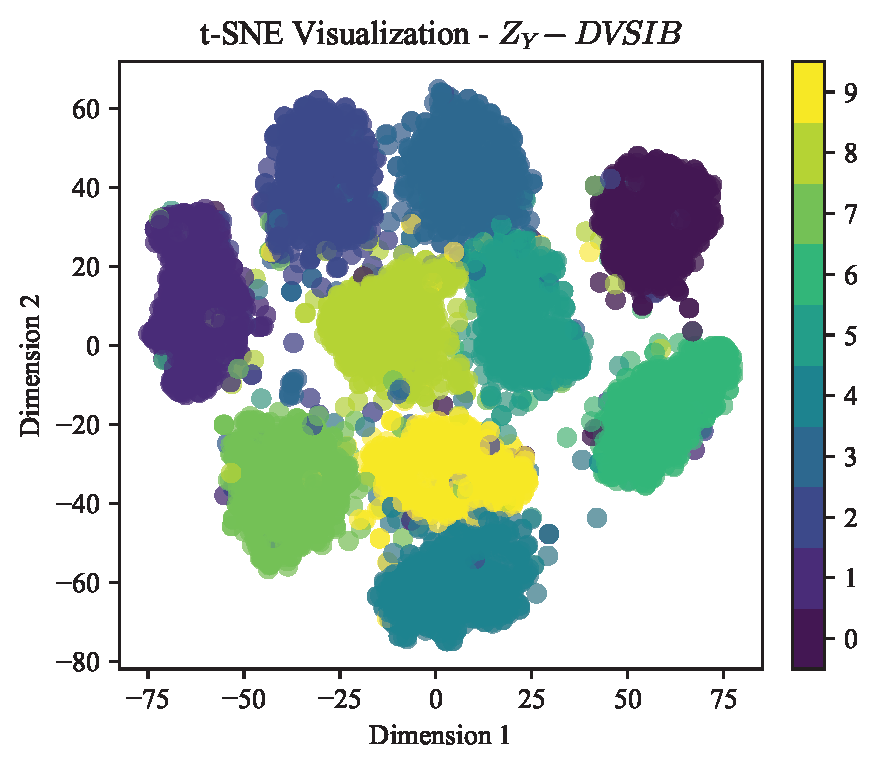}
\includegraphics[width=.4\textwidth]{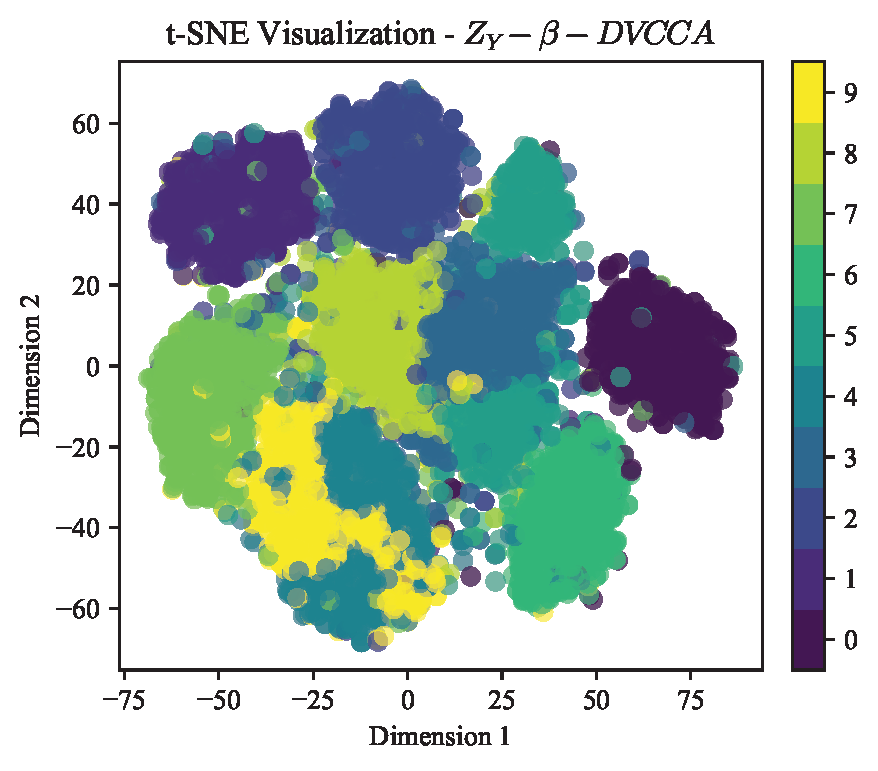}
\end{center}

\caption{Clustering of embeddings for best $\beta$, $k_{Z_X}$ (top) $k_{Z_Y}$ (bottom) for DVSIB and $\beta$-DVCCA.}
\label{Fig:TSNEbestz}
\end{figure}

Figure \ref{Fig:TSNEbestz} display 2d t-SNE embeddings for variables $Z_X$ and $Z_Y$ generated by various considered DR methods.

\subsection{t-SNE embeddings at \texorpdfstring{$k_{Z_X}=k_{Z_Y}=2$}{kZx=kZy=2}}
\label{App:2latent-mnist}
\begin{figure}[h!]
\begin{center}
\includegraphics[width=.8\textwidth]{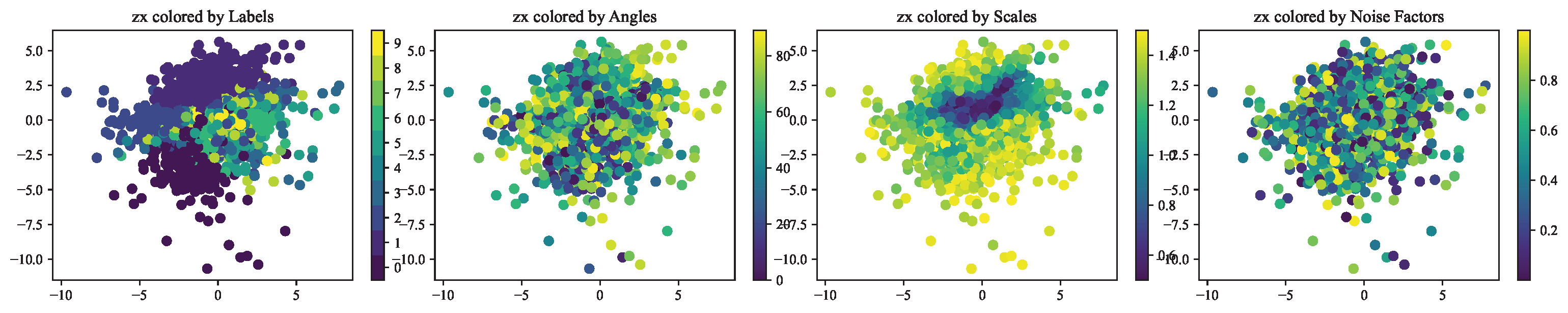}
\includegraphics[width=.8\textwidth]{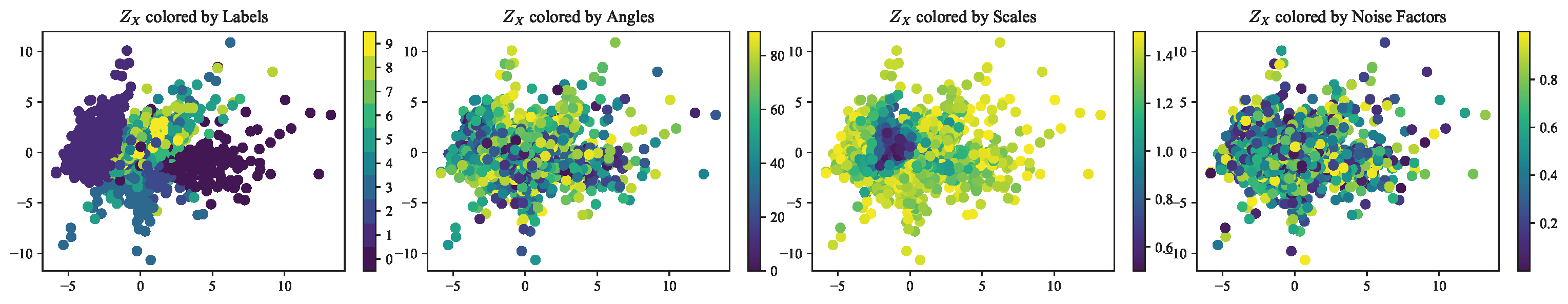}
\includegraphics[width=.8\textwidth]{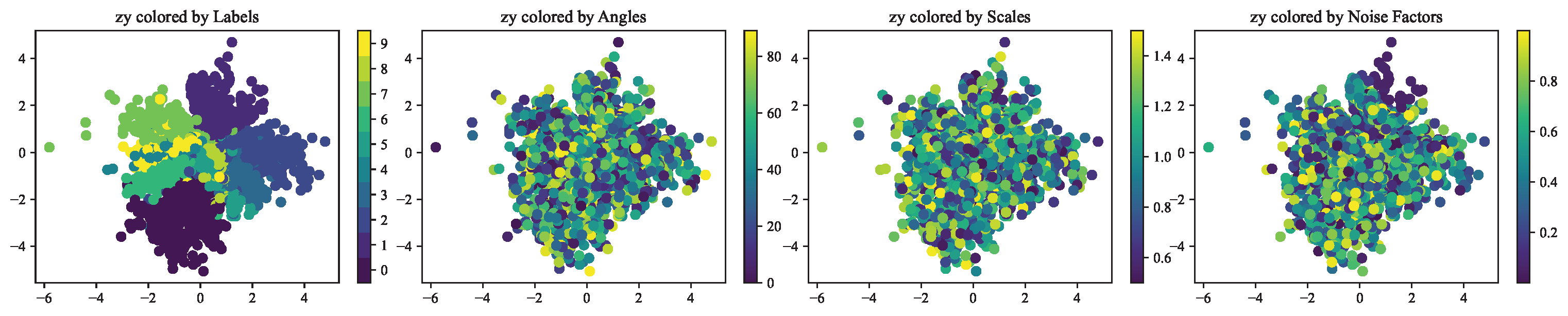}
\includegraphics[width=.8\textwidth]{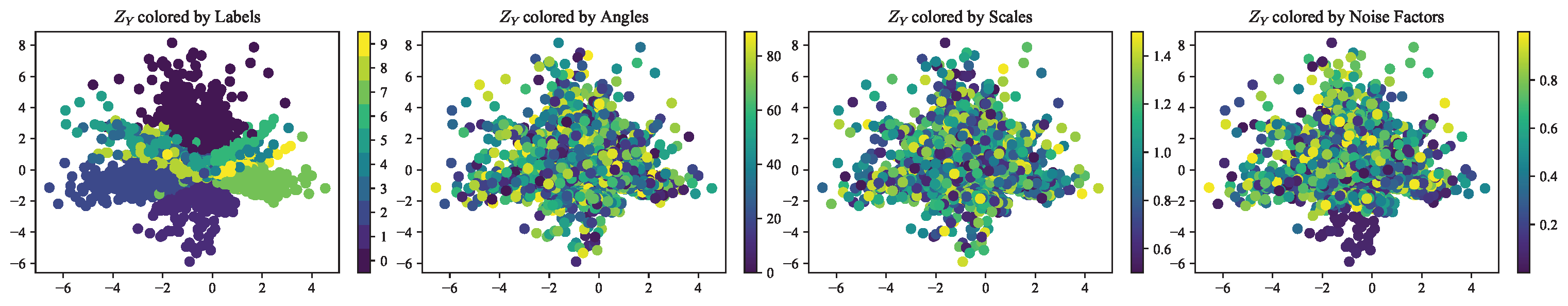}

\caption{Clustering of embeddings when restricting $k_Z=2$ for DVSIB and $\beta$-DVCCA colored by labels, rotations, scales, and noise factors, respectively. The top two rows show results for the $X$ dataset, while the bottom two rows correspond to the $Y$ dataset.}
\label{Fig:dz2}
\end{center}
\end{figure}

We now demonstrate how different DR methods behave when the compressed variables are restricted to have not more than 2 dimensions, cf.~Fig.~\ref{Fig:dz2}.

\subsection{DVSIB-private embeddings}
\subsubsection{At best parameters}

\begin{figure}[h!]
\begin{center}
\includegraphics[width=.4\textwidth]{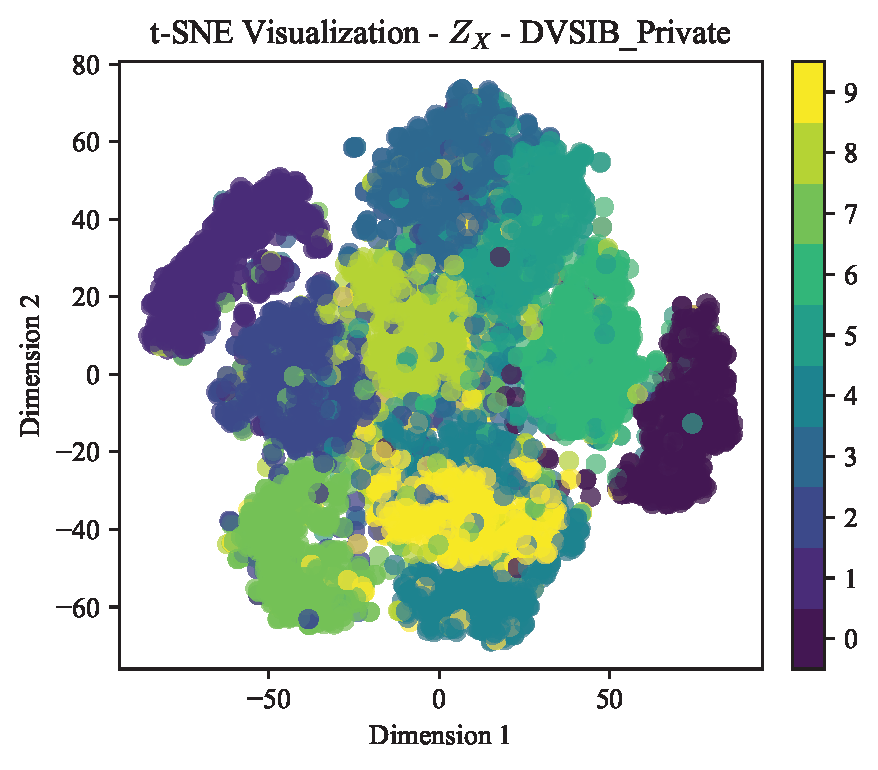}
\includegraphics[width=.4\textwidth]{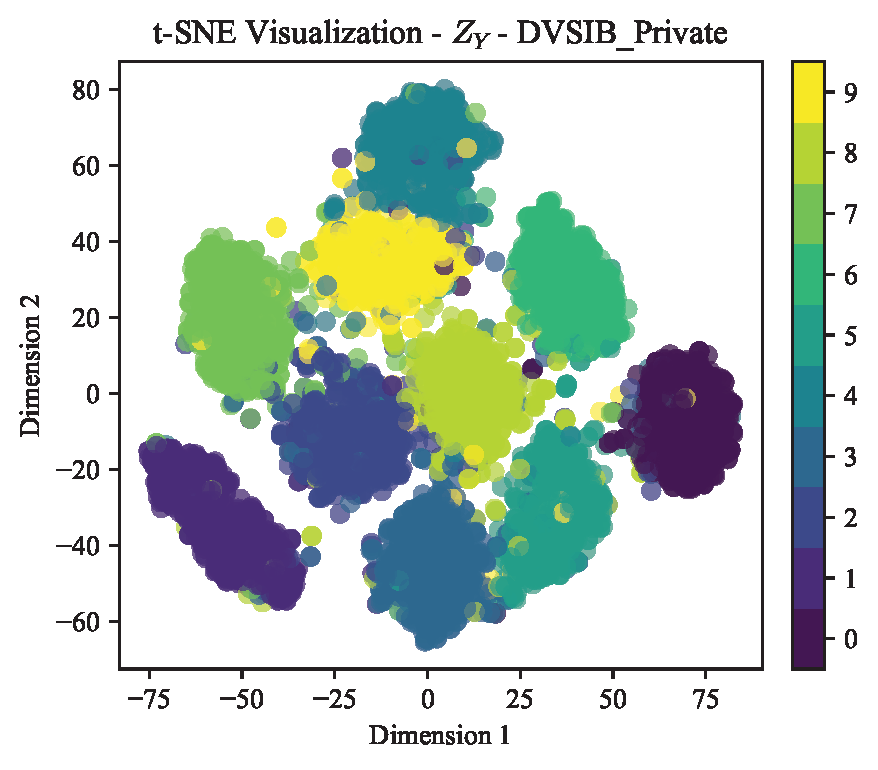}
\includegraphics[width=.8\textwidth]{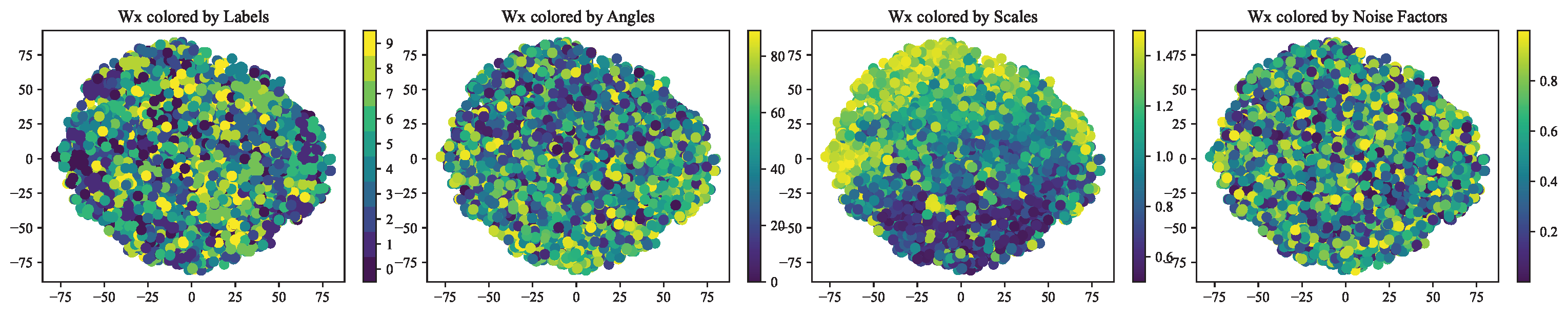}
\includegraphics[width=.8\textwidth]{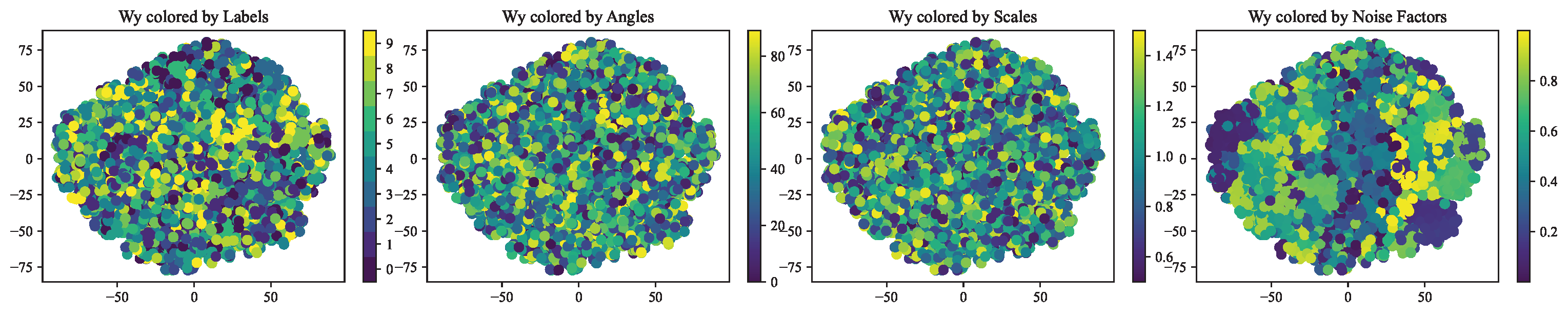}
\includegraphics[width=.8\textwidth]{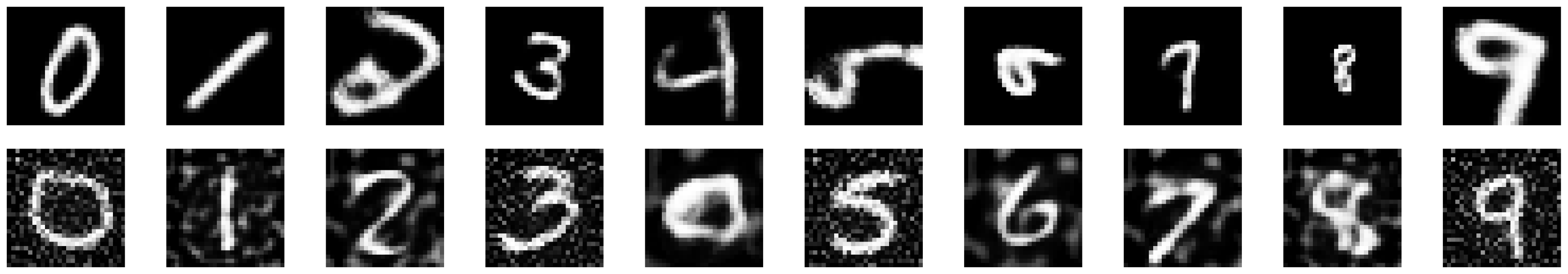}
\end{center}
\caption{DVSIB-private t-SNE embeddings at best $k_Z$ and $\beta$: Top row shows $Z_X$ (left) and $Z_Y$ (right). Second and third rows display $W_X$ and $W_Y$, colored by labels, rotations, scales, and noise factors, respectively. Bottom row presents digit reconstructions using both shared and private information, demonstrating how private components capture variations in background, scaling, and rotation.}
\label{Fig:TSNE-dvsib-p-best}
\end{figure}

Figure~\ref{Fig:TSNE-dvsib-p-best} shows the t-SNE embeddings of the private latent variables constructed by DVSIB-private, colored by the digit label. To the extent that the labels do not cluster, private latent variables do not preserve the label information shared between $X$ and $Y$.

\subsubsection{At \texorpdfstring{$k_{Z_X}=k_{Z_Y}=2$}{kZx=kZy=2}}

\begin{figure}[h!]
\begin{center}
\includegraphics[width=.8\textwidth]{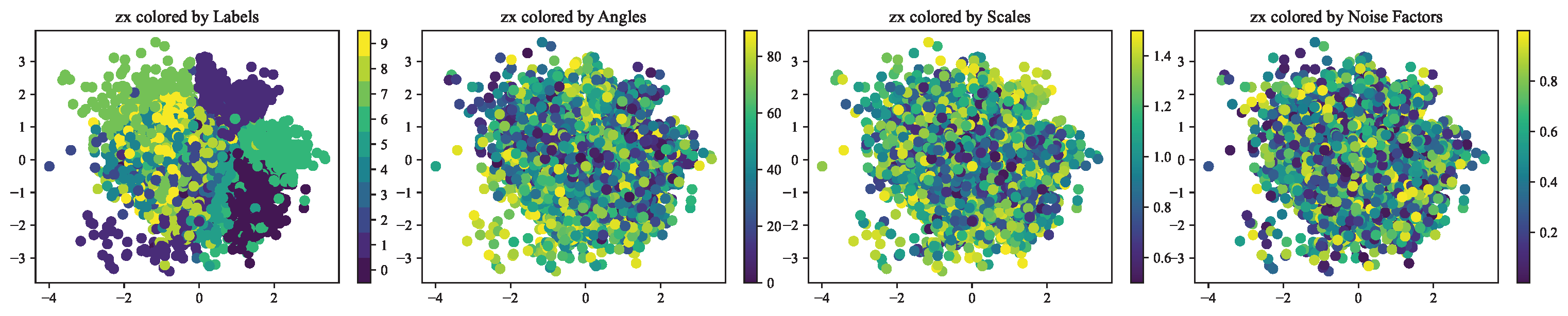}
\includegraphics[width=.8\textwidth]{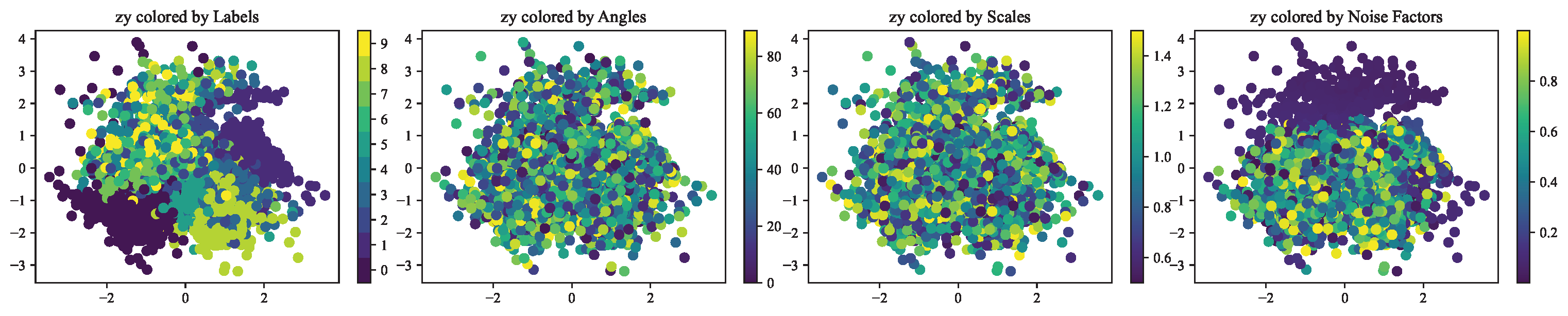}
\includegraphics[width=.8\textwidth]{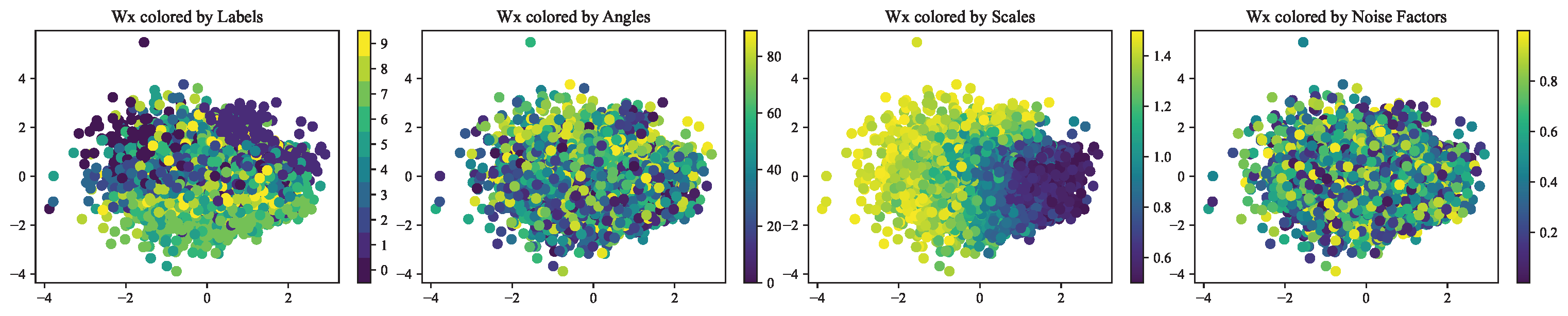}
\includegraphics[width=.8\textwidth]{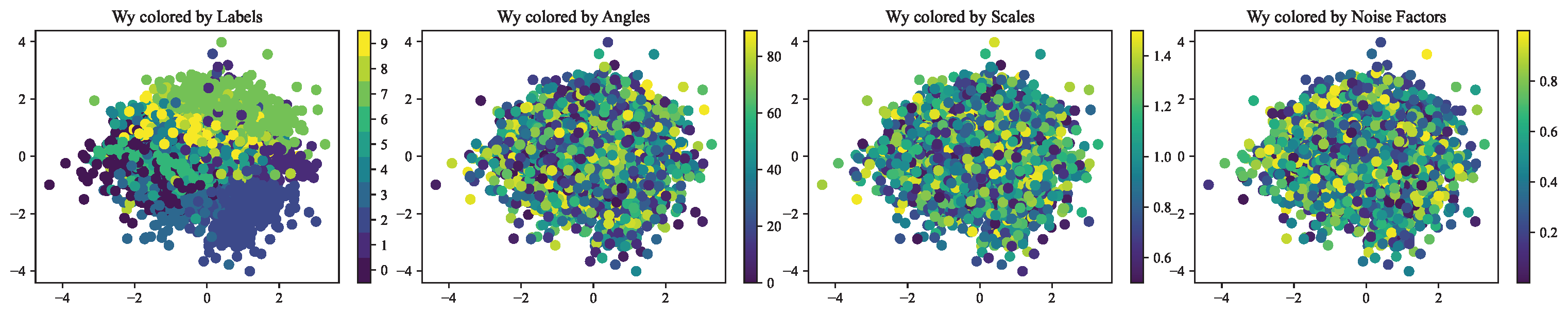}
\end{center}
\caption{DVSIB-private embeddings: The first two rows correspond to $Z_X$ and $Z_Y$, respectively, while the third and fourth rows show $W_X$ and $W_Y$. Each row is colored by labels, rotations, scales, and noise factors, respectively.}
\label{Fig:dvsibp-dz2w}
\end{figure}

Figure~\ref{Fig:dvsibp-dz2w} shows the embeddings constructed by DVSIB-private, colored by the digit label, rotations, scales, and noise factors for $X$ and $Y$. Private latent variables at 2 latent dimensions preserve a little about the label information shared between $X$ and $Y$, but clearly preserve the scale information for $X$, even at only two latent dimensions.

\subsection{Testing Training Efficiency}
\label{App:subsamples}
\begin{figure}[h!]
\begin{center}
\includegraphics[width=\textwidth]{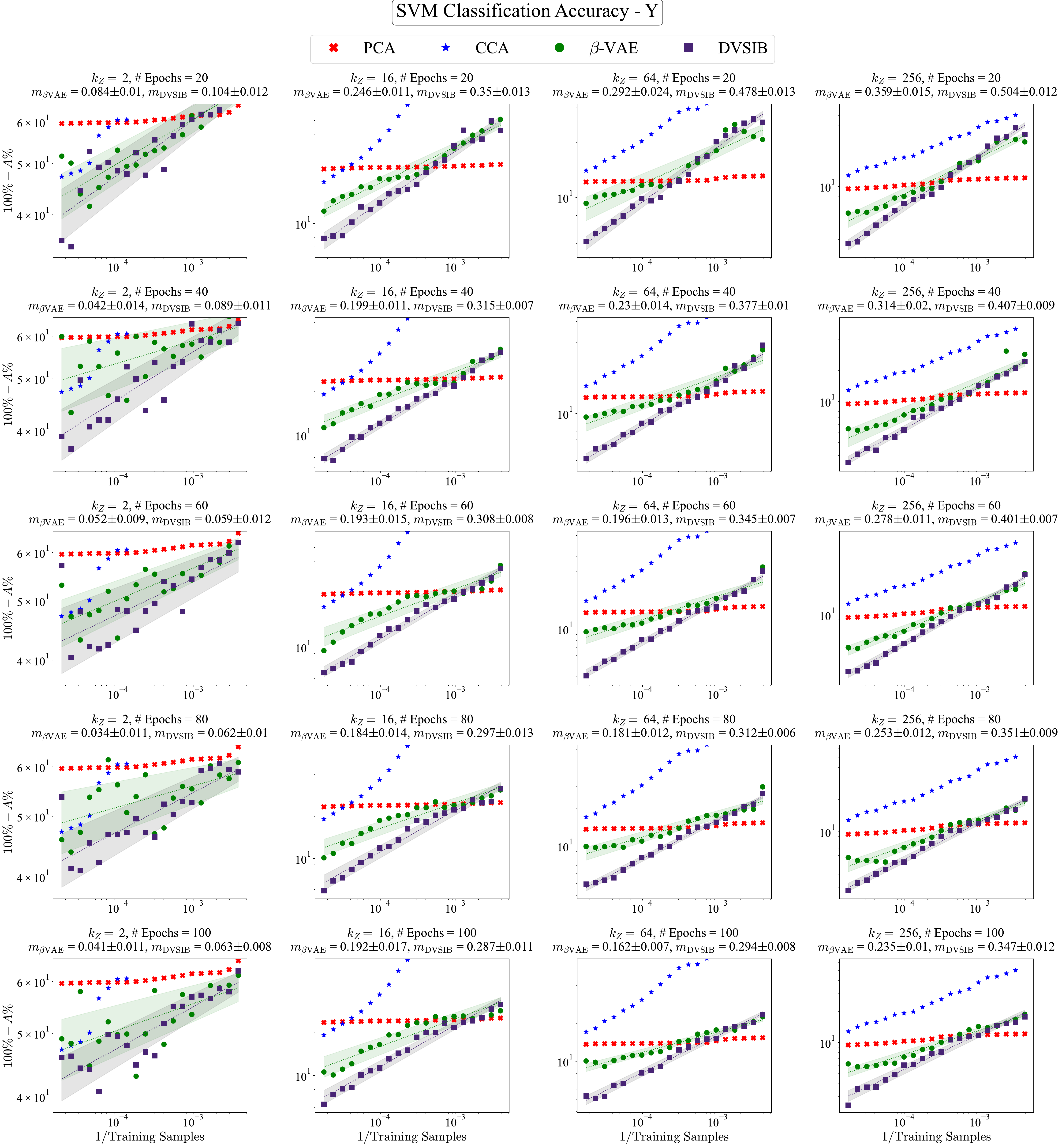}
\end{center}
\caption{Log-log plot of $100 -A$ vs $1/n$. DVSIB has a steeper slope than $\beta$-VAE corresponding to faster convergence with fewer samples for DVSIB. Plots vary $k_Z=2,16,64,256$ and training epochs $20,40,60,80,100$.
\label{Fig:Acc_T_SVM_all}}
\end{figure}

We tested an SVM's classification accuracy for distinguishing digits based on latent subspaces created by DVSIB, $\beta$-VAE, CCA, and PCA trained using different amounts of samples. Figure~\ref{Fig:Acc_T_SVM-Y} in the main text shows the results for 60 epochs of training with latent spaces of dimension $k_{Z_X}=k_{Z_Y}=64$. The DVSIB and $\beta$-VAE were trained with $\beta=1024$. Figure~\ref{Fig:Acc_T_SVM_all} shows the SVM's classification accuracy for a range of latent dimensions (from right to left): $k_{Z_X}=k_{Z_Y}=2, 16, 64, 256$. Additionally, it shows the results for different amounts of training time for the encoders ranging from 20 epochs (top row) to 100 epochs (bottom row). As explained in the main text, we plot a log-log graph of $100-A$ versus $1/n$. Plotted in this way, high accuracy appears at the bottom, and large sample sizes are at the left of the plots. DVSIB, $\beta$-VAE, and CCA often appear linear when plotted this way, implying that they follow the form $A=100-c/n^m$. Steeper slopes $m$ on these plots correspond to a faster increase in the accuracy with the sample size. This parameter sweep shows that the tested methods have not had time to fully converge at low epoch numbers. Additionally, increasing the number of latent dimensions helps the SVMs untangle the non-linearities present in the data and improves the corresponding classifiers.

\clearpage
\section{Convolutional  DVSIB}
\label{app:cifar}
\subsection{Noisy CIFAR-100 dataset}
\begin{figure}[h!]
    \centering
    \includegraphics[width=\textwidth]{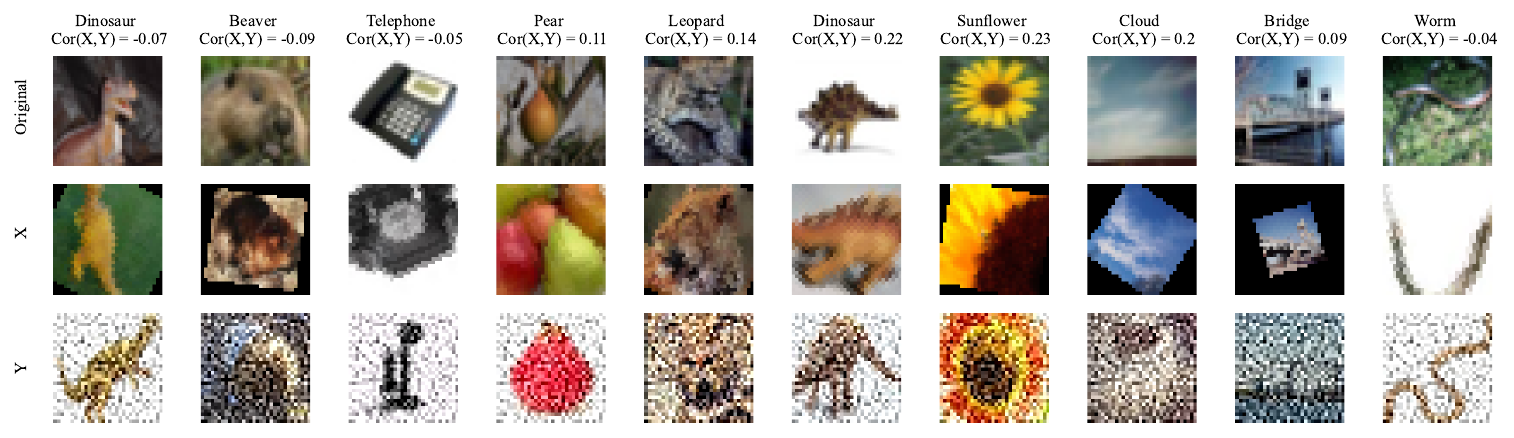}
    \caption{Examples from the Noisy CIFAR-100 dataset. Each column shows instances from a random category from CIFAR-100. For each category, we show an example of an unperturbed image of that class (top), a rotated and scaled example image from the class $X$ (middle), and a noised example image from the class with Perlin background noise $Y$ (bottom). The correlation between the $X$ and $Y$ views is calculated for the shown pairs.}
    \label{Fig:data_cifar}
\end{figure}
We evaluate different DR methods on an adaptation of the CIFAR-100 dataset \citep{Krizhevsky2009learning}. CIFAR-100 consists of $60,000$ color images in $100$ classes, with $600$ images per class. Each image is of size $32 \times 32$ pixels and has $3$ color channels (RGB), corresponding to a total number of $3072$ pixels.

To create the Noisy CIFAR-100 dataset, we generate two distinct views of each image class. The first view ($X$) is the original image, which is randomly rotated by an angle uniformly sampled between $0$ and $\frac{\pi}{2}$ and scaled by a factor uniformly distributed between $0.5$ and $1.5$. The second view ($Y$) is derived from the original image but with an added background Perlin noise \citep{Perlin1985}, where the noise factor is independently sampled for each RGB channel from a uniform distribution between $0$ and $1$. Both views are rescaled to have pixel intensities in the range $[0,1)$.

The dataset was shuffled within labels, retaining only the shared label identity between the two transformed views ($X$ and $Y$). The dataset is partitioned into training ($80\%$), testing ($10\%$), and validation ($10\%$) subsets.

Figure~\ref{Fig:data_cifar} illustrates some representative examples from different categories in the Noisy CIFAR-100 dataset. For each shown category, we show an original image, a transformed image from $X$ (rotated and scaled), and a noisy image from $Y$ (with Perlin noise).

\subsection{Architecture}
Since the Noisy CIFAR-100 dataset is more complex, and to showcase the flexibility of our framework with diverse data modalities, we extend our model to incorporate convolutional neural networks (CNNs) for the encoder and decoder architectures. Unlike the fully connected feed-forward networks used for Noisy MNIST, the convolutional structures are better suited for handling images with multiple channels.

We use convolutional encoders and decoders as detailed in the \ref{conv_enc}~\&~\ref{conv_dec} for the Variational Autoencoder (VAE), $\beta$-DVCCA, DVCCA ($\beta=1$), and DVSIB models. While Conv-VAE is well established in the literature \citep{rezende2014stochastic,pu2016variational,kingma2014semi}, we introduce novel adaptations: Conv-$\beta$-DVCCA and Conv-DVSIB. Additionally, we enhance DVSIB by integrating a clipped version of the MINE estimator, known as SMILE \citep{song2019understanding}. This modified estimator, set with a clipping factor $\tau=5$, is described in detail in the \ref{smile}. The SMILE estimator improves the estimation stability and accuracy of mutual information, leading to enhanced performance of Conv-DVSIB.

\subsubsection{Encoder \texorpdfstring{$I^E(X;Z_X)$}{Ie(X;Zx)}}
\label{conv_enc}

The convolutional encoder architecture is designed to handle the $32 \times 32 \times 3$ image input, using three convolutional layers. The first layer has $64$ filters, followed by batch normalization and ReLU activation, followed by a max-pooling layer to reduce the spatial dimensions. The second convolutional layer has $128$ filters, also followed by batch normalization, ReLU activation, and max-pooling. The final convolutional layer consists of $256$ filters, further reducing the image dimensions with max-pooling. All the convolutional layers have a kernel size of $3 \times 3$ and padding of $1$, with the max-pooling layers having a stride of $2$

The flattened output of the convolutional layers is passed through a fully connected layer of size $1024$ with ReLU activation and dropout of $0.5$ for regularization, then another fully connected linear layer of size $1024$. Then the $1024$ dimensional representation is mapped to two output layers that generate the mean ($\mu$) and log variance ($\log \sigma^2$) for the latent space. The fully connected layers are initialized with Xavier initialization \citep{glorot2010understanding} for stability.

\subsubsection{Decoder \texorpdfstring{$I^D(X;Z_X)$}{Id(X;Zx)}}
\label{conv_dec}
The convolutional decoder mirrors this structure in reverse. Starting from the latent representation, it expands the data through two fully connected layers of size $1024$, reshaping it to a $256 \times 4 \times 4$ tensor. The deconvolutional layers upsample the feature maps back to the original image size using transpose convolutions, with batch normalization and ReLU activation, and a final Sigmoid activation to ensure pixel intensities lie in the range $[0, 1]$.

\subsubsection{Mutual Information Estimator \texorpdfstring{$I_\mathrm{SMILE}(Z_X;Z_Y)$}{Ismile(Zx;Zy)}}\label{smile}

To enhance the mutual information estimation in Conv-DVSIB, we employ the SMILE estimator \citep{song2019understanding}, explained in more detail in Appx.~\ref{app:estimators}.

The SMILE estimator is implemented using a concatenated critic with three hidden layers, each containing 256 neurons. The input layer takes the joint low-dimensional representations from both views, $[Z_X, Z_Y]$, and maps them to a single-neuron output layer. This adaptation significantly stabilizes the learning of mutual information and contributes to improved performance in the Conv-DVSIB model.

\subsection{Results - Tables}

Table \ref{tab:results_cifar_y} summarizes the linear SVM classification accuracy for each method tested: Conv-VAE, Conv-$\beta$-DVCCA, Conv-DVCCA ($\beta=1$), and Conv-DVSIB. Since Noisy CIFAR-100 is a more challenging multi-class dataset compared to Noisy MNIST, and we have \textit{simple} implementations for the encoders and decoders, we report the top-$k$ accuracies for $k = 1, 5, 10, 20$, where top-$k$ refers to the fraction of instances where the true class is among the top-$k$ predicted classes. The table reports the test accuracies, using a separate dataset that was not seen during training or validation. The values of $k_Z$ shown are the ones that gave the highest accuracy for each method after sweeping over different $k_Z$ values. We found that $\beta=1024$ consistently gave the best results for all methods (Sec.~\ref{App:results_cifar_figs}). The results indicate that Conv-DVSIB outperforms all other methods across different top-$k$ accuracies. In addition, our newly improved method Conv-$\beta$-DVCCA clearly outperforms the original Conv-DVCCA keeping all other parameters fixed. Note that this is not meant to compete with the best possible accuracy for the Noisy CIFAR-100 algorithms which is beyond the message of this paper, but rather to show the comparable behavior of all the methods on equal footage. A better comparison moving towards the state-of-the-art level is in Sec.\ref{app:results-sota}.

\subsubsection{SVM on X}
\begin{longtable}{|l||c|c||c|c||c|c||c|c|}
\caption{Linear SVM classification accuracy on Noisy CIFAR (X data). We report the top-$k$ accuracies for $k=1, 5, 10, 20$, along with the corresponding $k_Z$ values that provided the highest accuracies for each method. The methods tested are Conv-VAE, Conv-$\beta$-DVCCA, Conv-DVCCA ($\beta=1$), and Conv-DVSIB. The best results are shown in bold.}
\label{tab:results_cifar_x} \\
\hline
Method & Top 1 & $k_Z$ & Top 5 & $k_Z$ & Top 10 & $k_Z$ & Top 20 & $k_Z$ \\
\hline
\endfirsthead

\hline
Method & Top 1 & $k_Z$ & Top 5 & $k_Z$ & Top 10 & $k_Z$ & Top 20 & $k_Z$ \\
\hline
\endhead

\hline
Conv-VAE & 8.43 & 128 & 24.67 & 128 & 36.97 & 128 & 54.28 & 128 \\
Conv-DVCCA & 0.72 & 2 & 3.72 & 2 & 7.85 & 2 & 16.55 & 2 \\
Conv-$\beta$-DVCCA & 10.65 & 128 & 30.17 & 128 & 42.83 & 128 & 57.83 & 128 \\
Conv-DVSIB & \textbf{14.72} & 128 & \textbf{38.23} & 128 & \textbf{52.58} & 16 & \textbf{68.48} & 16 \\
\hline
\end{longtable}
\subsubsection{SVM on Y}
\begin{longtable}{|l||c|c||c|c||c|c||c|c|}
\caption{Linear SVM classification accuracy on Noisy CIFAR (Y data). We report the top-$k$ accuracies for $k=1, 5, 10, 20$, along with the corresponding $k_Z$ values that provided the highest accuracies for each method. The methods tested are Conv-VAE, Conv-$\beta$-DVCCA, Conv-DVCCA ($\beta=1$), and Conv-DVSIB. The best results are shown in bold.}
\label{tab:results_cifar_y} \\
\hline
Method & Top 1 & $k_Z$ & Top 5 & $k_Z$ & Top 10 & $k_Z$ & Top 20 & $k_Z$ \\
\hline
\endfirsthead

\hline
Method & Top 1 & $k_Z$ & Top 5 & $k_Z$ & Top 10 & $k_Z$ & Top 20 & $k_Z$ \\
\hline
\endhead

\hline
Conv-VAE & 13.93 & 64 & 35.97 & 64 & 49.6 & 64 & 65.25 & 64 \\
Conv-DVCCA & 0.72 & 2 & 3.72 & 2 & 7.85 & 2 & 16.55 & 2 \\
Conv-$\beta$-DVCCA & 15.07 & 128 & 35.63 & 128 & 48.33 & 128 & 64.2 & 64 \\
Conv-DVSIB & \textbf{23.75} & 128 & \textbf{52.15} & 32 & \textbf{65.13} & 32 & \textbf{78.93} & 32 \\
\hline
\end{longtable}

\subsection{Results - Figures}
\label{App:results_cifar_figs}
\begin{figure}[h!]
    \centering
    \includegraphics[width=.9\textwidth]{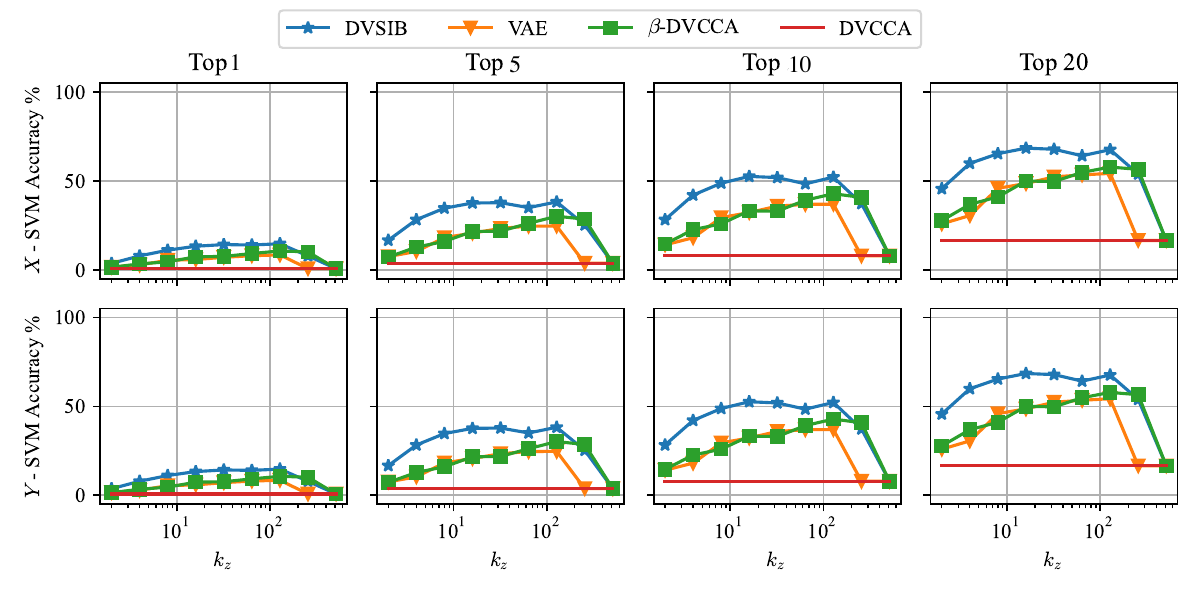}
    \caption{Top-$k$ classification accuracies (Top-1, Top-5, Top-10, Top-20) as a function of $k_Z$ for Noisy CIFAR. The x-axis shows different values of $k_Z$, and $\beta$ is fixed at the highest value tested ($\beta=1024$). The first row shows results for X data, while the second row shows results for Y data. We observe a saturation or decline in accuracy beyond $k_Z=128$.}
    \label{fig:kz_sweep}
    \vspace{-0.3in}
\end{figure}

In the supplementary figures, we further analyze the impact of $k_Z$ and $\beta$ on the performance:

1. Figure \ref{fig:kz_sweep} shows the effect of varying $k_Z$ on top-$k$ accuracy, with $\beta$ value chosen to maximize the reported accuracy. The x-axis represents different values of $k_Z$, while the four columns display the top-1, top-5, top-10, and top-20 accuracies. The first row shows results for the X data, while the second row shows results for the Y data.

\begin{figure}[h!]
    \centering
    \includegraphics[width=.9\textwidth]{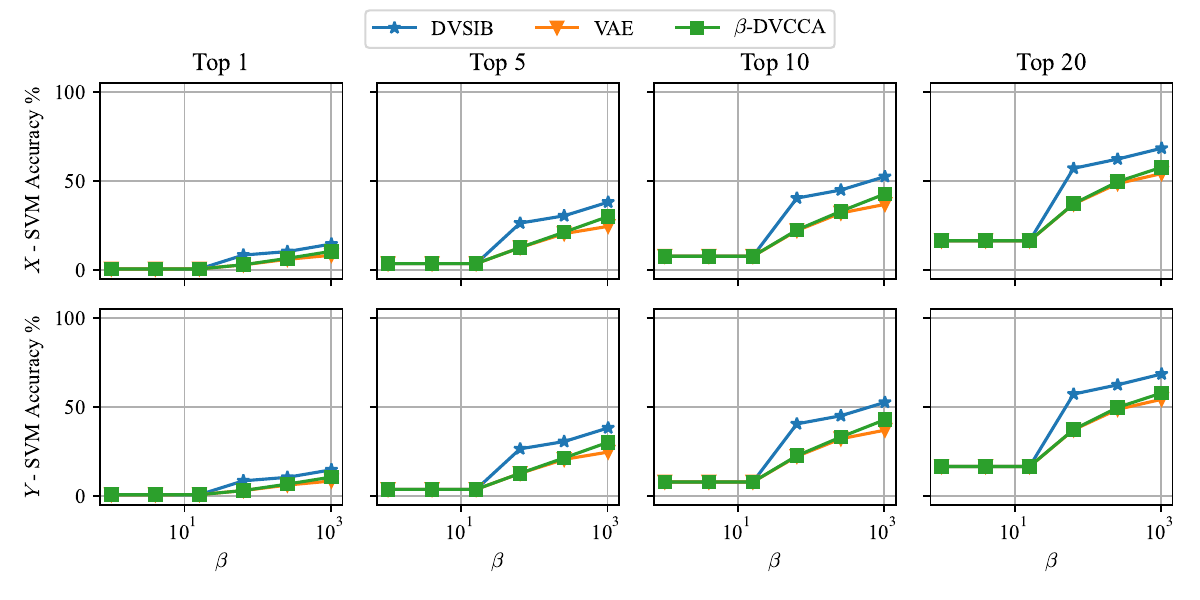}
    \caption{Top-$k$ classification accuracies (Top-1, Top-5, Top-10, Top-20) as a function of $\beta$ for Noisy CIFAR. The x-axis shows different values of $\beta$, with $k_Z$ fixed at the highest value tested ($k_Z=128$). The first row shows results for X data, while the second row shows results for Y data. We notice an increase in accuracy as $\beta$ increases up to $1024$.}
    \label{fig:beta_sweep}
    \vspace{-0.3in}
    \end{figure}

2. Figure \ref{fig:beta_sweep} illustrates the effect of varying $\beta$ with  $k_Z$ value chosen to maximize the reported accuracy. The x-axis represents different values of $\beta$, and the columns correspond to the different top-$k$ accuracies as described above.

\newpage
\subsubsection{\texorpdfstring{$k_{Z}$ sweeps}{kZ sweeps}}

\subsubsection{\texorpdfstring{$\beta$ sweeps}{beta sweeps}}

\clearpage
\section{ResNet-DVSIB}
\label{app:results-sota}
\subsection{Barlow Twins and Mixed Barlow Twins}

Self-supervised learning (SSL) methods such as Barlow Twins \citep{zbontar2021} have demonstrated remarkable performance in representation learning tasks. However, a key limitation of the standard Barlow Twins (BT) approach is its tendency to overfit the training data, particularly when trained on small or moderate-sized datasets. To address this issue, \cite{bandara2023guarding} introduced a mixup-based regularization technique that significantly improves generalization by encouraging better alignment between augmented views of the same data. 

The Barlow Twins objective is formulated as:
\begin{equation}
    \mathcal{L}_{BT} = \sum_{i} (1 - C_{ii})^2 + \lambda \sum_{i} \sum_{j \neq i} C_{ij}^2,
\end{equation}
where $C$ is the cross-correlation matrix of the normalized embeddings from two different augmentations of the same image, and $\lambda$ controls the weight of the off-diagonal terms.

To further improve generalization, \cite{bandara2023guarding} proposed a mixup-based regularization. Given an image batch $\mathbf{x}$, they generate two augmentations $\mathbf{y}_a$ and $\mathbf{y}_b$ and compute their corresponding embeddings $z_a$ and $z_b$ using a ResNet (18 or 50)\footnote{For all of this section, we use the ResNet-18 backbone} \citep{he2016deep} backbone. A mixed input is constructed as:
\begin{equation}
    \mathbf{y}_m = \alpha \mathbf{y}_a + (1 - \alpha) \mathbf{y}_b^{\pi},
\end{equation}
where $\alpha \sim \text{Beta}(1.0, 1.0)$ and $\pi$ is a random permutation of batch indices. The mixed embedding $z_m$ is then computed using the same encoder network. The regularization loss is computed based on the difference between the estimated and ground-truth cross-correlation matrices:
\begin{equation}
    \mathcal{L}_{mix} = \lambda_{mix} \lambda \left( \sum_{i, j} (C_m^a - C_m^{a, gt})^2 + \sum_{i, j} (C_m^b - C_m^{b, gt})^2 \right),
\end{equation}
where $C_m^a$ and $C_m^b$ are the cross-correlation matrices for the mixed embeddings with respect to $z_a$ and $z_b$, and $C_m^{a, gt}$ and $C_m^{b, gt}$ are their corresponding ground-truth values.

The total loss is then:
\begin{equation}
    \mathcal{L} = \mathcal{L}_{BT} + \mathcal{L}_{mix}.
\end{equation}

\subsection{ResNet-DVSIB: Adaptation for Variational Information Bottleneck}

We extend the DVSIB architecture (Sec.~\ref{sec:DVSIB}) by incorporating elements of the Barlow Twins approach and comparing it to Barlow Twins \citep{zbontar2021} and the Mixed Barlow Twins \citep{bandara2023guarding}.

\subsubsection{Encoder \texorpdfstring{$I^E(X;Z_X)$}{Ie(X;Zx)}}
The encoder follows the ResNet-18 backbone up to the 512-dimensional embedding layer, followed by a modified projection head, then a fully connected layer mapping (512, 512). The layer is followed by a Batch normalization and a ReLU activation. Then followed by two parallel output layers of size $(512, k_{Z_{\mu}})$ and $(512, k_{Z_{\log{\sigma^2}}})$ to produce the mean and log-variance of the variational posterior.

\subsubsection{Decoder \texorpdfstring{$I^D(X;Z_X)$}{Id(X;Zx)}}
The decoder reconstructs images from a lower-dimensional feature representation and is designed to mirror a ResNet-18 encoder. It consists of a decoder-projection head and a transposed convolutional decoder. The decoder-projection head is a two-layer fully connected network that projects the embedded dimensionality feature vector ($\mathbb{R}^{k_Z}$) to match the ResNet 18 encoder’s final output dimension ($\mathbb{R}^{512}$), with batch normalization and ReLU activation applied after each layer. The transposed convolutional decoder then progressively upsamples the feature map from $(512,1,1)$ to $(3,32,32)$ through a sequence of six transposed convolutional layers, each followed by batch normalization and ReLU activation, except for the final output layer, with proper padding and strides.

\subsubsection{Mutual Information Estimator \texorpdfstring{$I_{\mathrm{InfoNCE}}(Z_X;Z_Y)$}{InfoNCE(Zx;Zy)}}
\label{InfoNCE}

The mutual information term is estimated using the InfoNCE estimator with a separable critic \citep{oord2018representation}. The critic consists of two networks, $g$ and $h$, each with a fully connected layer mapping: $(k_Z, 512)$ with Xavier uniform initialization \citep{glorot2010understanding}, and a ReLU activation followed by a fully connected final layer of $(512, 8192)$, also with a Xavier uniform initialization. These networks compute $g(Z_X)$ and $h(Z_Y)$, which are then combined via a dot product to form a separable critic function $f(Z_X, Z_Y) = g(X) \cdot h(Y)$ for InfoNCE. The estimator is explained in more detail in Appx.~\ref{app:estimators}

\subsection{Training Setup and Results}
\begin{wrapfigure}{r}{0.5\textwidth}
    \centering
    \includegraphics[width=0.5\textwidth]{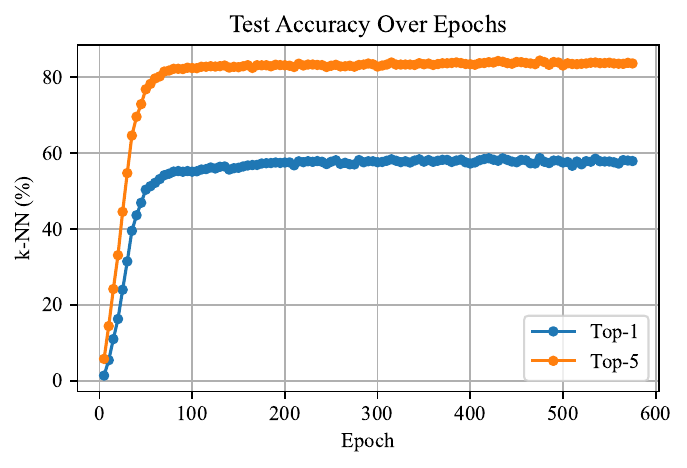}
    \caption{k-NN classification accuracy over training for top-1 and top-5 on the ResNet-DVSIB model using the CIFAR-100 dataset. The figure shows saturation around 50 epochs, with minor fluctuations afterward.}
    \label{sifig:resnet_dvsib_cifar100}
\end{wrapfigure}

To make our ResNet-DVSIB architecture resemble BT, we enforce both encoding terms, $I^E(X;Z_X)$ and $I^E(Y;Z_Y)$, to share the same encoder. Similarly, the decoder is also shared for both views. The representations $Z_X$ and $Z_Y$ are then obtained by passing the two augmented versions, $X$ and $Y$, through this common encoder network\footnote{More generally, this setup can be applied in DVSIB when we want to ensure that the same features are learned from different inputs.}.

We initialize the encoder portion of ResNet-DVSIB (before the projection head) using the pre-trained model from \cite{bandara2023guarding} trained on CIFAR-100 at a $k_Z$ of 1024, where the reported top-1 accuracy is 61\%\footnote{Saved model and code of \cite{bandara2023guarding} available at \url{https://github.com/wgcban/mix-bt/tree/main}. We adapted parts of this code to integrate with ours and made the necessary modifications to align it with our ResNet-DVSIB setup}. We replace the projection head with our variational head at $k_z = 128$, while initializing the decoder and mutual information estimator from scratch. 

During training, and because the encoder portion is already a good starting point, we set its loss weight to zero ($\mathcal{L} = 0*I^E - \beta I^D$) but allow it to train indirectly through the other terms $I^D$ in the loss, but with a learning rate of $0.1 \times$ the base learning rate. The full model is optimized using cosine annealing with an initial learning rate of $10^{-3}$, running for 1000 epochs with early stopping after 100 epochs of no improvement. Training stabilizes around 50 epochs, with a highest top-1 accuracy of 58.6\% (top-5: 84.28\%) achieved around 500 epochs, as shown in Fig.~\ref{sifig:resnet_dvsib_cifar100}.

Noting that \cite{bandara2023guarding} report the top-1 accuracy of the original BT on the CIFAR-100 dataset to be around 50\% for $k_Z=128$ and around 58\% for $k_Z=1024$, while the mixed BT reports around 60\% and 61\% for $k_Z=128, 1024$ respectively\footnote{Refer to Figure 3 of \cite{bandara2023guarding} for BT and Figure 4 for mixed BT. Note that the only explicitly reported value is for the mixed BT model at $k_Z=1024$, which was 61\%, while other values were inferred from the graph.}.

\subsection{Discussion}

Our results show that the adapted DVSIB framework achieves comparable performance to the 1024-dimensional BT model while using only 128 latent dimensions. Additionally, our model not only produces useful representations, but also learns a latent distribution and includes a decoder, allowing for further interpretability. Although no exhaustive tuning was performed, these results demonstrate the viability of the DVSIB method, and the DVMIB framework in general, in achieving near-the-state-of-the-art performance in self-supervised learning.

\clearpage

\bibliography{ref}

\end{document}